\documentclass[letterpaper]{article} 
\usepackage{aaai2026}  
\usepackage{times}  
\usepackage{helvet}  
\usepackage{courier}  
\usepackage[hyphens]{url}  
\usepackage{graphicx} 
\usepackage{natbib}
\usepackage{caption}
\usepackage[ruled,vlined,linesnumbered]{algorithm2e}

\usepackage{balance}
\usepackage{float}

\usepackage{amsmath}
\usepackage{amssymb}
\usepackage{booktabs}
\usepackage{subcaption}
\usepackage{colortbl}
\usepackage{multirow}
\usepackage{enumerate}
\usepackage{nicefrac}
\usepackage{xurl}

\usepackage{xcolor}
\definecolor{rowHL}{HTML}{EAF2FF} 
\newcommand{\colorRowOne}{%
  \rowcolor{rowHL!70}%
  \cellcolor{white}%
  \rule{0pt}{2.0ex}%
}
\newcommand{\colorRow}{%
  \rowcolor{rowHL!70}%
  \rule{0pt}{2.0ex}%
}

\definecolor{mygreen}{HTML}{009900}
\definecolor{deltagreen}{HTML}{308A45}
\definecolor{deltapurple}{HTML}{682075}
\definecolor[named]{ACMBlue}{cmyk}{1,0.1,0,0.1}
\definecolor[named]{ACMYellow}{cmyk}{0,0.16,1,0}
\definecolor[named]{ACMOrange}{cmyk}{0,0.42,1,0.01}
\definecolor[named]{ACMRed}{cmyk}{0,0.90,0.86,0}
\definecolor[named]{ACMLightBlue}{cmyk}{0.49,0.01,0,0}
\definecolor[named]{ACMGreen}{cmyk}{0.20,0,1,0.19}
\definecolor[named]{ACMPurple}{cmyk}{0.55,1,0,0.15}
\definecolor[named]{ACMDarkBlue}{cmyk}{1,0.58,0,0.21}
\definecolor[named]{ACMTangerine}{cmyk}{0,0.60,0.95,0.01}
\definecolor{rowHL}{HTML}{EAF2FF} 

\usepackage[hidelinks,pagebackref=true]{hyperref} 
\renewcommand*\backref[1]{\ifx#1\relax \else (Cited on page #1) \fi}
\hypersetup{
  colorlinks,
  linkcolor=ACMBlue,  
  citecolor=ACMDarkBlue,  
  urlcolor=ACMDarkBlue,  
  filecolor=ACMDarkBlue,
}
\usepackage[nameinlink]{cleveref}

\title{Out-of-Distribution Generalization with a SPARC:\\ Racing 100 Unseen Vehicles with a Single Policy}
\author{
    Bram Grooten\textsuperscript{\rm 1,2}\ \ 
    Patrick MacAlpine\textsuperscript{\rm 1}\ \   
    Kaushik Subramanian\textsuperscript{\rm 1}\ \ 
    Peter Stone\textsuperscript{\rm 1,3}\ \ 
    Peter R. Wurman\textsuperscript{\rm 1}
}
\affiliations{
    \textsuperscript{\rm 1}Sony AI\ \ \ \ 
    \textsuperscript{\rm 2}TU Eindhoven\ \ \ \ 
    \textsuperscript{\rm 3}The University of Texas at Austin
}

\begin{document}

\maketitle

\begin{abstract}
Generalization to unseen environments is a significant challenge in the field of robotics and control. In this work, we focus on contextual reinforcement learning, where agents act within environments with varying contexts, such as self-driving cars or quadrupedal robots that need to operate in different terrains or weather conditions than they were trained for. We tackle the critical task of generalizing to out-of-distribution (OOD) settings, \textit{without} access to explicit context information at test time. Recent work has addressed this problem by training a context encoder and a history adaptation module in separate stages. While promising, this two-phase approach is cumbersome to implement and train. We simplify the methodology and introduce SPARC: \textbf{s}ingle-\textbf{p}hase \textbf{a}daptation for \textbf{r}obust \textbf{c}ontrol.
We test SPARC on varying contexts within the high-fidelity racing simulator \textit{Gran Turismo~7} and wind-perturbed \textit{MuJoCo} environments, and find that it achieves reliable and robust OOD generalization.
\end{abstract}

\begin{links}
    \link{Code}{https://github.com/bramgrooten/sparc}
\end{links}

\section{Introduction}

Deep reinforcement learning (RL) has demonstrated successful performance in fields such as robotics \cite{mahmood2018benchmarking}, nuclear fusion \cite{degrave2022magnetic}, and high-fidelity racing simulators \cite{wurman2022outracing}. Despite these successes, generalizing RL agents to unseen environments with varying contextual factors remains a critical challenge. In real-world applications, environmental conditions such as friction, wind speed, or vehicle dynamics can change unpredictably, often leading to catastrophic failures when the agent encounters out-of-distribution (OOD) contexts that it was not trained for.

A promising approach to tackle this issue is context-adaptive reinforcement learning \cite{benjamins2021carl}, where agents infer and adapt to latent environmental factors by leveraging past interactions. Rapid Motor Adaptation (RMA) \cite{kumar2021rma} is a notable framework in this direction, introducing a two-phase learning procedure. In the first phase, a context encoder is trained using privileged information about the environment. The second phase then employs supervised learning to train a history-based adaptation module, enabling the agent to infer latent context solely from past state-action trajectories. While effective, this two-phase approach introduces complexity during implementation and training.

\begin{figure*}[t]
    \centering
    \includegraphics[width=\linewidth]{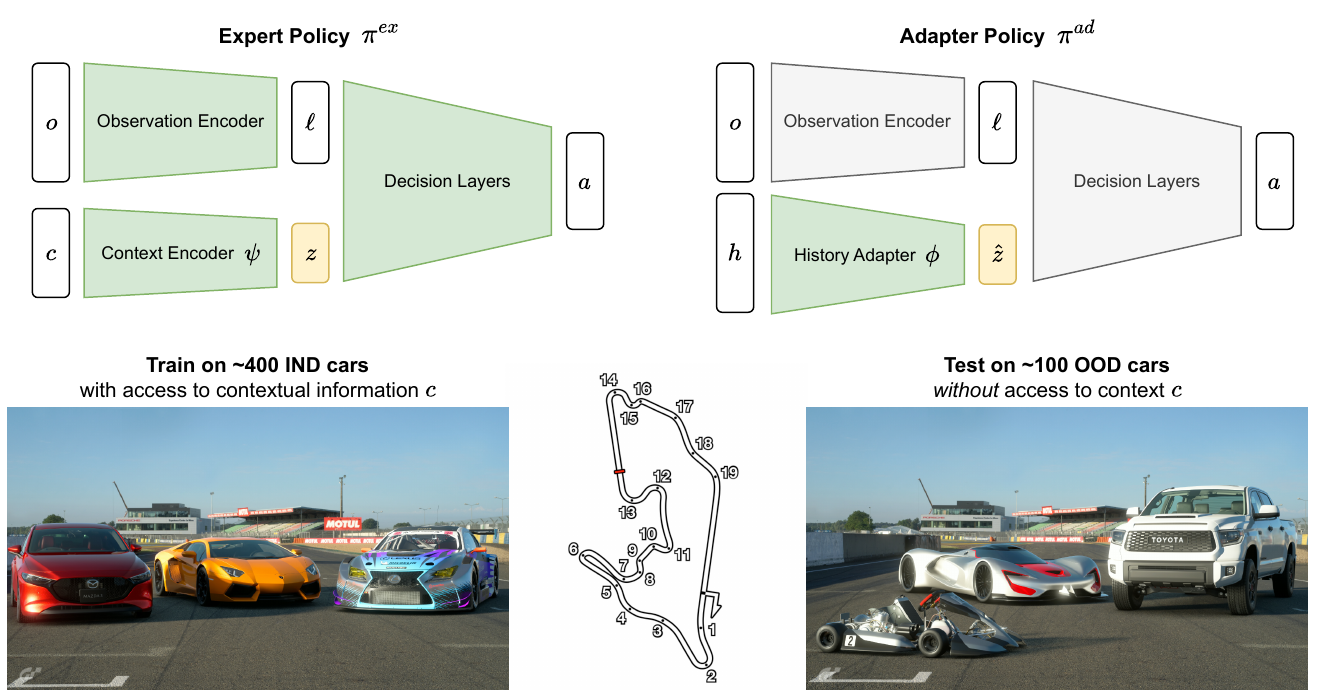}
    \caption{Overview of our algorithm SPARC (\textbf{top}) and the problem setting in Gran Turismo 7 (\textbf{bottom}). SPARC trains an expert policy $\pi^{ex}$ and an adapter policy $\pi^{ad}$ simultaneously in a single phase. The adapter policy does not require access to privileged contextual information, facilitating deployment to OOD real-world scenarios.  
    Observations $o$, contextual information $c$, and a history of recent observation-action pairs $h$ are passed into the networks. Latent encodings $\ell$ and $z$ are concatenated and passed to the final layers, producing action $a$.
    Similar to RMA \cite{kumar2021rma}, $\pi^{ex}$ is trained with reinforcement learning, while the History Adapter $\phi$ of $\pi^{ad}$ is trained with supervised learning to regress its encoding $\phi(h)=\hat{z}$ to the Context Encoder's output $\psi(c)=z$. Note that since SPARC trains in one phase, the context encoding $z$ is a moving target, instead of a traditionally fixed target in RMA. Trainable modules are in green. The black modules regularly copy weights from their counterpart in $\pi^{ex}$. 
    }
    \label{fig:overview_sparc}
\end{figure*}

In this work, we introduce \textbf{SPARC} (\textbf{s}ingle-\textbf{p}hase \textbf{a}daptation for \textbf{r}obust \textbf{c}ontrol), a novel algorithm that unifies context encoding and adaptation into a single training phase, as illustrated in \Cref{fig:overview_sparc}. 
SPARC is straightforward to implement and naturally integrates with off-policy training as well as asynchronous distributed computation on cloud-based rollout workers.
Algorithms such as SPARC and RMA are advantageous when explicit context labels are unavailable at test time, a frequent limitation in real-world robotic deployment.
By collapsing adaptation into a single training loop, SPARC is naturally compatible with on-device continual learning---especially applicable in settings where retraining in the cloud is prohibitive due to privacy or latency constraints. In contrast, RMA is unable to perform continual learning in a straightforward manner.

We evaluate SPARC on two distinct domains: (1) a set of \textit{MuJoCo} environments featuring strongly varying environment dynamics through the use of wind perturbations, and (2) a high-fidelity racing simulator, \textit{Gran Turismo 7}, where agents must adapt to different car models on multiple tracks. 
SPARC achieves state-of-the-art generalization performance and consistently produces Pareto-optimal policies when evaluated across multiple desiderata.

Our contributions are summarized as follows.
\begin{itemize}
    \item We introduce SPARC, a novel single-phase training method for context-adaptive reinforcement learning, eliminating the need for separate encoder pre-training.
    \item We empirically validate SPARC’s generalization ability across OOD environments, demonstrating competitive or superior performance compared to existing approaches.
    \item We perform and analyze several ablation studies, examining key design choices such as history length and the selection of rollout policy during training.
\end{itemize}

\section{Related Work}
\label{sec:related_work}

Generalization to out-of-distribution environments is a fundamental challenge in reinforcement learning, hindering its deployment in real-world applications, particularly in robotics and control tasks \cite{kirk2023survey}. The learning dynamics of RL methods often struggle to adapt to novel environmental conditions \cite{lyle2022learning}. Contextual reinforcement learning \cite{langford2017,benjamins2021carl} provides a framework to address this problem by training agents capable of adapting to varying environmental factors.

\subsection{Contextual RL}

Robust RL often depends on effective contextual adaptation. 
Recent work has explored context-aware policies that integrate contextual cues into decision-making \cite{beukman2023dynamics, chen2021context, lahmer2024fast} or employ world models to capture environment dynamics \cite{lee2020context, prasanna2024dreaming}.
In addition, several studies have focused on modifying the environment itself—such as by varying gravity or adjusting agent component dimensions—to promote the development of more versatile controllers \cite{benjamins2021carl, leon2024duplex}.

\subsection{What if the Agent has No Access to Context?}

In many real-world scenarios, agents are deprived of explicit contextual information during deployment. In these cases, the agent must infer the relevant environmental factors indirectly. For instance, \citet{lee2020science} advanced robust legged locomotion by introducing a two-phase learning process. It first trains an expert policy, which includes a context encoder using the privileged contextual information. 
The second phase involves an adapter policy that tries to imitate the expert's action, while a history-based adaptation component aims to minimize the difference between its history encoding and the expert's context encoding.
Rapid Motor Adaptation \cite{kumar2021rma} refines this methodology by only imitating the context encoding, not the action. The adapter policy can be deployed, as it does not require access to the privileged context.

\subsection{Other Techniques for Generalization}

Several complementary approaches have been proposed to enhance generalization. Domain randomization \cite{tobin2017domain, peng2018sim} and procedurally generated environments \cite{procgen, gisslen2021adversarial} introduce diversity during training, thereby encouraging robust policy behavior.
We employ domain randomization by default in our experiments.
System identification methods \cite{yu2017preparing}—whether performed explicitly or through implicit online adaptation, as in SPARC and RMA—also contribute to improved performance under varying conditions. 
Moreover, techniques such as data augmentation \cite{laskin2020reinforcement, hansen2021stabilizing, Wang_Wu_Hu_Wang_Lin_Lv_2024} and masking \cite{grooten2024madi, huang2022spectrum} have been shown to further enhance generalization, particularly for pixel-based inputs.

Meta-reinforcement learning offers an alternative paradigm for learning adaptable policies \cite{wang2016learning, rabinowitz2018machine, duan2016rl}.
Foundational algorithms like Model-Agnostic Meta-Learning (MAML) \cite{finn2017model} enable rapid task adaptation, and emerging methods using hypernetworks generate task-specific policy parameters on the fly \cite{beck2023hypernetworks, rezaei2023hypernetworks, beukman2023dynamics}.

\section{Background}
\label{sec:background}

In this section, we formalize the underlying problem framework and examine the core techniques that form the foundation for SPARC, enabling context-adaptive behavior.

\subsection{Problem Formulation}
\label{sec:problem_formulation}

We consider a contextual Markov decision process (CMDP) \cite{hallak2015contextual, abbasi2014online}, defined by \citet{kirk2023survey} as a tuple
$\mathcal{M} = (\mathcal{S}, \mathcal{A}, \mathcal{O}, \mathcal{C}, R, T, O, p_s, p_c)$
where:

\begin{itemize}
    \item $\mathcal{S}$ is the state space, 
    \item $\mathcal{A}$ is the action space, 
    \item $\mathcal{O}$ is the observation space,
    \item $\mathcal{C}$ is the context space,
    \item $R : \mathcal{S} \times \mathcal{A} \times \mathcal{C} \rightarrow \mathbb{R}$ is the reward function,
    \item $T : \mathcal{S} \times \mathcal{A} \times \mathcal{C} \rightarrow \Delta(\mathcal{S})$ defines the stochastic transition dynamics conditioned on a context $c \in \mathcal{C}$,
    \item $O: \mathcal{S} \times \mathcal{C} \rightarrow \mathcal{O}$ is the observation function,
    \item $p_s : \mathcal{C} \rightarrow \Delta(\mathcal{S})$ is the distribution over initial states $s_0$ given a context $c \in \mathcal{C}$, and 
    \item $p_c \in \Delta(\mathcal{C})$ is the distribution over contexts.
\end{itemize}

During training, the agent will be exposed to a certain subset of contexts $\mathcal{C}_{\text{IND}} \subset \mathcal{C}$, which are in-distribution (IND), short for \textit{within the training distribution}. 
To test generalization ability, we hold out a different subset of contexts $\mathcal{C}_{\text{OOD}} \subset \mathcal{C}$ that are out-of-distribution (OOD). We ensure that there is no overlap: $\mathcal{C}_{\text{IND}} \cap \mathcal{C}_{\text{OOD}} = \varnothing$. 
This separation defines two sub-CMDPs: $\mathcal{M}_{\text{IND}}$ and $\mathcal{M}_{\text{OOD}}$. We specify the context distributions to be uniform over their respective subsets:
\begin{equation}\label{eq:pc}
    p_c^{i}(c) = \begin{cases}
        \frac{1}{|\mathcal{C}_i|}  & \text{if } c \in \mathcal{C}_i \\
        0  & \text{otherwise,}
    \end{cases}
\end{equation}
for $i \in \{ \text{IND, OOD}\}$.

In our setting, the agents do not observe $c$ at test time and must infer it through other means, for example from their interaction history. However, for comparison, we will also present results of an expert policy that \textit{does} have access to the privileged context information $c \in \mathcal{C}_{\text{OOD}}$ at evaluation.

Our objective is to train a policy $\pi$ that maximizes expected return across both in-distribution and out-of-distribution contexts, while only having access to privileged contextual information $c \in \mathcal{C}_{\text{IND}}$ during training.

\subsection{Pure History-based Policies}
\label{sec:history}

History-based policies have emerged as a powerful approach in reinforcement learning for inferring hidden environmental context from past interactions. Instead of relying solely on the current observation \(o_t \in \mathcal{O}\), these policies condition action selection on a sequence of recent observation-action pairs. 
Let $H$ be the history length and
$\mathcal{H} = (\mathcal{O}\!\times\!\mathcal{A})^{H}$, the space of possible histories.  
For time $t$ we define the corresponding history $h_t$ as
\[
h_t=(o_{t-H:t-1},a_{t-H:t-1})\in\mathcal{H}.
\]
This history input results in policies of the form 
\( \pi: \mathcal{O} \times \mathcal{H} \rightarrow \Delta(\mathcal{A}) \).
Including the history may enable the agent to implicitly capture latent context information \(c \in \mathcal{C}\), as the context $c$ may influence the environment dynamics.

A pure history-based approach is presented by \citet{lee2020science} as a strong baseline. In their work on quadrupedal locomotion over challenging terrains, the authors demonstrate that leveraging an extended history of proprioceptive data via a temporal convolutional network (TCN) enables robust control in diverse settings.

\section{Method}
\label{sec:method}

\subsection{Leveraging Contextual Information}
\label{sec:rma}

Training with privileged contextual information---even if not available at test time---has been shown to be particularly useful for generalizing to OOD contexts.
In that regard, the approaches by \citet{lee2020science} and \citet{kumar2021rma} are almost equivalent; we will focus on Rapid Motor Adaptation (RMA) \cite{kumar2021rma}.
In RMA, two policies are trained in separate phases. First, the expert policy 
\[
\pi^{ex}_{\theta}: \mathcal{O} \times \mathcal{C} \rightarrow \Delta(\mathcal{A})
\]
which includes a context encoder $\psi(\cdot)$ with access to the environment's privileged information, is trained using a reinforcement learning algorithm. While the original RMA work uses PPO \cite{schulman2017proximal}, we employ the more sample-efficient QR-SAC, proven to work well in Gran Turismo \cite{wurman2022outracing}.

Once training of $\pi^{ex}_{\theta}$ has converged to a sufficient level, the best model checkpoint $\pi^{ex}_{\theta^*}$ needs to be determined.
This selection requires careful evaluation across multiple dimensions \cite{morrill2023composing}, a cumbersome intermediate step that SPARC skips, as it is trained in a single phase.

The second stage of RMA trains the adapter policy 
\[
\pi^{ad}_\theta: \mathcal{O} \times \mathcal{H} \rightarrow \Delta(\mathcal{A})
\]
while keeping the expert policy $\pi^{ex}_{\theta^*}$ frozen. 
In the adapter policy, a \emph{history adapter} $\phi_\theta$ processes a sequence of recent observation-action pairs $h_t$
to produce a latent representation
$\hat{z}_t = \phi_\theta(h_t)$. The history adapter is trained by minimizing the distance between $\hat{z}_t = \phi_\theta(h_t)$ and $z_t = \psi_{\theta^*}(c_t)$ through the mean squared error loss:
\begin{equation}\label{eq:loss_adapter}    
\mathcal{L}_{\phi}(c_t, h_t) = \mathbb{E}_{c_t,h_t}[(z_t - \hat{z}_t)^2].
\end{equation}

The history-inferred latent context $\hat{z}_t$ is integrated into the policy.
By conditioning on both the current observation \(o_t\) and the latent context \(\hat{z}_t\), the policy can adjust its behavior to handle unseen or varying environmental conditions.

\begin{table*}[]
    \centering
    \caption{Performance summary on IND and OOD settings across all test tracks in Gran Turismo, averaged over 3 seeds. Results show the mean built-in AI (BIAI) ratio across cars (ratio = the RL agent's lap time divided by the BIAI lap time, lower is better). If an algorithm fails to complete a lap with a specific vehicle, it receives a BIAI ratio of 2.0 for that car model. Additionally, we show the percentage of cars with a successfully completed lap ($\pm$ s.e.m.). We \textbf{bold} the best out-of-distribution results across algorithms without access to context at test time (all except Oracle, see \Cref{tab:inputs}). We include IND results for reference. SPARC achieves the fastest OOD lap times on 2 of 3 tracks and completes the most laps with OOD vehicles overall.} 
    \label{tab:gt_summary_biai2}
    \resizebox{0.85\textwidth}{!}{%
    \begin{tabular}{ll@{\hspace{1em}}cc@{\hspace{1em}}cc}
        \toprule
        \multirow{2}{*}{Race Track} & \multirow{2}{*}{Method} & \multicolumn{2}{c}{IND} & \multicolumn{2}{c}{OOD} \\
        \cmidrule(lr){3-4} \cmidrule(lr){5-6}
         & & BIAI ratio ($\downarrow$) & Success \% ($\uparrow$) & BIAI ratio ($\downarrow$) & Success \% ($\uparrow$) \\
        \midrule
        \multirow{5}{*}{Grand Valley} 
            & Only Obs       & 0.9929 $\pm$ 0.0007   & 100.00 $\pm$ 0.00   & 1.0641 $\pm$ 0.0058   & 95.15 $\pm$ 0.56   \\
            & History Input    & 0.9904 $\pm$ 0.0001   &99.68 $\pm$ 0.08   & 1.0826 $\pm$ 0.0203   & 92.56 $\pm$ 2.12   \\
            & RMA              & 1.0046 $\pm$ 0.0054   & 99.84 $\pm$ 0.16   & 1.0560 $\pm$ 0.0134   & 97.09 $\pm$ 1.12   \\
            \colorRowOne
            & SPARC            & 0.9999 $\pm$ 0.0061 & 99.76 $\pm$ 0.14 & \textbf{1.0491 $\pm$ 0.0055} & \textbf{98.06 $\pm$ 0.56} \\
            & Oracle    & 0.9884 $\pm$ 0.0005   & 100.00 $\pm$ 0.00   & 1.1348 $\pm$ 0.0137   & 90.94 $\pm$ 2.27   \\
        \midrule
        \multirow{5}{*}{N\"{u}rburgring} 
            & Only Obs       & 1.0202 $\pm$ 0.0163   & 95.87 $\pm$ 1.48   & 1.1745 $\pm$ 0.0129   & 81.88 $\pm$ 1.17   \\
            & History Input    & 0.9984 $\pm$ 0.0030   & 97.49 $\pm$ 0.32   & 1.1204 $\pm$ 0.0132   & 86.73 $\pm$ 1.29   \\
            & RMA              & 1.1085 $\pm$ 0.0195   & 88.03 $\pm$ 1.76   & 1.2995 $\pm$ 0.0306   & 77.99 $\pm$ 3.19   \\
            \colorRowOne
            & SPARC            & 1.0254 $\pm$ 0.0061 & 95.87 $\pm$ 0.49 & \textbf{1.1199 $\pm$ 0.0076} & \textbf{89.00 $\pm$ 0.86} \\
            & Oracle    & 0.9804 $\pm$ 0.0027   & 99.27 $\pm$ 0.28   & 1.1182 $\pm$ 0.0215   & 89.64 $\pm$ 2.53   \\
        \midrule
        \multirow{5}{*}{Catalunya Rallycross} 
            & Only Obs       & 0.9319 $\pm$ 0.0009   & 100.00 $\pm$ 0.00   & 0.9560 $\pm$ 0.0006   & \textbf{100.00 $\pm$ 0.00}   \\
            & History Input    & 0.9294 $\pm$ 0.0001   & 100.00 $\pm$ 0.00   & \textbf{0.9553 $\pm$ 0.0068}   & 99.33 $\pm$ 0.67   \\
            & RMA              & 0.9445 $\pm$ 0.0010   & 99.82 $\pm$ 0.18   & 0.9667 $\pm$ 0.0030   & \textbf{100.00 $\pm$ 0.00}   \\
            \colorRowOne
            & SPARC            & 0.9432 $\pm$ 0.0027 & 100.00 $\pm$ 0.00 & 0.9631 $\pm$ 0.0026 & \textbf{100.00 $\pm$ 0.00} \\
            & Oracle    & 0.9282 $\pm$ 0.0001   & 100.00 $\pm$ 0.00   & 1.1354 $\pm$ 0.0595   & 85.33 $\pm$ 5.81   \\
        \bottomrule
    \end{tabular}%
    }
\end{table*}

\subsection{Single-Phase Adaptation}
\label{sec:sparc_single}

Our algorithm illustrated in \Cref{fig:overview_sparc}, SPARC, greatly simplifies the implementation and training of agents capable of generalizing to out-of-distribution environments without access to privileged contextual information.
In SPARC, the expert policy $\pi^{ex}$ and the adapter policy $\pi^{ad}$ are trained simultaneously, in contrast to the two-phase approach of RMA. This means that the context encoding $\psi(c)= z$ is a non-stationary target for the history adapter $\phi$, instead of a fixed target. The results in \Cref{sec:results} demonstrate that the adapter policy is able to manage these new learning dynamics. 

An important detail in RMA is which model acts in the environment to collect experience. Policy $\pi^{ex}$ acts in the first training phase, while $\pi^{ad}$ does so in the second.
This raises the question which policy should gather experience for SPARC, as both are trained together. One option would be to let the expert policy $\pi^{ex}$ control the actions, since it is updated and improved through QR-SAC. 

However, the expert policy, $\pi^{ex}$, is not the goal of the SPARC approach. A robust adapter policy, $\pi^{ad}$, is the overall learning target and using this policy to gather experience allows the learning algorithm to correct for any inaccuracies before final deployment. This brings the learning dynamics of $\pi^{ad}$ closer to an on-policy setting, even though its history adapter \(\phi\) is trained through supervised learning as shown in Equation \ref{eq:loss_adapter}. 
We perform an ablation study on this choice of rollout policy in the supplementary material. 

SPARC's critic networks---necessary to run QR-SAC on $\pi^{ex}$---have the same architecture as the expert policy 
and thus have access to the context $c$. We can still run inference without knowing $c$, because at test time only $\pi^{ad}$ is needed.

Reducing training of SPARC to one phase provides several benefits: (i) no intermediate selection of the best model checkpoint of the first phase is necessary, (ii) training can be easily continued indefinitely, without having to retrain the second phase, (iii) the simpler implementation facilitates the use of SPARC on asynchronous distributed systems.

\section{Experimental Setup}
\label{sec:experimental_setup}

In this section, we describe the experimental setup used to evaluate SPARC, our proposed single-phase adaptation method, in comparison with several baselines. 

\subsection{Environments}
\label{sec:environments}

We evaluate our approach on two distinct domains:
\begin{itemize}
    \item \textbf{MuJoCo:} A suite of continuous control tasks including \textit{HalfCheetah}, \textit{Hopper}, and \textit{Walker2d} \cite{todorov2012mujoco}. We induce contextual variability by perturbing the environment's wind speed in multiple dimensions and scales, creating challenging OOD scenarios.
    \item \textbf{Gran Turismo 7:} A high-fidelity racing simulator that features diverse car models and realistic vehicle-track dynamics \cite{wurman2022outracing}. The simulator’s rich contextual variability makes it an ideal testbed for assessing generalization to unseen conditions. 
\end{itemize}
Within Gran Turismo, we experiment on two settings: (1) generalization across car models, and (2) generalization across differing engine power and vehicle mass settings for one specific car.
The in-distribution (IND) training set and OOD test set are selected as follows:
\begin{enumerate}[(1)]
    \item \emph{Car Models}: we sort all $\sim$500 vehicles by their anomaly score through an isolation forest 
    on the car's contextual features such as mass, length, width, weight distribution, power source type, drive train type, wheel radius, etc. 
    We hold out the 20\% most \emph{outlier} vehicles as a test set (OOD) and train on the 80\% most \emph{inlier} cars (IND).
    \item \emph{Power \& Mass:} for a more controlled experiment, we pick a relatively standard racing car, but tune its engine power and mass in each training episode to randomly sampled values within the range [75\%, 125\%] of their defaults. During evaluation, we test on fixed-spaced intervals within [50\%, 150\%], covering IND and OOD contextual settings.
\end{enumerate}
For the wind-perturbed MuJoCo environments, we similarly train on a certain range of wind speeds, while testing on intervals twice as large.
In Gran Turismo, we experiment on three different tracks, presented in \Cref{tab:tracks}. These tracks represent highly varying settings, with \emph{Catalunya Rallycross} even including a mixed dirt and tarmac racing path.

\begin{table}[]
    \centering
    \caption{The Gran Turismo tracks which we experiment on in the \emph{Car Models} setting. The road type and track length pose varying challenges.}
    \label{tab:tracks}
    \resizebox{\columnwidth}{!}{%
    \begin{tabular}{lll}
        \toprule
        Track & Length & Road Type \\ 
        \midrule
        Grand Valley & 5.099 km & Tarmac \\   
        N\"{u}rburgring & 25.378 km & Tarmac + Concrete \\ 
        Catalunya Rallycross & 1.133 km & Dirt + Tarmac \\  
        \bottomrule
    \end{tabular}
    }
\end{table}

\begin{table}[]
    \centering
    \caption{The set of inputs that each algorithm receives.}
    \label{tab:inputs}
    \resizebox{\columnwidth}{!}{%
    \begin{tabular}{lll}
        \toprule
        Method & Inputs during \textbf{Training} & Inputs at \textbf{Test Time} \\ 
        \midrule
        Only Obs & \texttt{obs} & \texttt{obs} \\
        History Input & \texttt{obs}, \texttt{history} & \texttt{obs}, \texttt{history} \\
        RMA & \texttt{obs}, \texttt{history}, \texttt{context} & \texttt{obs}, \texttt{history} \\
        \colorRow
        SPARC & \texttt{obs}, \texttt{history}, \texttt{context} & \texttt{obs}, \texttt{history} \\
        Oracle & \texttt{obs}, \texttt{context} & \texttt{obs}, \texttt{context} \\
        \bottomrule
    \end{tabular}
    }
\end{table}

\subsection{Training Details}

We repeat our runs with independent random seeds to ensure statistical robustness: three seeds for the compute-heavy Gran Turismo simulator, and five for MuJoCo environments. Key training hyperparameters—such as the history length \(H\), learning rates, and network architectures—are tuned through preliminary experiments with grid search.
We train all methods asynchronously, collecting experience on distributed rollout workers. Further training details and analyses are provided in the supplementary material.

\begin{figure}[]
    \centering
    \includegraphics[width=\columnwidth]{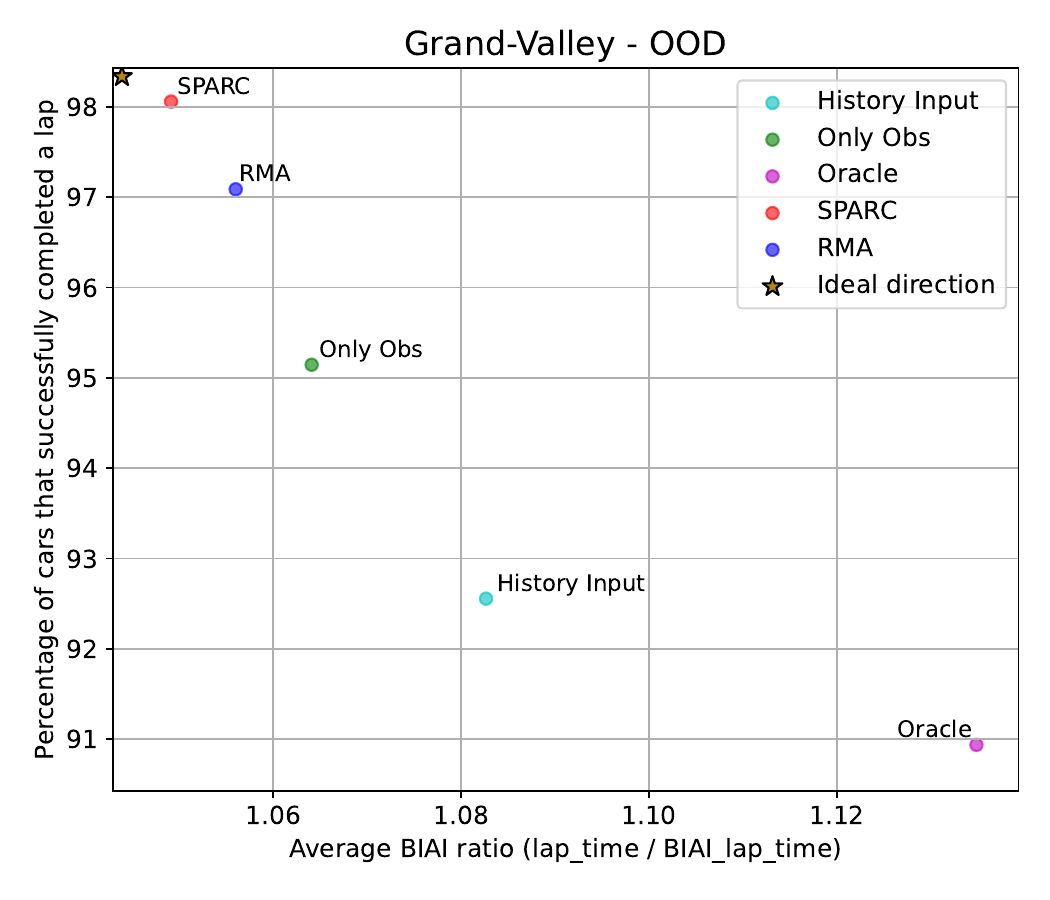}
    \caption{Results on Grand Valley averaged over three seeds. For each algorithm, we plot the percentage of cars that successfully completed laps, and the built-in AI ratio lap time. SPARC is able to complete the most and the fastest laps on out-of-distribution cars.}
    \label{fig:main_gt_scatter}
\end{figure}

\subsection{Evaluation Protocol}
We evaluate policy performance under two settings:
\begin{itemize}
    \item \textbf{In-Distribution (IND):} Environments with contextual parameters that lie within the training distribution.
    \item \textbf{Out-of-Distribution (OOD):} Contextual parameters that deviate significantly from the training set, testing the model's generalization capabilities.
\end{itemize}
During training, we regularly evaluate the policy on three predetermined IND settings. 
These training evaluations form a Pareto-front, from which we select the best model checkpoint for each run. 
We then test the policy on a wide range of IND \& OOD contexts. For \emph{Car Models}, this means all unique vehicles, while for \emph{Power \& Mass} and \emph{MuJoCo} we divide the widest context ranges into fixed intervals, providing up to $21^2=441$ test environments. 

Performance metrics include the return for \emph{MuJoCo} and lap times in \emph{Gran Turismo}. However, for particularly difficult outlier cars, some algorithms may not be able to complete any laps. For this reason, we present the racing results along two dimensions: (1) percentage of cars with a completed lap, and (2) the average lap time. Note that (2) is a biased metric, so (1) needs to be taken into account. 

When averaging raw lap times, slower cars have a larger impact on the average. To avoid skewed results, we divide by the built-in AI (BIAI) lap time for each specific car. The BIAI is a classical control method implemented in Gran Turismo 7 to follow a preset driving line. This \emph{BIAI ratio} of RL lap time over BIAI lap time provides an informative normalized value. When a car was unable to complete a lap, we set its BIAI ratio to $2.0$ before averaging over all vehicles.

\subsection{Baselines}
We compare the performance of the following algorithms.
\begin{itemize}
    \item \textbf{Only Obs:} This QR-SAC \cite{wurman2022outracing} policy is trained without any context information. Only the current observation is provided as input. 
    \item \textbf{History Input:} A baseline policy \cite{lee2020science} that additionally receives a history of observation-action pairs $h_t = (o_{t-H:t-1},a_{t-H:t-1})$.
    \item \textbf{RMA:} The two-phase approach of Rapid Motor Adaptation~\cite{kumar2021rma}, first trains an expert policy with context input, then learns the adapter policy from history.
    \item \textbf{SPARC:} Our single-phase adaptation technique introduced in this work. At test time it only receives an observation-action history and the current observation.
    \item \textbf{Oracle:} A policy that has access to the ground-truth unencoded contextual features, even at test time. 
\end{itemize}
Benchmarking SPARC against the listed baselines allows us to isolate the benefits of our single-phase training paradigm, especially regarding implementation simplicity and OOD generalization. See \Cref{tab:inputs} for a concise overview of the inputs per algorithm.

\section{Results}
\label{sec:results}

We present the performance of SPARC and several baselines on \textit{Gran Turismo} and \textit{MuJoCo} environments, focusing on generalization to unseen contexts.

\subsection{Gran Turismo: Car Models}
\label{sec:gt_results}

The scatterplot in \Cref{fig:main_gt_scatter} summarizes the performance of each algorithm averaged over all out-of-distribution cars on the race track Grand Valley. 
The results indicate that SPARC outperforms the baselines across unseen vehicles during training. SPARC completes laps with the most cars and with the fastest average built-in AI ratio lap time. 

\Cref{tab:gt_summary_biai2} provides a quantitative summary of our findings across all three tracks. 
On IND settings, SPARC is competitive, but it is particularly designed to handle OOD dynamics. When racing untrained cars, SPARC is the fastest of all algorithms without access to context at test time on 2 out of 3 tracks. Furthermore, our method manages to complete laps with the most OOD vehicles on aggregate. 

SPARC even outperforms its two-phase counterpart RMA. We believe this occurs because SPARC avoids the brittle selection of a phase-1 checkpoint, required by RMA. Training an adapter $\phi$ against one checkpoint of $\psi$ can overfit to parts of the context space, while SPARC learns against multiple strong checkpoints over time.

\begin{table}[]
    \centering
    \caption{Performance summary of the \textit{Power \& Mass} experiments, averaged over 3 seeds. Results show the mean built-in-AI lap-time ratio (2.0 if no lap completed) across all OOD power \& mass settings, and the percentage of these settings with a successfully completed lap ($\pm$ s.e.m.). SPARC completes the most and has the fastest laps.}
    \label{tab:bop_ood}
    \resizebox{\linewidth}{!}{%
    \begin{tabular}{l@{\hspace{1em}}cc}
        \toprule
        Method & \multicolumn{1}{c}{Built-in-AI lap-time ratio ($\downarrow$)} & \multicolumn{1}{c}{Success \% ($\uparrow$)} \\
        \midrule
        Only Obs          & 1.0131 $\pm$ 0.0136 & 98.75 $\pm$ 1.25 \\
        History Input     & 1.0135 $\pm$ 0.0013 & 98.33 $\pm$ 0.10 \\
        RMA               & 1.0004 $\pm$ 0.0030 & 99.17 $\pm$ 0.28 \\
        \colorRow
        SPARC             & \textbf{0.9907 $\pm$ 0.0011} & \textbf{99.90 $\pm$ 0.10} \\
        Oracle            & 0.9962 $\pm$ 0.0067 & 99.27 $\pm$ 0.58 \\
        \bottomrule
    \end{tabular}%
    }
\end{table}

\subsection{Gran Turismo: Power \& Mass}
\label{sec:bop_results}

\begin{figure*}
  \centering
  \hfill
  \begin{subfigure}{0.32\textwidth}
    \includegraphics[width=\linewidth]{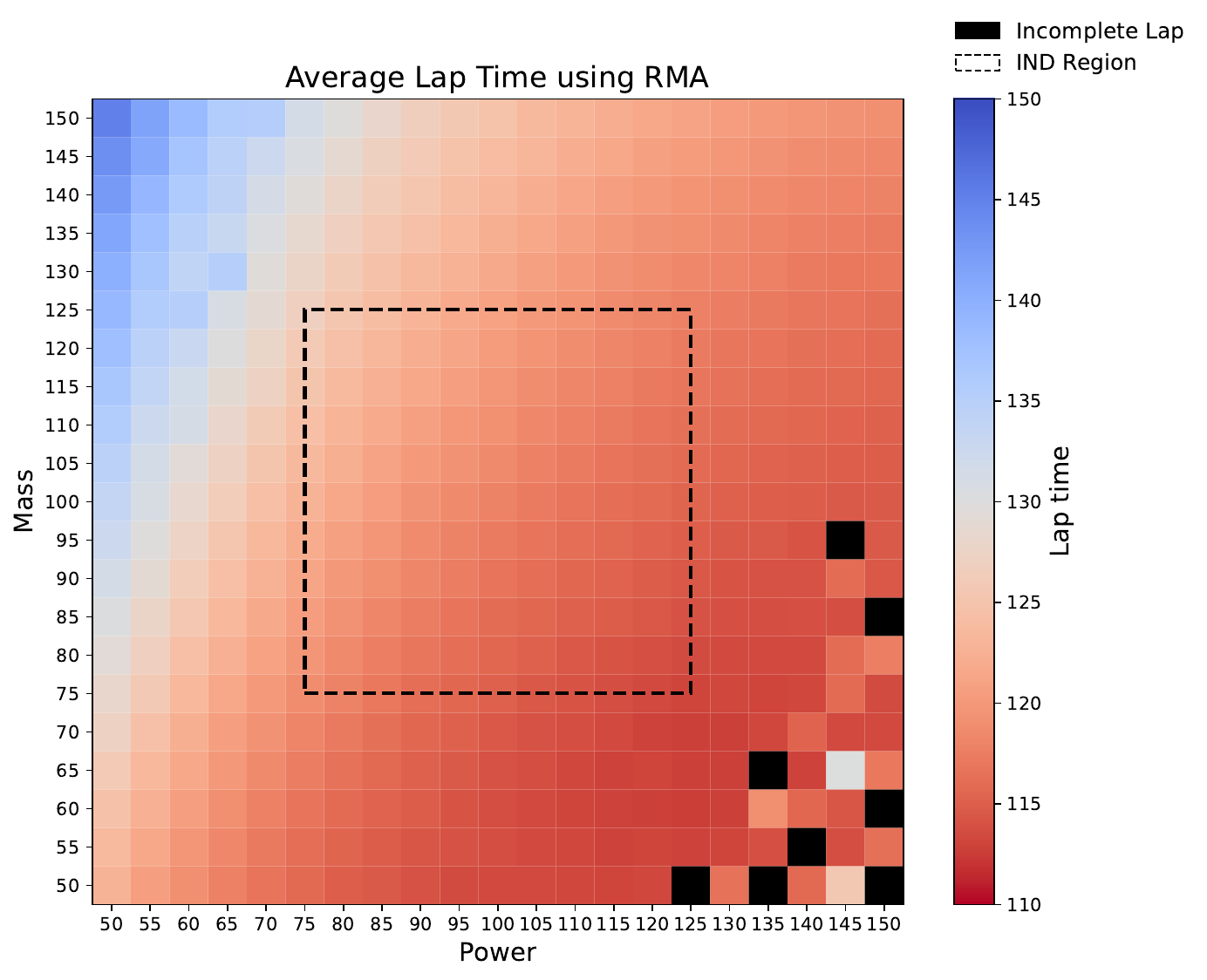}
    \caption{RMA}
    \label{fig:sub1}
  \end{subfigure}
  \hfill
  \begin{subfigure}{0.32\textwidth}
    \includegraphics[width=\linewidth]{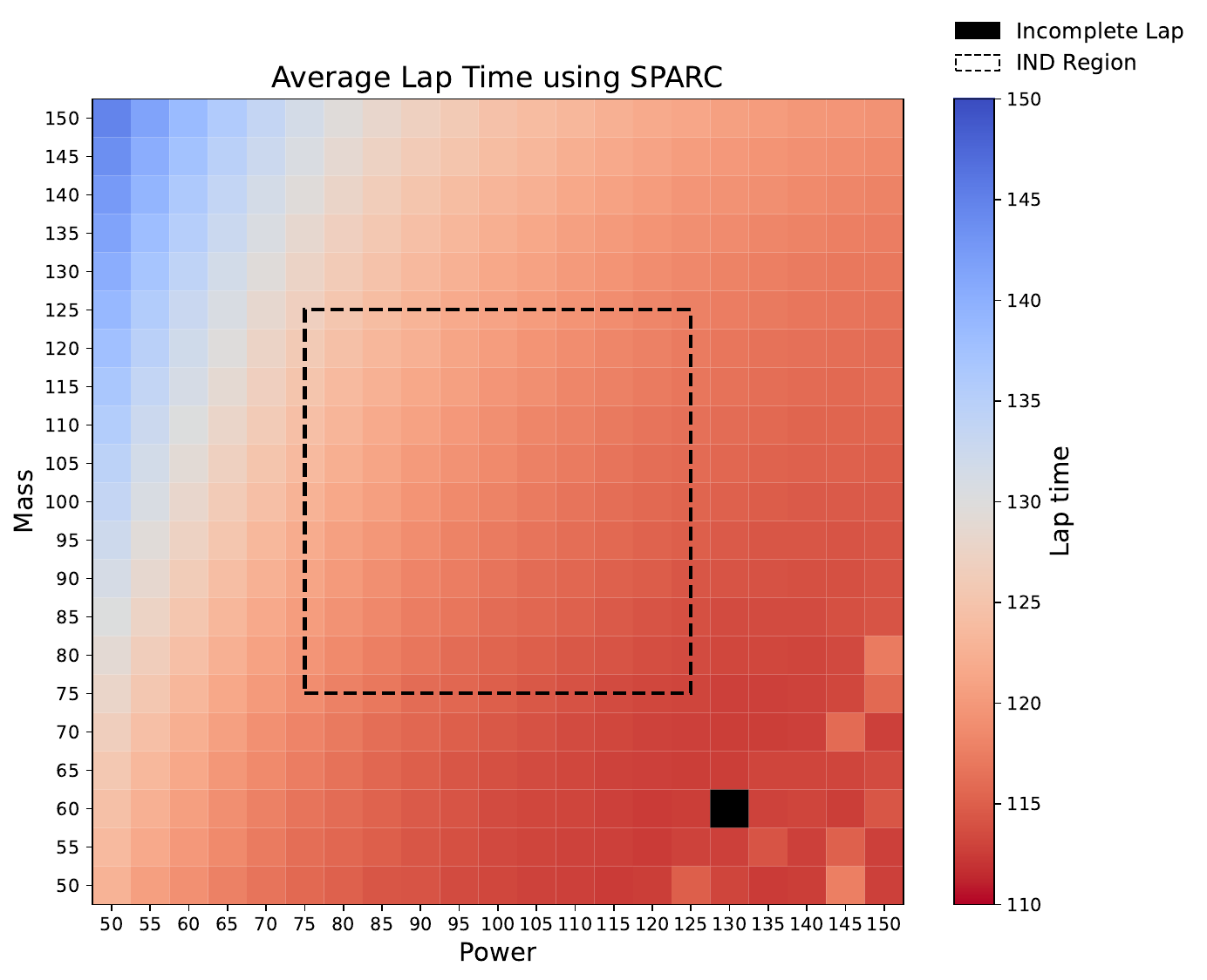}
    \caption{SPARC}
    \label{fig:sub2}
  \end{subfigure}
  \hfill
  \begin{subfigure}{0.32\textwidth}
    \includegraphics[width=0.92\linewidth]{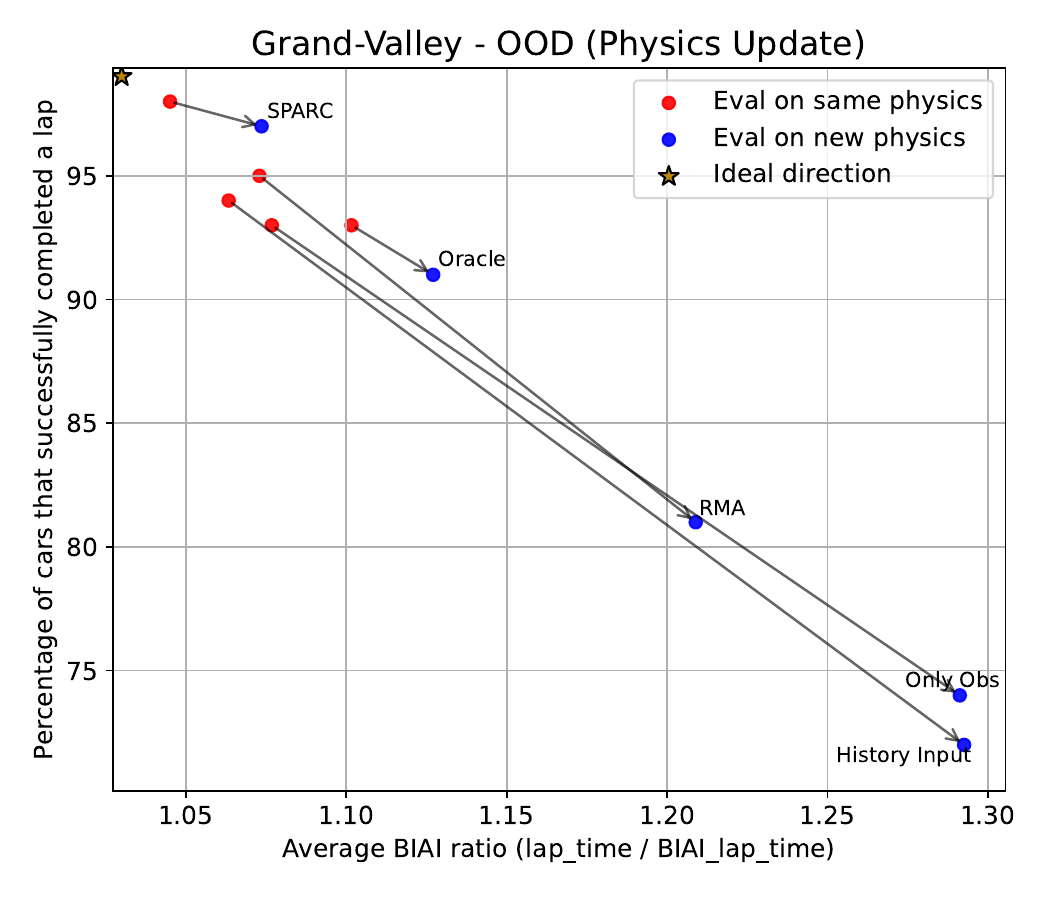}
    \caption{OOD Physics}
    \label{fig:across_game_dynamics}
  \end{subfigure}
  \hfill
  \caption{
  \textbf{(a)} and \textbf{(b)}:
  Lap times on the \emph{Power \& Mass} experiment. 
  Colours denote average lap time over 3 seeds (red\ =\ fast, blue\ =\ slow); black squares indicate at least one unfinished lap.
  Even though both algorithms are trained only on settings within the IND region (dashed box), SPARC is able to handle challenging OOD settings in the bottom-right corner (high power and low mass).
  \textbf{(c)}: Performance difference between old and new game dynamics. These algorithms have only been trained on old physics settings, and are tested zero-shot on the new physics after a game update of Gran Turismo. SPARC shows the best OOD generalization, with only slightly slower lap times on new dynamics, while other methods degrade significantly.
  }
  \label{fig:bop}
\end{figure*}

In \Cref{fig:bop} we show the difference between (a) the strongest baseline and (b) our method. SPARC is able to complete laps in almost all OOD contextual settings, while RMA struggles in the most difficult scenarios of lightweight cars with high engine power.
\Cref{tab:bop_ood} provides a summary of the average results across all OOD contexts, indicating that SPARC outperforms all baselines, including the Oracle method. 
The Oracle does not receive history as part of its inputs, as opposed to other baselines. We believe that SPARC is even able to outperform the Oracle in this environment because knowledge of some history may be useful to mitigate the partial observability in our contextual MDP $\mathcal{M}$ (\Cref{sec:problem_formulation}).
SPARC is the most robust in this experiment---completing laps in all but one setting---and also achieves the fastest average built-in-AI lap-time ratio.

\newcolumntype{R}{%
  r@{\hspace{0.725em}} 
  >{$\pm$\hspace{0.7em}}  
  l                            
}

\begin{table}[]
    \centering
    \caption{Performance across MuJoCo environments, averaged over 5 seeds. Results show the mean return over all out-of-distribution wind perturbations ($\pm$ s.e.m.). SPARC outperforms all baselines in 2 out of 3 environments.}
    \label{tab:mujoco}
    \resizebox{\linewidth}{!}{%
    \begin{tabular}{lRRR}
    \toprule
    Method
          & \multicolumn{2}{c}{HalfCheetah-v5 ($\uparrow$)}
          & \multicolumn{2}{c}{Hopper-v5 ($\uparrow$)}
          & \multicolumn{2}{c}{Walker2d-v5 ($\uparrow$)}\\
    \midrule
    Only Obs      & 5724.51 & 1624.98 & 1274.13 & 133.78 & 2495.77 & 220.69\\
    History Input & 8760.12 &  161.53 & 1367.09 &  67.79 & 1534.86 & 144.26\\
    RMA           & 9033.87 &  634.11 & 1307.96 &  45.65 & 2306.23 & 222.09\\
    \colorRow
    SPARC         & \textbf{10017.90}&  \textbf{476.19} & 1348.22 &  53.67 & \textbf{2528.25} & \textbf{263.58}\\
    Oracle        & 7821.42 & 1156.77 & \textbf{1710.14} & \textbf{98.98} & 2325.30 & 576.48\\
    \bottomrule
    \end{tabular}
    }
\end{table}

\begin{figure}
  \centering
  \includegraphics[width=0.96\linewidth]{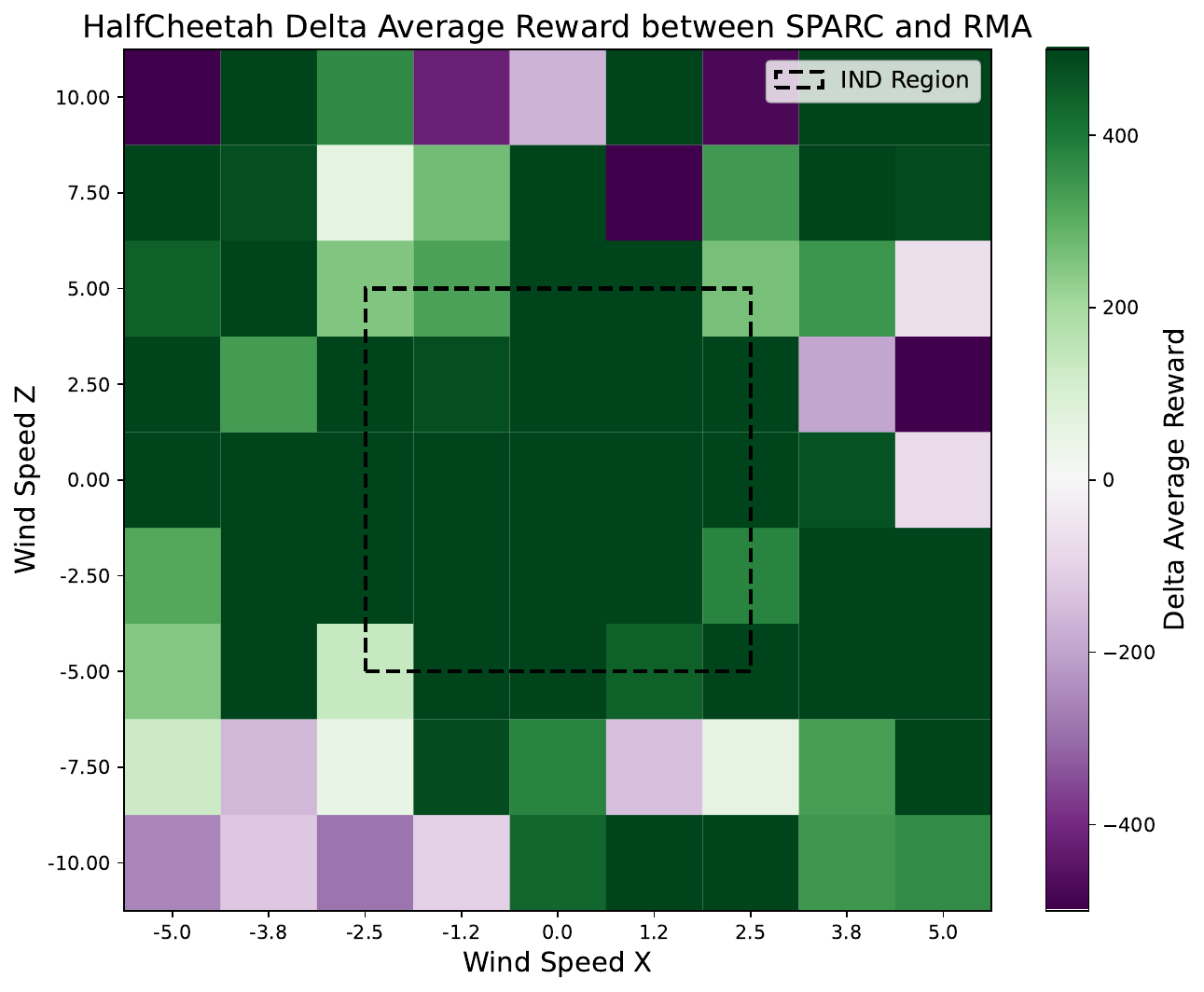}
  \caption{Difference in average return of SPARC versus RMA with varying wind perturbations over 5 seeds. In \textcolor{deltagreen}{green: SPARC is better} in that wind setting, while in \textcolor{deltapurple}{purple: RMA scores higher}. Our method outperforms the two-phase baseline across many IND and OOD contextual settings.}
  \label{fig:mujoco_delta}
\end{figure}



\subsection{MuJoCo: Wind Perturbations}
\label{sec:mujoco_results}

In \Cref{fig:mujoco_delta}, we present results on HalfCheetah by calculating the difference in performance between SPARC and its main baseline RMA, in each wind perturbation tested. The green squares show SPARC outperforming RMA, while purple indicates the opposite. Overall, SPARC beats RMA in significantly more IND and OOD settings, demonstrating a robust performance across varying contexts. 

\Cref{tab:mujoco} shows the OOD results for all baselines and MuJoCo enviroments. Again, SPARC presents strong generalization ability to unseen contexts. On Hopper, the Oracle performs best; note that this baseline has access to true context at test-time, in contrast to all others (see \Cref{tab:inputs}).

\subsection{Transferability to Updated Game Dynamics}
\label{sec:physics}

The \textit{Gran Turismo} developers regularly deploy game updates, where the simulation physics can be adjusted.
Reinforcement learning agents that are trained on previous game dynamics may struggle to adapt.
In this section we evaluate policies on the newest game dynamics, while they were solely trained on a previous version of Gran Turismo.

In \Cref{fig:across_game_dynamics}, we show that SPARC outperforms all baselines in OOD generalization, this time not only across different car models, but also across other unseen environment dynamics. The Oracle policy with access to ground-truth context is not able to finish laps with around 10\% of the OOD cars, while SPARC reduces this to less than 5\%, with significantly faster lap times.
Note that the context $c \in \mathcal{C}$ that we provide to the Oracle contains information about the car model only, as the exact simulator physics adjustments are unknown to us. This missing information highlights the importance of SPARC's ability to adapt to unseen contexts without access to comprehensive contextual details, e.g., when training in simulation and transferring to a real-world environment.

\section{Conclusion}

This paper introduces SPARC, a novel \textbf{s}ingle-\textbf{p}hase \textbf{a}daptation method for \textbf{r}obust \textbf{c}ontrol in contextual environments. The algorithm unifies context encoding and history-based adaptation into one streamlined training procedure. By eliminating the need for separate phases, SPARC not only simplifies implementation but also facilitates continual learning and deployment in real-world scenarios.

Our extensive experiments in both the high-fidelity Gran Turismo 7 simulator and various MuJoCo tasks demonstrate that SPARC achieves competitive or superior performance in both in-distribution and OOD settings. In particular, SPARC excels at generalizing to unseen contexts while maintaining robust control, a critical capability for robotics applications where explicit contexts are unknown during deployment.

While our results are promising, the work also highlights opportunities for future research. In particular, testing SPARC on physical robotic platforms and further optimizing its training efficiency remain important next steps.
Overall, SPARC represents a significant advance toward practical, adaptable agents that can thrive in dynamic environments.

\section*{Acknowledgments}

We are grateful to Florian Fuchs for initiating this research project. Further appreciation to everyone at Sony AI---especially the \textit{Gran Turismo} team members---for many fruitful discussions.
Finally, we thank all anonymous reviewers who helped to improve the quality of our work.

\section*{Ethics Statement}

This work advances core machine learning capabilities by improving the performance and generalizability of reinforcement learning agents. While we focus on algorithmic improvements, we acknowledge that, like most technical advances in ML, this work may have various societal impacts. We encourage thoughtful consideration of these implications when building upon this research.

\balance
\bibliography{aaai2026}

\setcounter{secnumdepth}{2}      
\renewcommand\thesection{\Alph{section}}   
\renewcommand\thesubsection{\thesection.\arabic{subsection}} 

\newpage
\onecolumn
\appendix
\section*{\huge Appendix}
\

\section{Training and Implementation Details}
\label{sec:details}

All experiments are conducted using the off-policy QR-SAC algorithm~\cite{wurman2022outracing} as the base reinforcement learning method.
In the \textit{MuJoCo} experiments, we train for $3M$ policy updates. 
For Gran Turismo, in the \textit{Power \& Mass} setting we perform $6M$ updates, 
while across \textit{Car Models} we train for $9M$ steps. 
The famously long and difficult \emph{N\"{u}rburgring} track is an exception, where we perform additional updates: $12M$, 
see \Cref{tab:train_steps}. 

We tune the history length (see \Cref{sec:history_length}) and the history adapter's learning rate (values: $\{2.5 \times 10^{-4}, 2.5 \times 10^{-5}$\}). The list of hyperparameters used for training our Gran Turismo and MuJoCo experiments is presented in \Cref{table:hyperparameters}.

In each training episode, a new IND setting is sampled according to Equation~\ref{eq:pc}, determining either the wind speeds, the car's power \& mass, or the full car model. Generalization complexity is further raised for the \textit{Car Model} experiment by sampling uniformly over 9 tire types, from least traction \textit{Comfort Hard} up to most traction \textit{Racing Soft}. 

\subsection{Code}

The code for \textit{Gran Turismo 7} is proprietary. The game runs exclusively on PlayStation (PS4 and PS5), which we use for training. However, for reproducibility we do provide open-source code for the multiple wind-perturbed MuJoCo environments that we designed, in the supplementary material. Furthermore, see \Cref{alg:sparc} for pseudocode of our algorithm SPARC. The resulting adapter policy $\pi^{ad}_\theta(o,h)$ no longer requires the context $c$ as input, and is the model deployed at test time.

\begin{table}[H]
    \centering
    \renewcommand{\arraystretch}{1.2}
    \caption{Number of training steps for each environment.}
    \label{tab:train_steps}
    \begin{tabular}{ll}
        \toprule
        \textbf{Environment} & \textbf{Training steps} \\
        \midrule
        MuJoCo & $3M$ \\
        Gran Turismo (Power \& Mass) & $6M$ \\
        Gran Turismo (Car Models) & $9M$ (N\"{u}rburgring track: $12M$) \\
        \bottomrule
    \end{tabular}
\end{table}

\begin{table}[b]
\centering
\renewcommand{\arraystretch}{1.2}
\caption{Training hyperparameters for Gran Turismo and MuJoCo experiments.}
\label{table:hyperparameters}
\begin{tabular}{lcc}
\toprule
\textbf{Hyperparameter} & \textbf{Gran Turismo} & \textbf{MuJoCo} \\
\midrule
Optimizer & Adam~\citep{kingma2015adam} & Adam \\
Batch size ($B$)  & 1024 & 32 \\
History length ($H$) & 50 & 50 \\
History adapter learning rate ($\alpha_{ad}$) & $2.5 \times 10^{-4}$ & $3 \times 10^{-4}$ \\
SAC learning rate for $\pi^{ex}$ and critics ($\alpha_{\text{SAC}}$) & $2.5 \times 10^{-5}$ & $3 \times 10^{-4}$ \\
Target critics update rate ($\tau$)  & 0.005  & 0.005 \\
Global norm of critics gradient clipping & 10 & 10 \\
Discount factor ($\gamma$) & 0.9896 & 0.99 \\
SAC entropy temperature~\citep{sac} & 0.01 & 0.01 \\
Number of quantiles & 32 & 32 \\
\bottomrule
\end{tabular}
\end{table}

\subsection{Model Architecture}

The architecture of SPARC used in our experiments is detailed in \Cref{tab:architecture}.
We use QR-SAC as the base reinforcement learning algorithm for the expert policy, which involves two critic networks and two critic target networks. All four of these have the same architecture, with access to the true context $c$. This is possible since the critics are not needed at deployment, only during training. We use 32 quantiles within the critic networks of QR-SAC.

The observation $o$ is a large vector of relevant information, similar to the one used by \citet{wurman2022outracing}. 
The action $a$ has a dimension of 2 for Gran Turismo, and varying numbers for the MuJoCo environments.
For our MuJoCo experiments, we decrease the width of the FC layers from 2048 and 64 to 256 and 32, respectively.

\newcommand{\thspace}{\hspace{1.5em}}
\begin{table}
  \centering
  \renewcommand{\arraystretch}{1.2} 
  \caption{Model architecture of SPARC. See \Cref{fig:overview_sparc} for a schematic overview.}
  \label{tab:architecture}
  \begin{tabular}{lcc}
    \toprule
    Layer Description & Input Dimensions & Output Dimensions \\ \midrule
    \rowcolor{rowHL!70} \multicolumn{3}{l}{Expert Policy $\pi^{ex}$} \\ 
    \multicolumn{3}{l}{\thspace Observation Encoder} \\ 
    \thspace \thspace FC: 2048 units, ReLU & $o$ & 2048 \\
    \thspace \thspace FC: 2048 units, ReLU & 2048 & 2048 \\
    \multicolumn{3}{l}{\thspace Context Encoder} \\ 
    \thspace \thspace FC: 64 units, ReLU & $c$ & 64 \\
    \thspace \thspace FC: 64 units, ReLU & 64 & 64 \\
    \multicolumn{3}{l}{\thspace Decision Layers} \\ 
    \thspace \thspace FC: 2048 units, ReLU & 2112 (2048 + 64) & 2048 \\
    \thspace \thspace FC: 2048 units, ReLU & 2048 & 2048 \\
    \thspace \thspace FC: 2$a$ units, Tanh & 2048 & 2$a$ \\
    \midrule
    \rowcolor{rowHL!70} \multicolumn{3}{l}{Adapter Policy $\pi^{ad}$} \\ 
    \multicolumn{3}{l}{\thspace Observation Encoder} \\ 
    \thspace \thspace FC: 2048 units, ReLU & $o$ & 2048 \\
    \thspace \thspace FC: 2048 units, ReLU & 2048 & 2048 \\
    \multicolumn{3}{l}{\thspace History Adapter} \\ 
    \thspace \thspace FC: 64 units, ReLU & $(o + a)$x50  & 64x50 \\
    \thspace \thspace Conv1D: 32 filters, kernel 8, stride 4, ReLU & 64x50 & 32x13 \\
    \thspace \thspace Conv1D: 32 filters, kernel 5, stride 1, ReLU &   32x13 & 32x13  \\
    \thspace \thspace Conv1D: 32 filters, kernel 5, stride 1, ReLU &   32x13 & 32x13  \\
    \thspace \thspace Flatten & 32x13  & 416 \\
    \thspace \thspace FC: 64 units, ReLU & 416  & 64 \\
    \multicolumn{3}{l}{\thspace Decision Layers} \\ 
    \thspace \thspace FC: 2048 units, ReLU & 2112 (2048 + 64) & 2048 \\
    \thspace \thspace FC: 2048 units, ReLU & 2048 & 2048 \\
    \thspace \thspace FC: 2$a$ units, Tanh & 2048 & 2$a$ \\
    \midrule
    \rowcolor{rowHL!70} \multicolumn{3}{l}{Critic Networks} \\ 
    \multicolumn{3}{l}{\thspace Observation + Action Encoder} \\ 
    \thspace \thspace FC: 2048 units, ReLU & $o + a$ & 2048 \\
    \thspace \thspace FC: 2048 units, ReLU & 2048 & 2048 \\
    \multicolumn{3}{l}{\thspace Context Encoder} \\ 
    \thspace \thspace FC: 64 units, ReLU & $c$ & 64 \\
    \thspace \thspace FC: 64 units, ReLU & 64 & 64 \\
    \multicolumn{3}{l}{\thspace Decision Layers} \\ 
    \thspace \thspace FC: 2048 units, ReLU & 2112 (2048 + 64) & 2048 \\
    \thspace \thspace FC: 2048 units, ReLU & 2048 & 2048 \\
    \thspace \thspace FC: 32 units, identity & 2048 & 32 \\
    \bottomrule
  \end{tabular}
\end{table}

  \begin{center}               
    \begin{minipage}{0.86\linewidth} 
\begin{algorithm}[H]
\caption{SPARC: \textbf{S}ingle‑\textbf{P}hase \textbf{A}daptation for \textbf{R}obust \textbf{C}ontrol}
\label{alg:sparc}
\setlength{\algomargin}{100pt}   
\DontPrintSemicolon
\SetKwInOut{Input}{Input}\SetKwInOut{Output}{Output}
\Input{%
  Environment $\mathcal{M}$ with latent context $c$
  (observable \emph{only during training})\\
  History length $H$, replay buffer $\mathcal{D}$, batch size $B$\\
  Discount $\gamma$, target‑update rate $\tau$, learning rates $\alpha_{\text{SAC}},\alpha_{ad}$
}
\Output{Adapter policy $\pi^{ad}_{\theta}$}
\textbf{Initialise:}\\
\hspace{0.5em}Expert policy $\pi^{ex}$ with context encoder $\psi$ \\
\hspace{0.5em}Adapter policy $\pi^{ad}$ with history adapter $\phi$ \\
\hspace{0.5em}Critic networks $Q^{ex}$ and target critics $\tilde Q^{ex}\!\leftarrow Q^{ex}$ \\
\hspace{0.5em}History $h \gets \mathbf{0}^{H}$ \\
\hspace{0.5em}Sample IND context $c$ for environment $\mathcal{M}$ \\
\BlankLine
\For{$t=1,2,\dots$}{

  \BlankLine
  \tcp{Collect experience with adapter policy}
  Observe $o_t$ from environment $\mathcal{M}_c$\; 
  Sample action $a_t \sim \pi^{ad}_\theta(o_t,h_t)$ \;
  Concatenate $(o_t, a_t)$ to history $h_t=(o_{t-H:t-1},a_{t-H:t-1})$; (and pop last entry) \;
  Execute $a_t$, receive reward $r_t$, next observation $o_{t+1}$ and done flag $d_t$ \;
  Store transition $(o_t,a_t,r_t,o_{t+1},d_t,h_t,c) \;\rightarrow\; \mathcal{D}$ \;

  \BlankLine
  \If{$|\mathcal{D}|\ge B$}{
    Sample minibatch $\mathcal{B}$ of $B$ tuples from $\mathcal{D}$ \;

    \tcp{Expert Policy update via QR-SAC}
    $\theta^{Q}\!\leftarrow\!
      \theta^{Q}-\alpha_{\text{SAC}}\nabla_{\theta^{Q}}\!\mathcal{L}_{\text{SAC}}
      \bigl(\pi^{ex},Q^{ex},\tilde Q^{ex}\bigr)$\;
    $\theta^{ex}\!\leftarrow\!
      \theta^{ex}+\alpha_{\text{SAC}}\nabla_{\theta^{ex}}\!J_{\text{SAC}}(\pi^{ex})$\;

    \tcp{History Adapter regression}
    $\theta^{\phi}\!\leftarrow\!
      \theta^{\phi}-\alpha_{ad}
      \nabla_{\theta^{\phi}}\bigl\|\,\phi(h)-\psi(c)\bigr\|_2^{2}$\;

    \tcp{Polyak averaging of target critics}
    $\tilde Q^{ex}\leftarrow(1-\tau)\tilde Q^{ex}+\tau Q^{ex}$\;
    \tcp{Copy obs. encoder \& decision layers from $\pi^{ex}$ to $\pi^{ad}$}
    $\theta^{ad}\leftarrow\text{copy\_modules}(\theta^{ex},\theta^{ad})$\;

  }

  \BlankLine
  \If{$d_t$}{
    reset environment $\mathcal{M}$; sample new IND context $c$ \;
    reset history $h \gets \mathbf{0}^{H}$ \;
  }
}
\end{algorithm}
    \end{minipage}
  \end{center}

\subsection{Rollouts and Distributed Training}

We train our algorithms asynchronously, performing rollouts on up to 20 distributed PlayStations simultaneously for Gran Turismo runs. The collected experience is added to a central replay buffer, from which a single GPU trains the neural networks. We use NVIDIA A100 GPUs with 40GB of memory. The wall-clock training time of an average Gran Turismo run is about 6 days, for MuJoCo it is approximately 14 hours. See \citet{wurman2022outracing} for further details on the Gran Turismo training setup.

At the start of each episode, we sample a contextual setting uniformly at random from the in-distribution (IND) set. For Gran Turismo, the agent's initial position is randomly chosen from on-track regions as well as from off-track zones to the left and right, extending up to 5\% of the track width. The agent is oriented to face a point 30 meters ahead along the center line, and its initial speed is drawn uniformly from 0 to 104.607 km/h. Each episode runs for up to 150 seconds before being reset.

\vspace{10em}
\subsection{Race Cars}

\begin{figure}[t]
  \centering
  \begin{subfigure}[b]{0.48\linewidth}
    \centering
    \includegraphics[width=\linewidth]{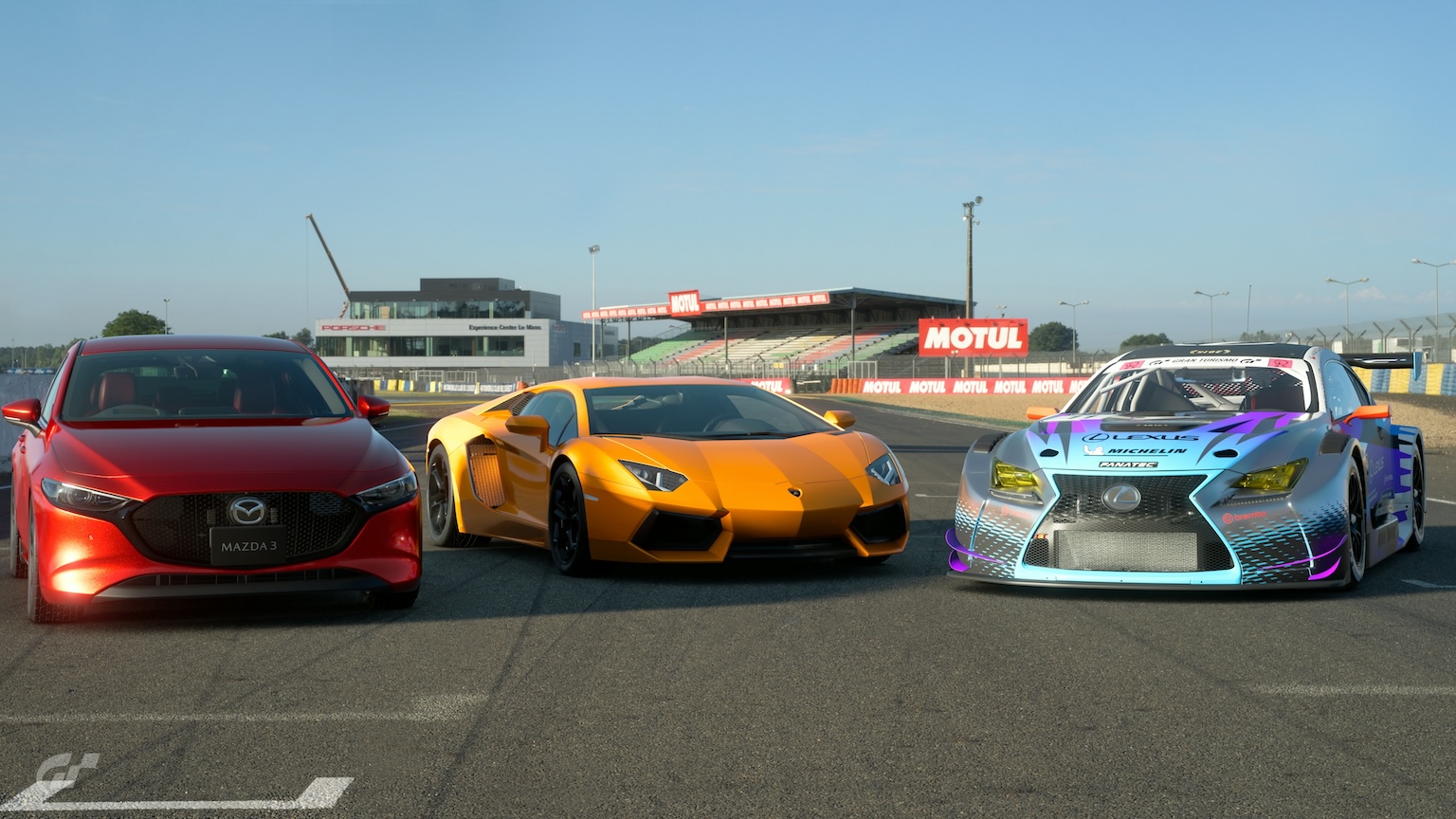}
    \caption{IND cars sampled from the $\sim$400‑vehicle training set.  
    The Mazda, Lamborghini, and Lexus (from left to right) are representative, standard racing cars in \textit{Gran Turismo 7}.}
    \label{fig:ind_cars}
  \end{subfigure}
  \hfill
  \begin{subfigure}[b]{0.48\linewidth}
    \centering
    \includegraphics[width=\linewidth]{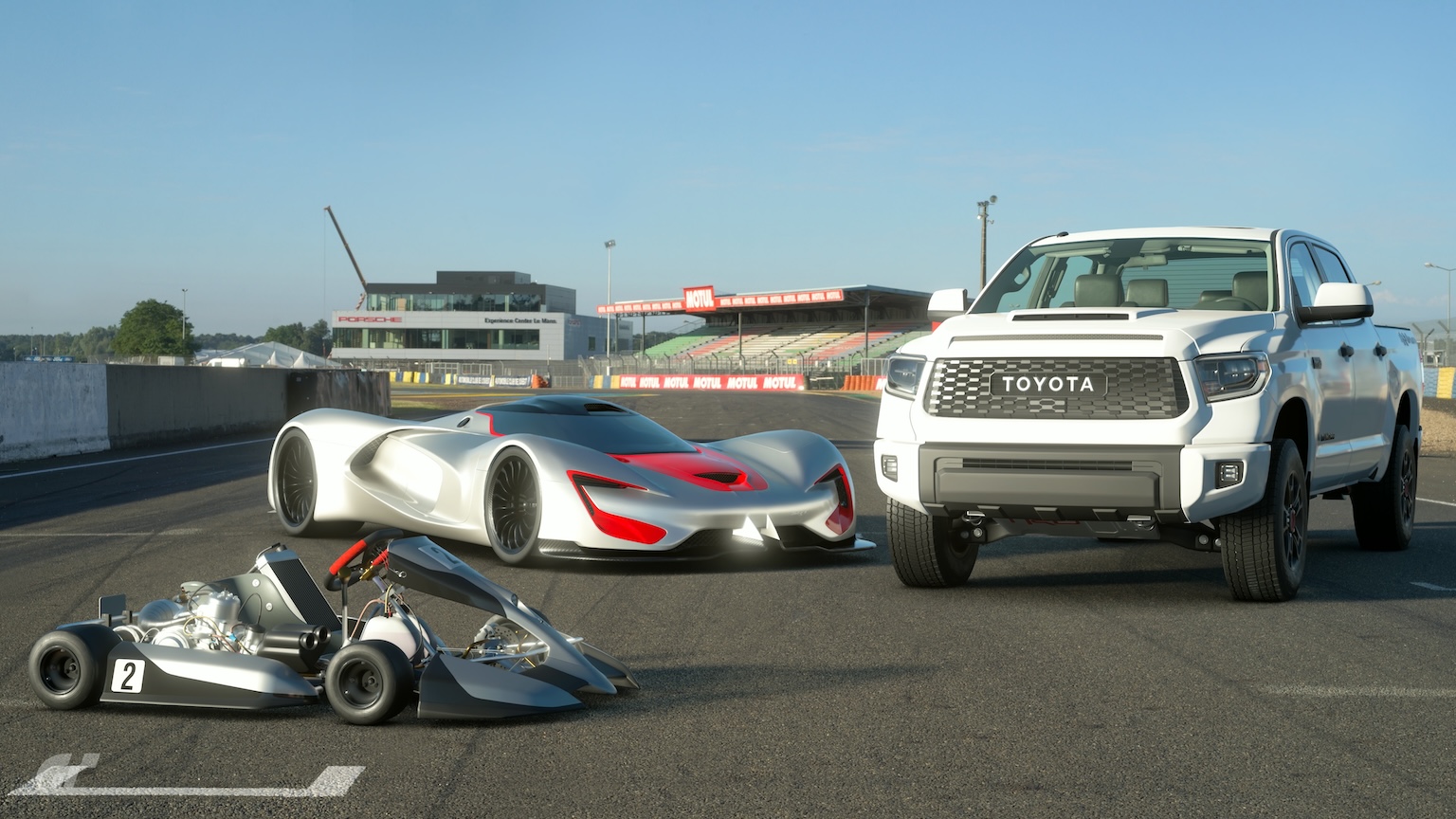}
    \caption{OOD cars drawn from the $\sim$100‑vehicle test set.  
    The racing kart, the Tomahawk, and the Toyota pickup truck (from left to right) are highly unusual outlier vehicles.}
    \label{fig:ood_cars}
  \end{subfigure}

  \caption{Illustrative in‑distribution (IND) versus out‑of‑distribution (OOD) vehicles used in our \textit{Gran~Turismo~7} experiments.}
  \label{fig:car_examples}
\end{figure}

As detailed in \Cref{sec:experimental_setup}, we hold out the 20\% most \emph{outlier} vehicles as a test set (OOD) and train on the 80\% most \emph{inlier} cars (IND). This means we train on about $\sim$400 vehicles, while testing on $\sim$100 unseen cars. In \Cref{fig:ind_cars,fig:ood_cars} we present a few examples of IND and OOD vehicles, respectively. The sampled cars in these figures are from left to right:
\

\begin{center}
\begin{tabular}{p{0.45\linewidth} p{0.38\linewidth}}
    \textbf{Example In-Distribution cars (\Cref{fig:ind_cars})} & \textbf{Example Out-of-Distribution cars (\Cref{fig:ood_cars})} \\
    \begin{itemize}
        \item Mazda3 '19
        \item Lamborghini Aventador LP 700-4 '11
        \item Lexus RC F GT3 '17
    \end{itemize} &
    \begin{itemize}
        \item GT Racing Kart 125 Shifter
        \item Dodge SRT Tomahawk X Vision GT
        \item Toyota Tundra TRD Pro '19
    \end{itemize} 
\end{tabular}
\end{center}
The \textbf{kart} (on the left in \Cref{fig:ood_cars}) is one of the most outlier vehicles: very tiny and over-responsive. The \textbf{Tomahawk} (in the middle) is extremely fast and also famously difficult to control. The Toyota \textbf{pickup truck} (on the right) is an outlier in its mass, being significantly heavier and bulkier than most IND cars.

\subsection{Race Tracks}
\label{sec:tracks}

We evaluate our approach on three diverse tracks in Gran Turismo 7: Grand Valley Highway 1, N\"{u}rburgring 24h, and Circuit de Barcelona - Catalunya Rallycross. These tracks vary widely in length, layout, and surface composition, enabling robust testing of our model's generalization capabilities.

\begin{itemize}
    \item \textbf{Grand Valley} is a medium-length road track featuring tight turns, a tunnel section, and a bridge crossing. It provides a balanced mix of technical and high-speed sections.
    \item \textbf{N\"{u}rburgring} is the longest and one of the most complex tracks in the game, combining the Grand Prix circuit with the Nordschleife. Its high difficulty and track length pose a significant generalization challenge.
    \item \textbf{Catalunya Rallycross} is the shortest of the three. It introduces mixed terrain, with both asphalt and dirt sections. Driving here requires a distinct policy due to the lower traction and rapid transitions between surface types.
\end{itemize}

\begin{figure}[H]
  \centering
  \begin{subfigure}[b]{0.3\textwidth}
    \includegraphics[width=\textwidth]{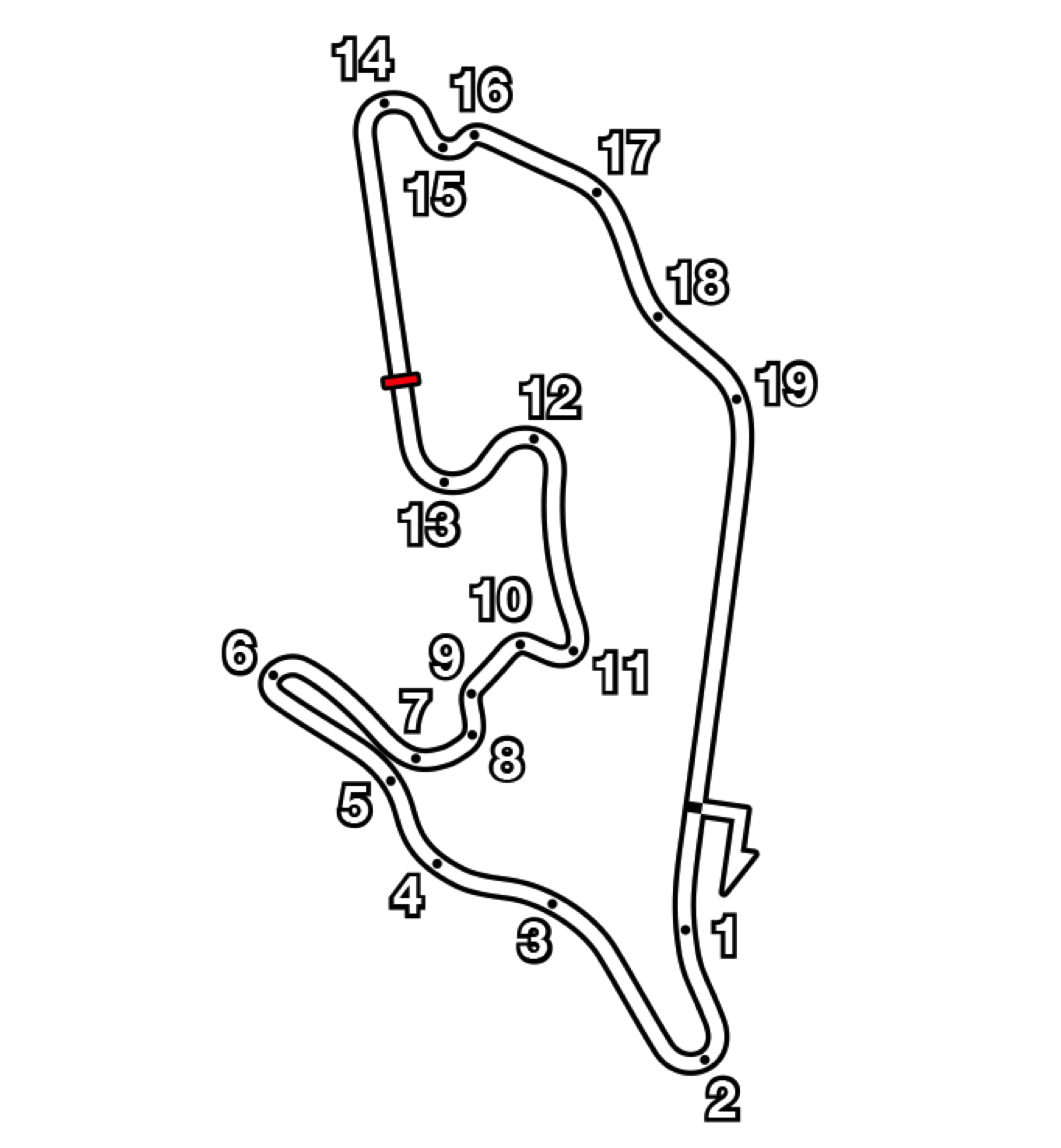}
    \caption{Grand Valley}
  \end{subfigure}
  \hfill
  \begin{subfigure}[b]{0.3\textwidth}
    \includegraphics[width=\textwidth]{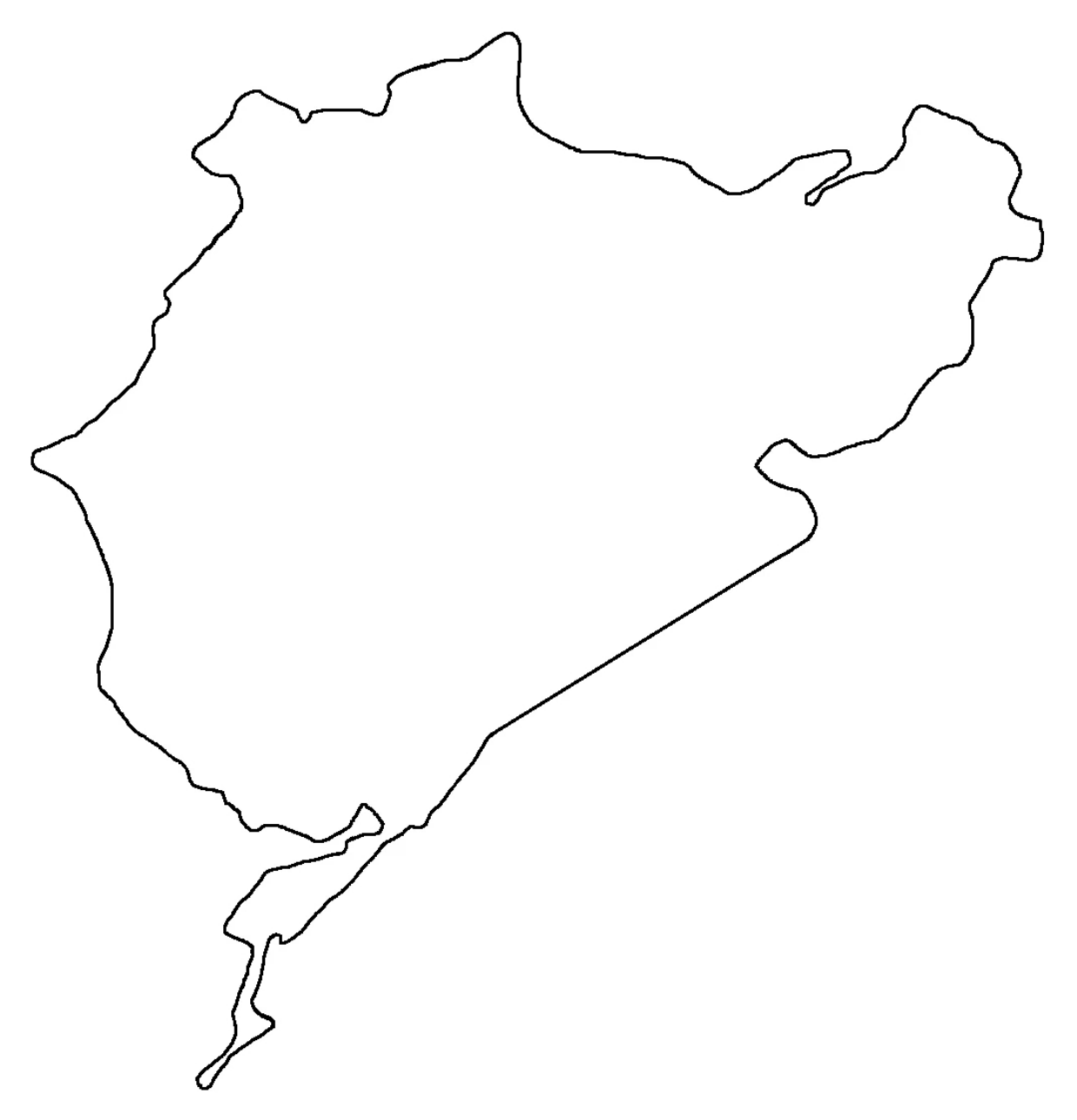}
    \caption{N\"{u}rburgring}
  \end{subfigure}
  \hfill
  \begin{subfigure}[b]{0.3\textwidth}
    \includegraphics[width=\textwidth]{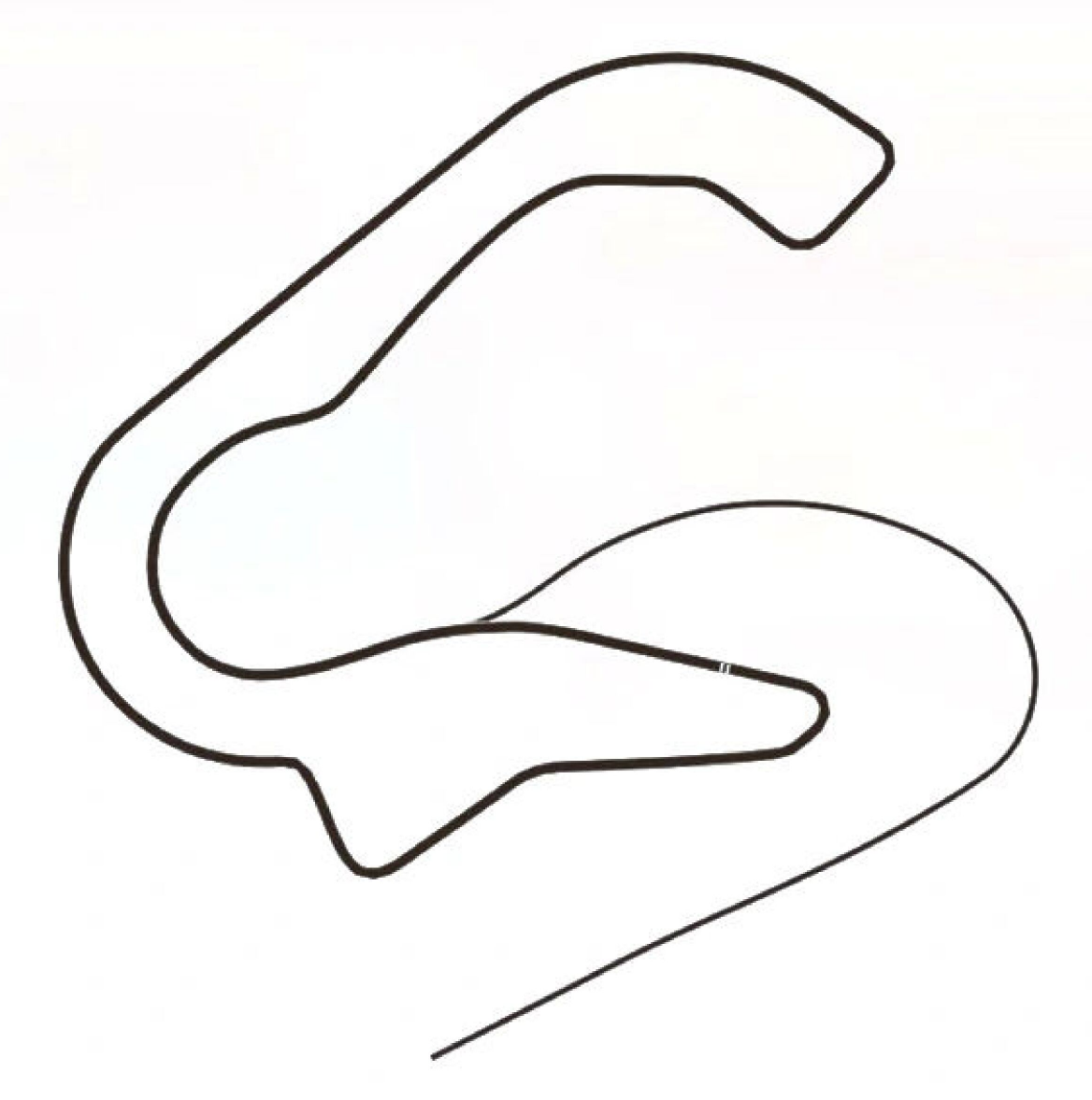}
    \caption{Catalunya Rallycross}
  \end{subfigure}
  \caption{Map layouts of the three tracks in Gran Turismo 7 used in our experiments. Note that these are not in the same scale, N\"{u}rburgring is by far the longest track. Catalunya Rallycross is the shortest and features a separate start ramp, which is not a part of our experiments. See \Cref{tab:tracks} for the exact lengths of the tracks.}
  \label{fig:track_layouts}
\end{figure}

\begin{figure}[H]
  \centering
  \begin{subfigure}[b]{0.32\textwidth}
    \includegraphics[width=\textwidth]{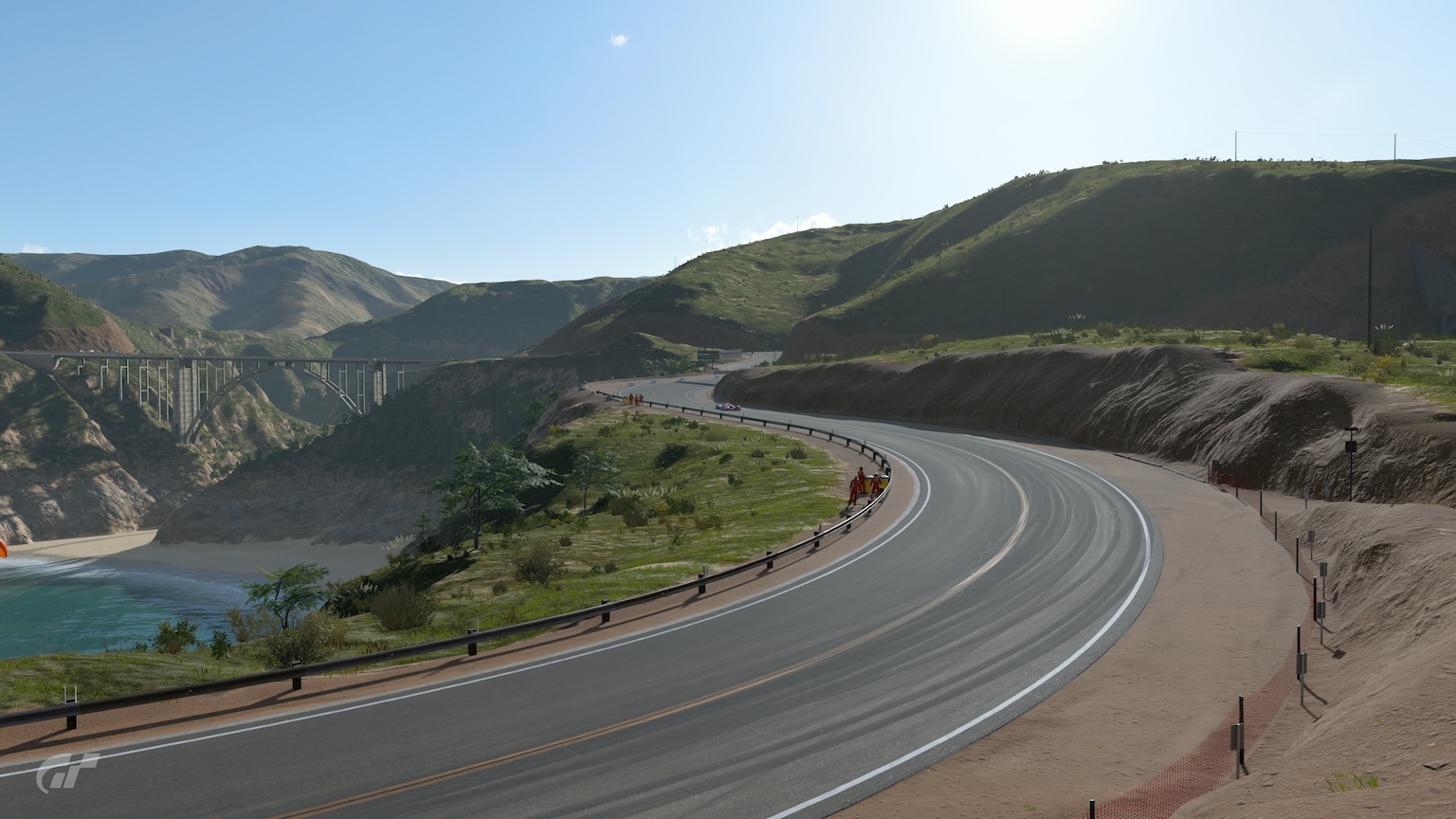}
    \caption{Grand Valley}
  \end{subfigure}
  \hfill
  \begin{subfigure}[b]{0.32\textwidth}
    \includegraphics[width=\textwidth]{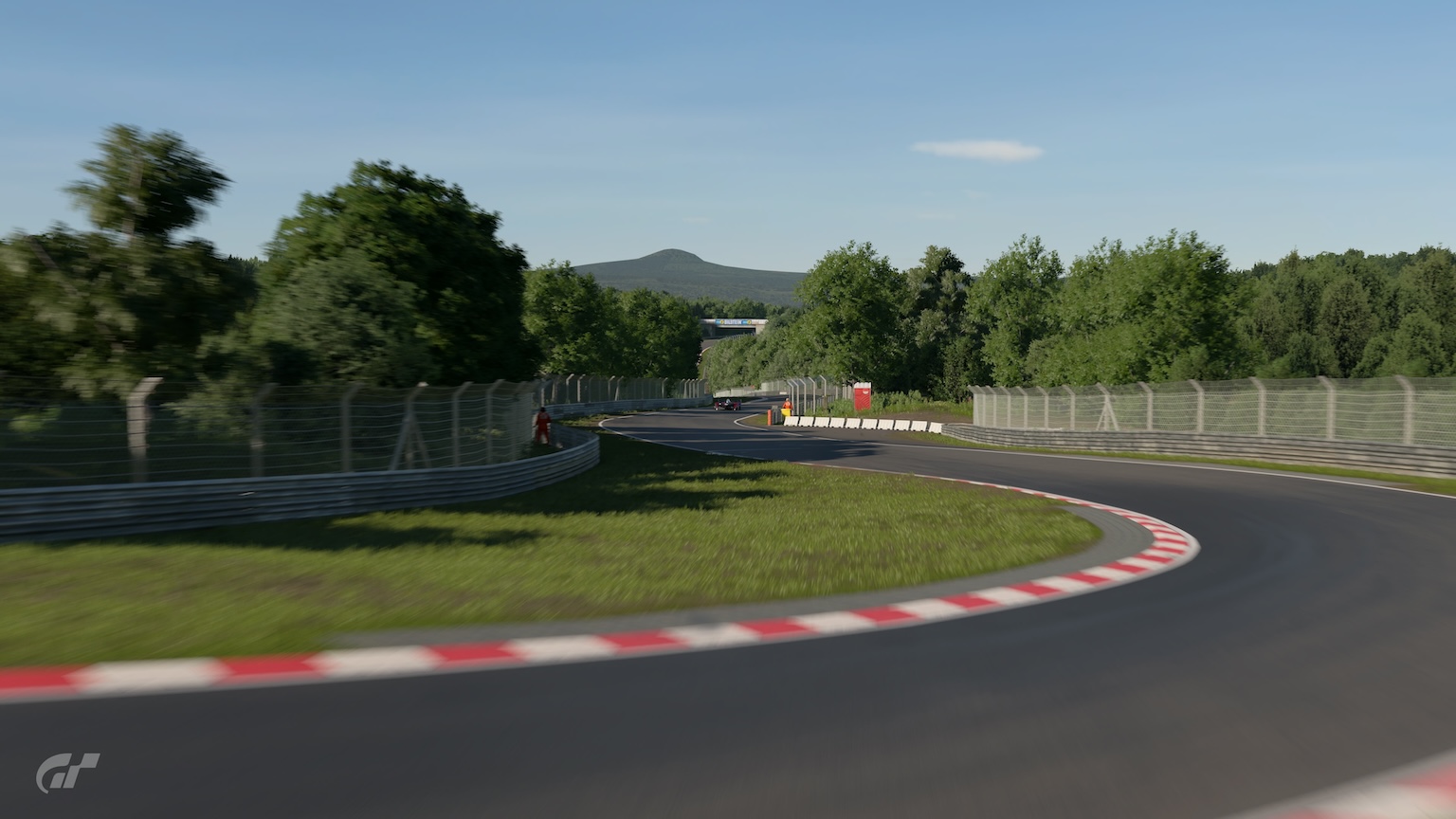}
    \caption{N\"{u}rburgring}
  \end{subfigure}
  \hfill
  \begin{subfigure}[b]{0.32\textwidth}
    \includegraphics[width=\textwidth]{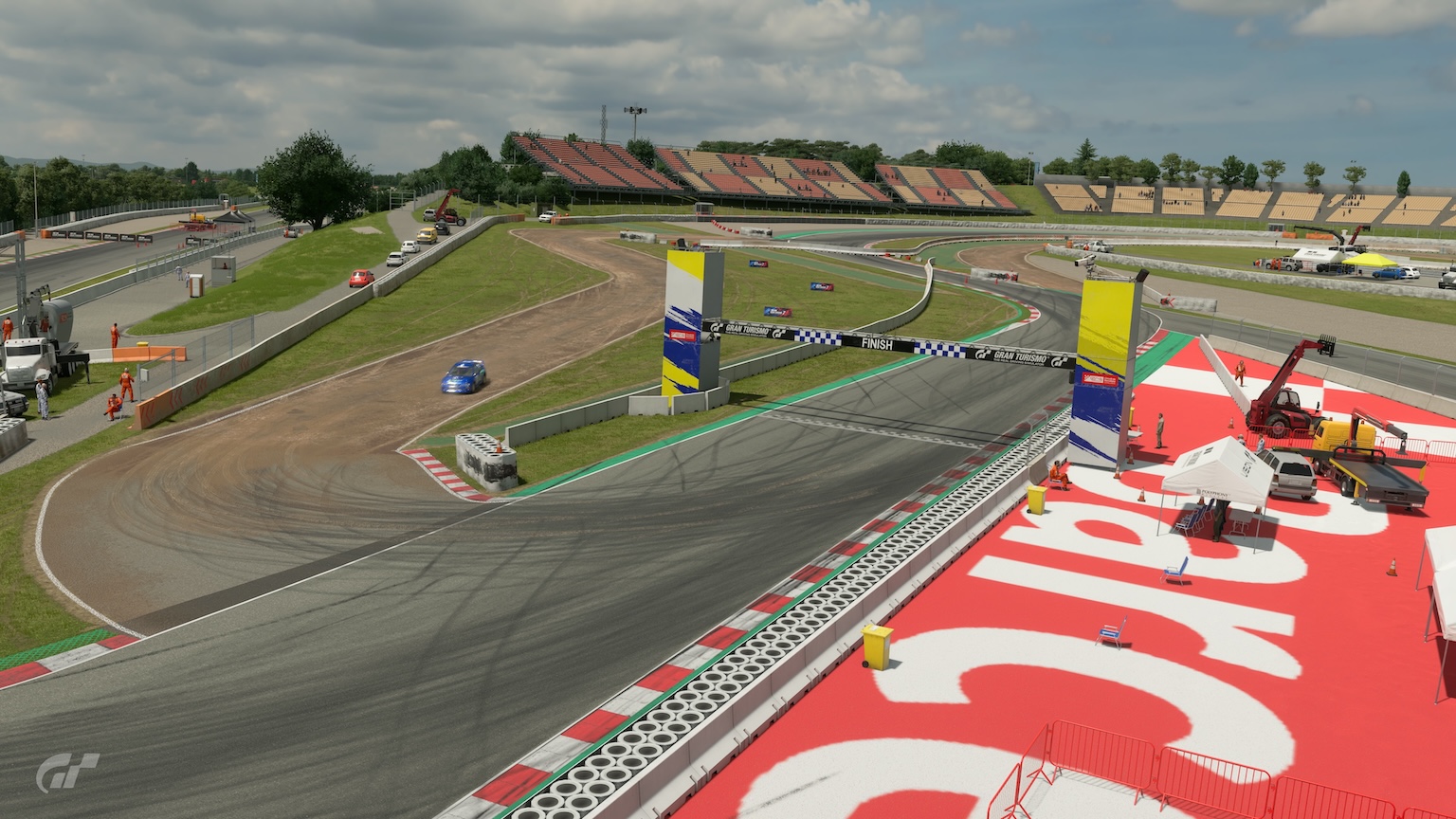}
    \caption{Catalunya Rallycross}
  \end{subfigure}
  \caption{In-game pictures of the three tracks in Gran Turismo 7 used in our experiments. The Grand Valley track is placed on a regular road, going over a bridge and a through a tunnel. N\"{u}rburgring is a replica of the real-world race track located in Germany. Catalunya Rallycross is a mixed surface track, with tarmac and dirt areas. Specific dirt tires are required for this track.}
  \label{fig:track_pics}
\end{figure}

The map layouts of these tracks are shown in \Cref{fig:track_layouts}, while in-game screenshots can be found in \Cref{fig:track_pics}. Exact track lengths and characteristics are summarized in \Cref{tab:tracks}.

\section{Further Analysis}

To gain a deeper understanding of the contributions of our design choices, we perform an analysis of key factors, including the ideal rollout policy (\Cref{sec:rollout_policy}) and the optimal history length (\Cref{sec:history_length}).

\subsection{Selecting the Ideal Rollout Policy}
\label{sec:rollout_policy}

In this ablation, we compare the performance of SPARC under two scenarios: (i) when experience is collected with the expert policy ($\pi^{ex}$), or (ii) when this is done by the adapter policy ($\pi^{ad}$). As detailed in \Cref{sec:sparc_single}, by default SPARC performs rollouts with the adapter policy to ensure a more on-policy style of learning for the final product, which is $\pi^{ad}$. 

\Cref{tab:naive_IND_OOD} presents results across all IND and OOD cars on three tracks. Although the differences are small, the results demonstrate that naively using $\pi^{ex}$ for rollouts leads to slower lap times on 2 out of 3 tracks when looking at OOD generalization.
Moreover, SPARC finishes every track with at least as many OOD cars as the alternative method.

\begin{table}[]
    \centering
    \captionsetup{width=.82\linewidth}
    \caption{Ablation on rollout policy. Results across all IND/OOD cars and 3 seeds. Mean $\pm$ s.e.m.\ of lap-time BIAI ratio and \% successful laps. Especially on OOD, naively collecting experience with the expert policy ($\pi^{ex}$) generally does not perform as well as directly using the adapter policy ($\pi^{ad}$) for rollouts.}
    \label{tab:naive_IND_OOD}
    \begin{tabular}{l@{\hspace{1em}}cc@{\hspace{1em}}cc}
        \toprule
        Method & \multicolumn{2}{c}{IND} & \multicolumn{2}{c}{OOD}\\
        \cmidrule(lr){2-3}\cmidrule(lr){4-5}
               & BIAI ratio ($\downarrow$) & Success \% ($\uparrow$)
               & BIAI ratio ($\downarrow$) & Success \% ($\uparrow$)\\
        \midrule
        \multicolumn{5}{l}{\textit{Grand Valley}}\\
        SPARC ($\pi^{ex}$ rollouts) & \textbf{0.9922 $\pm$ 0.0034} & \textbf{100.0 $\pm$ 0.00} & \textbf{1.0417 $\pm$ 0.0024} & \textbf{98.06 $\pm$ 0.00} \\
        SPARC       & 0.9999 $\pm$ 0.0061 & 99.76 $\pm$ 0.14 & 1.0491 $\pm$ 0.0055 & \textbf{98.06 $\pm$ 0.56} \\
        \addlinespace
        \multicolumn{5}{l}{\textit{Nürburgring}}\\
        SPARC ($\pi^{ex}$ rollouts) & \textbf{1.0107 $\pm$ 0.0150} & \textbf{96.93 $\pm$ 1.21} & 1.1531 $\pm$ 0.0158 & 85.44 $\pm$ 1.12\\
        SPARC       & 1.0254 $\pm$ 0.0061 & 95.87 $\pm$ 0.49 & \textbf{1.1199 $\pm$ 0.0076} & \textbf{89.00 $\pm$ 0.86} \\
        \addlinespace
        \multicolumn{5}{l}{\textit{Catalunya Rallycross}}\\
        SPARC ($\pi^{ex}$ rollouts) & 0.9434 $\pm$ 0.0012 & \textbf{100.00 $\pm$ 0.00} & 0.9659 $\pm$ 0.0032 & \textbf{100.00 $\pm$ 0.00} \\
        SPARC        & \textbf{0.9432 $\pm$ 0.0027} & \textbf{100.00 $\pm$ 0.00} & \textbf{0.9631 $\pm$ 0.0026} & \textbf{100.00 $\pm$ 0.00} \\
        \bottomrule
    \end{tabular}%
\end{table}

\subsection{Optimal History Length} 
\label{sec:history_length}

We perform a sensitivity analysis of SPARC to different lengths of the observation-action history \(H\). Recall from \Cref{fig:overview_sparc} that SPARC's history adapter $\phi$ uses this recent experience to recognize its current contextual environment. The history input is defined as $h_t=(o_{t-H:t-1},a_{t-H:t-1})\in\mathcal{H}$, where $\mathcal{H} = (\mathcal{O}\!\times\!\mathcal{A})^{H}$ is the space of possible histories.

\begin{figure}[h]
    \centering
    \includegraphics[width=0.58\linewidth]{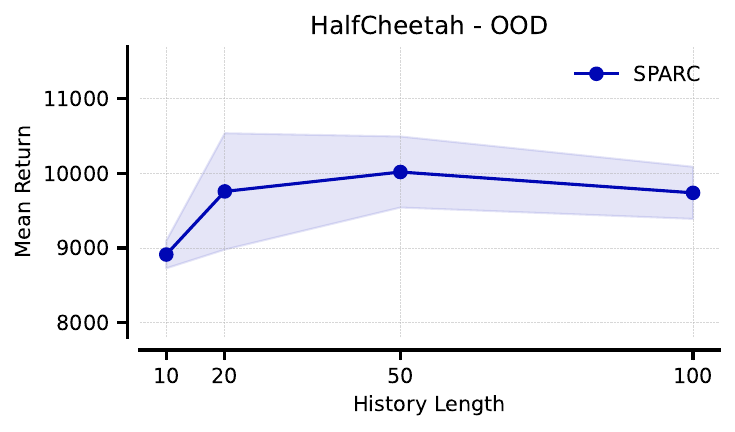}
    \captionsetup{width=.6\linewidth}
    \caption{Analysis of the optimal history length for SPARC, averaged over 5 seeds ($\pm$ s.e.m.). We show the mean return over all tested out-of-distribution wind perturbations. A history length of 50 yields the highest average return.}
    \label{fig:history_analysis}
\end{figure}

The results in \Cref{fig:history_analysis} show that a history length of $H = 50$ timesteps seems to be optimal for SPARC. Too few observation-action pairs do not provide enough information to robustly distinguish between contexts, whereas too many may distract the agent and waste computational resources.

\section{Additional Results}

In this section we provide the graphs and heatmaps of all our experiments. We start with \textit{Gran Turismo}, showing the \textit{Car Models} generalization experiments in \Cref{sec:add_results_gt_cars} and the \textit{Power \& Mass} runs in \Cref{sec:add_results_gt_bop}. The MuJoCo experiments are presented in \Cref{sec:add_results_mujoco}.

\subsection{Gran Turismo: Car Models}
\label{sec:add_results_gt_cars}

\begin{figure}[H]
  \centering
  \begin{subfigure}{0.48\textwidth}
    \includegraphics[width=\linewidth]{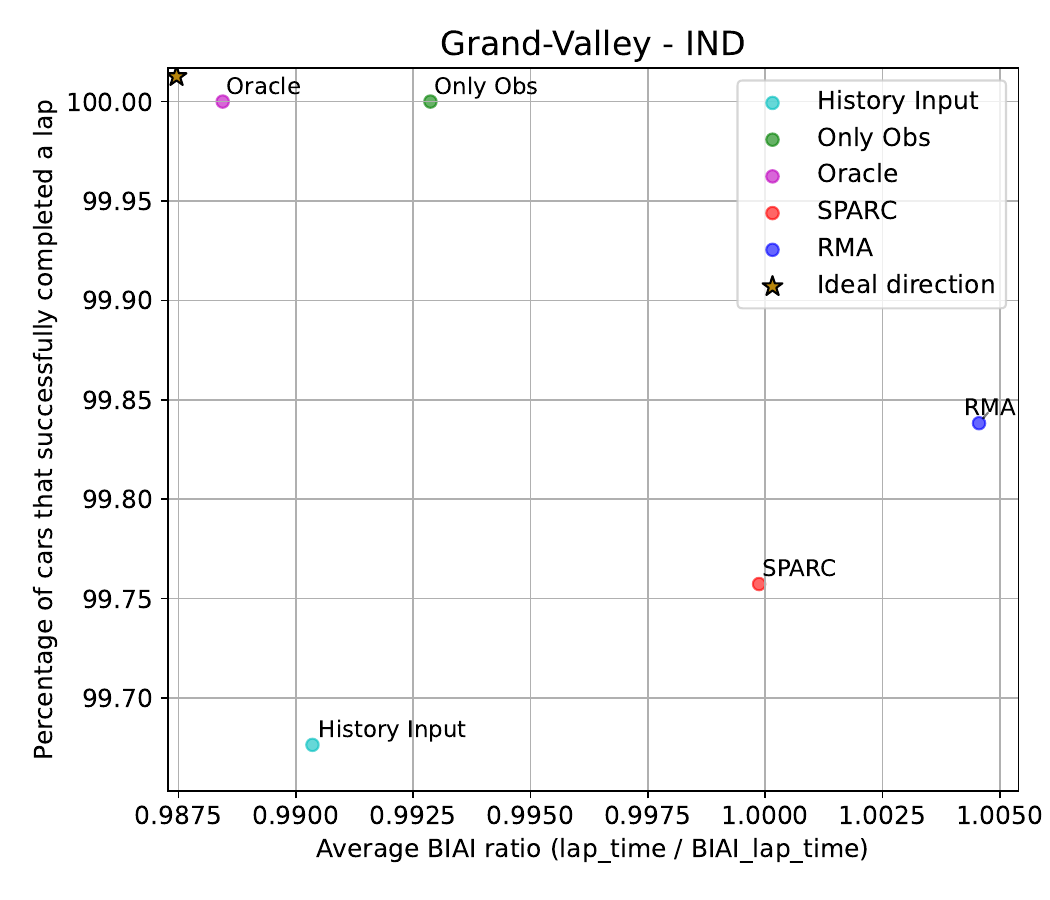}
    \caption{Grand Valley}
    \label{fig:grand_valley2}
  \end{subfigure}
  \hfill
  \begin{subfigure}{0.48\textwidth}
    \includegraphics[width=\linewidth]{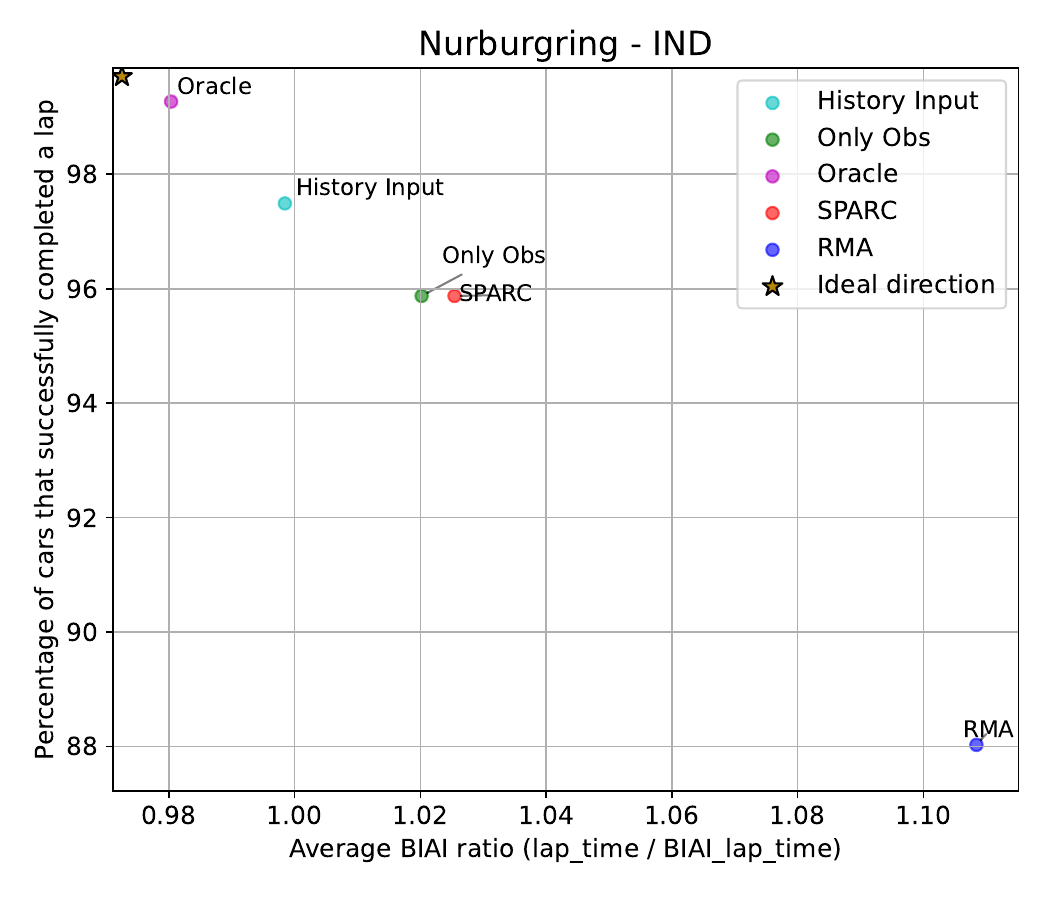}
    \caption{N\"{u}rburgring}
    \label{fig:nurburgring2}
  \end{subfigure}
    \hfill
  \begin{subfigure}{0.48\textwidth}
    \includegraphics[width=\linewidth]{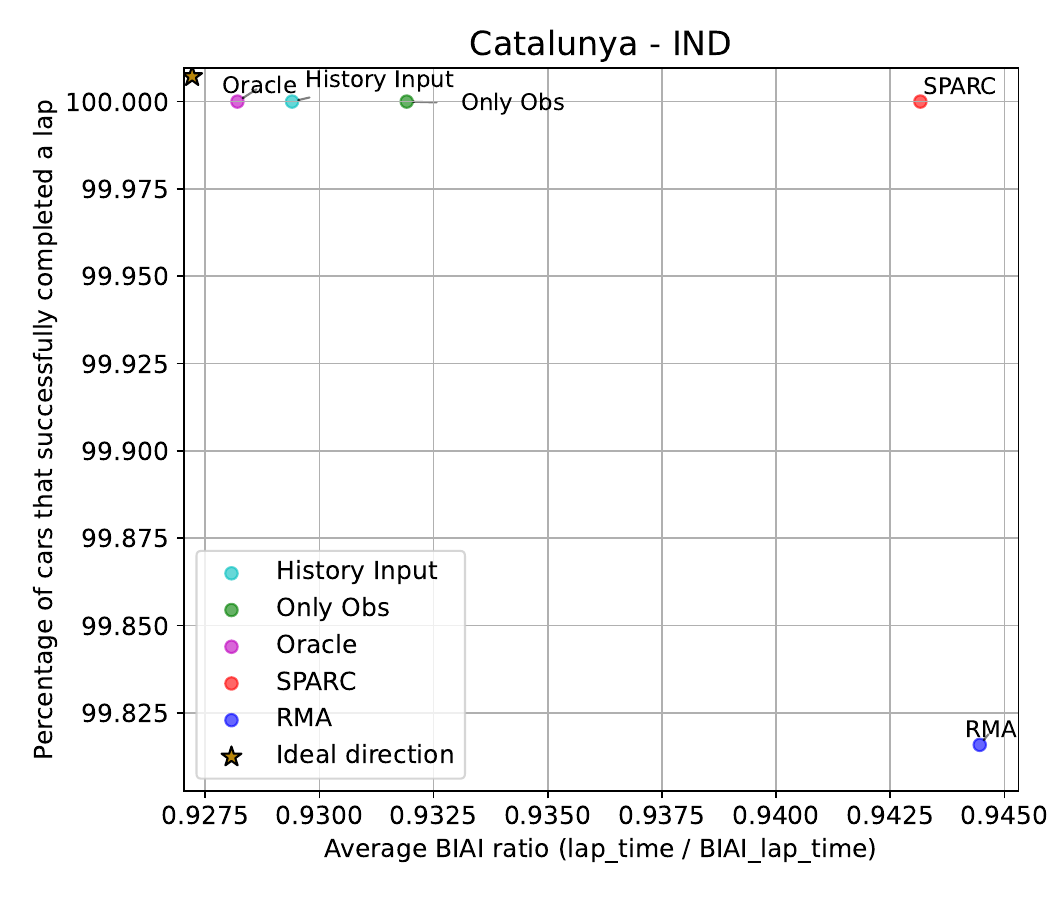}
    \caption{Catalunya Rallycross}
    \label{fig:catalunya2}
  \end{subfigure}
  \caption{
  Built-in-AI ratio lap times and the percentage of cars that are able to complete a lap for \textbf{in-distribution} (\textbf{IND}) vehicles. SPARC is designed to generalize well on OOD contexts, but also attains results relatively close to the Oracle baseline in the IND setting. Note that the numbers on these axes contain smaller intervals than the OOD plot (\Cref{fig:gt_all_ood_biai2}).}
  \label{fig:gt_all_ind_biai2}
\end{figure}

\begin{figure}[H]
  \centering
  \begin{subfigure}{0.48\textwidth}
    \includegraphics[width=\linewidth]{figures/gt/scatter_Grand-Valley_biai2_OOD.pdf}
    \caption{Grand Valley}
    \label{fig:grand_valley2}
  \end{subfigure}
  \hfill
  \begin{subfigure}{0.48\textwidth}
    \includegraphics[width=\linewidth]{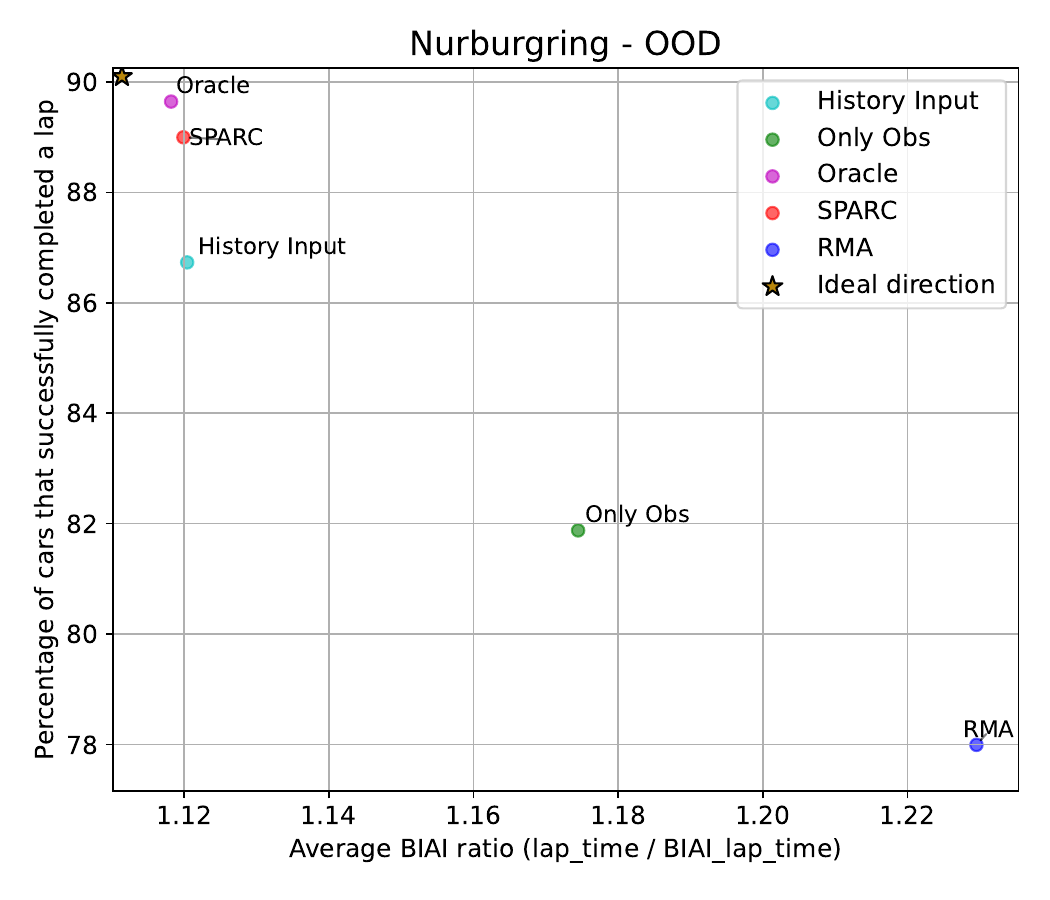}
    \caption{N\"{u}rburgring}
    \label{fig:nurburgring2}
  \end{subfigure}
    \hfill
  \begin{subfigure}{0.48\textwidth}
    \includegraphics[width=\linewidth]{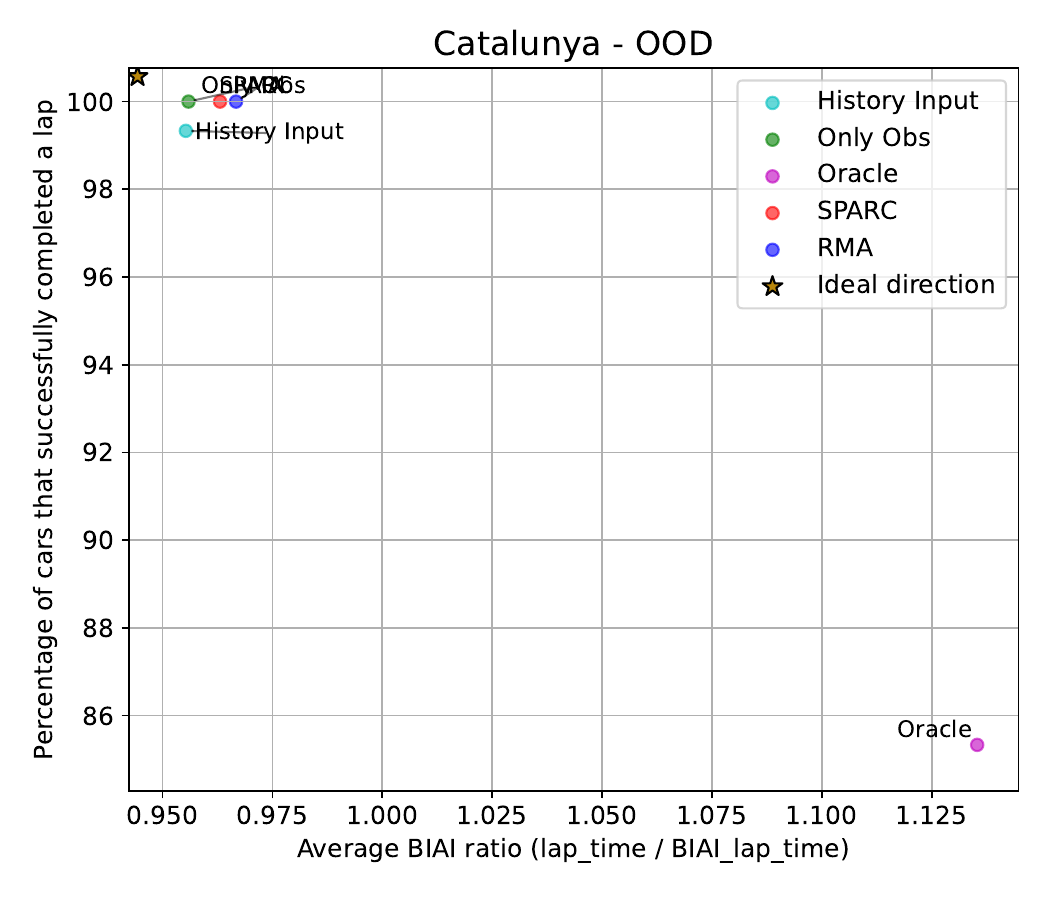}
    \caption{Catalunya Rallycross}
    \label{fig:catalunya2}
  \end{subfigure}
  \caption{
  Built-in-AI ratio lap times and the percentage of cars that are able to complete a lap for \textbf{out-of-distribution} (\textbf{OOD}) vehicles. On Grand~Valley SPARC has the fastest laps and finishes the most OOD cars. On N\"{u}rburgring, only the Oracle (which has access to priviledged context at test time) is able to surpass SPARC. The mixed dirt-and-tarmac track (Catalunya Rallycross) is much shorter in length, and thus many algorithms are able to complete laps with all cars.}
  \label{fig:gt_all_ood_biai2}
\end{figure}

Figure~\ref{fig:gt_all_ind_biai2} (IND) and Figure~\ref{fig:gt_all_ood_biai2} (OOD) show the full scatter plots of built‑in–AI (BIAI) lap–time ratio versus completion rate for every algorithm on each of the three tracks.  Each marker corresponds to the average over three independent seeds. These plots make the trade‑off between speed and reliability visible at a glance: points closer to the upper–left corner denote faster laps and a higher percentage of completed cars.

Across all tracks, SPARC is on or near the OOD Pareto front, despite lacking privileged context at test time. Note that on the IND plots, the numbers on the axes are generally much closer together than with OOD cars.
Table~\ref{tab:gt_summary_biai2} summarises these results numerically.  SPARC attains the best out‑of‑distribution BIAI ratio on two out of three tracks and, averaged over all cars and circuits, completes the most laps with OOD vehicles.

\begin{table}[]
    \centering
    \captionsetup{width=.78\linewidth}
    \caption{Performance summary on \textbf{IND} vehicles for the physics update experiment on Grand Valley. Results show the mean built-in AI (BIAI) ratio across cars. Additionally, we show the percentage of cars with a successfully completed lap.
    SPARC suffers only a modest slowdown (+0.027) and retains 98\% completion, outperforming all non‑Oracle baselines. 
    }
    \label{tab:gt_summary_physics_update_ind}
    \begin{tabular}{ll@{\hspace{1em}}cc@{\hspace{1em}}cc}
        \toprule
         \multirow{2}{*}{Method} & \multicolumn{2}{c}{Before Physics Update} & \multicolumn{2}{c}{After Physics Update}  \\
        \cmidrule(lr){2-3} \cmidrule(lr){4-5}
         & BIAI ratio ($\downarrow$) & Success (\%) ($\uparrow$) & BIAI ratio ($\downarrow$) & Success (\%) ($\uparrow$) \\
        \midrule
            Only Obs        & 0.9902  & \textbf{100.0}  & 1.1334 & 89.08  \\
            History Input    & 0.9932  & 99.27  & 1.1100   &89.32  \\
            RMA               & 1.0093  & 99.27 & 1.0584  & 95.15  \\
            \colorRow
            SPARC            & 0.9964  & \textbf{100.0}  & 1.0234 &  98.30  \\
            Oracle     & \textbf{0.9880}  & \textbf{100.0}  & \textbf{0.9981}   & \textbf{99.76}  \\
        \bottomrule
    \end{tabular}%
\end{table}

\begin{table}[H]
    \centering
    \captionsetup{width=.78\linewidth}
    \caption{Performance summary on \textbf{OOD} vehicles for the physics update experiment on Grand Valley. Results show the mean built-in AI (BIAI) ratio across cars. Additionally, we show the percentage of cars with a successfully completed lap.
    SPARC maintains the highest completion rate (97\%) and the fastest lap‑times (1.0736) among all methods.}
    \label{tab:gt_summary_physics_update_ood}
    \begin{tabular}{ll@{\hspace{1em}}cc@{\hspace{1em}}cc}
        \toprule
         \multirow{2}{*}{Method} & \multicolumn{2}{c}{Before Physics Update} & \multicolumn{2}{c}{After Physics Update}  \\
        \cmidrule(lr){2-3} \cmidrule(lr){4-5}
         & BIAI ratio ($\downarrow$) & Success (\%) ($\uparrow$) & BIAI ratio ($\downarrow$) & Success (\%) ($\uparrow$) \\
        \midrule
            Only Obs        & 1.0769  & 93.20  & 1.2912 & 74.76  \\
            History Input    & 1.0633  & 94.17  & 1.2926   &72.82  \\
            RMA               & 1.0729  & 95.15 & 1.2089  & 81.55  \\
            \colorRow
            SPARC            & \textbf{1.0451 } & \textbf{98.06}  & \textbf{1.0736} & \textbf{97.09}  \\
            Oracle     & 1.1016  & 93.20  & 1.1271   & 91.26  \\
        \bottomrule
    \end{tabular}%
\end{table}

\begin{figure}[H]
    \centering
    \includegraphics[width=0.6\linewidth]{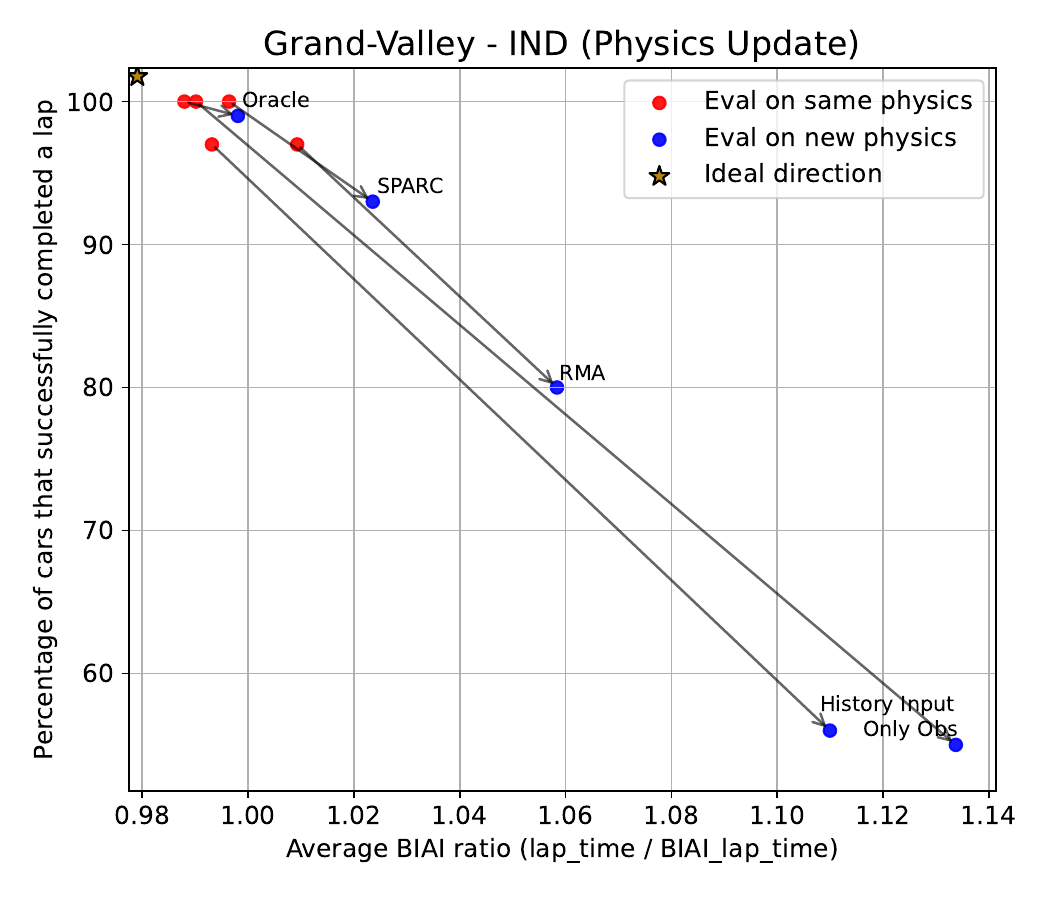}
    \caption{Performance difference between old and new game dynamics. These algorithms have only been trained on old physics, and are tested zero-shot for generalization to the new environment. This graph is on the set of IND cars, but the new physics is out-of-distribution. SPARC and the Oracle baseline have the lowest drops in performance.
    See \Cref{fig:across_game_dynamics} for the plot with OOD cars. }
    \label{fig:across_game_dynamics_ind}
\end{figure}

\paragraph{Robustness to a physics patch.}
The March‑2025 update to \textit{Gran~Turismo~7} subtly altered the vehicle‑dynamics model.  
To measure zero‑shot transfer we evaluate agents trained on the \emph{old} physics on Grand~Valley \emph{before} and \emph{after} the patch.  
\Cref{sec:physics} showed a plot with OOD dynamics \textit{and} cars, but in \Cref{fig:across_game_dynamics_ind} we evaluate on the in‑distribution cars seen during training. Although the vehicles themselves are familiar, the simulator’s updated physics offer an unseen domain shift, so each agent must generalise zero‑shot to new dynamics. As the plot shows, SPARC preserves both speed and reliability under this shift, while the other baselines without access to contextual information suffer noticeable drops in completion rate and/or lap‑time. This highlights that SPARC’s single‑phase adaptation is not merely memorising vehicle‑specific behaviour; it learns transferrable representations that remain robust when the underlying environment changes.

\Cref{tab:gt_summary_physics_update_ind} reports results on in‑distribution (IND) cars, while \Cref{tab:gt_summary_physics_update_ood} repeats the test on out‑of‑distribution (OOD) cars.  
Each table lists the mean built‑in–AI (BIAI) lap‑time ratio ($\downarrow$~lower~is~better) and the percentage of cars that complete a lap ($\uparrow$).  
Across both IND and OOD car sets, SPARC shows the smallest increase in lap‑time ratio and the least drop in completion rate among methods that do \emph{not} receive privileged context at test time, demonstrating strong resilience to unforeseen dynamics changes.

\subsection{Gran Turismo: Power \& Mass}
\label{sec:add_results_gt_bop}

For a controlled stress‑test, in \Cref{sec:bop_results} we picked a single baseline car and randomized its \emph{engine power} (x‑axis) and \emph{vehicle mass} (y‑axis) independently in every episode. In this section, we present additional results from that experiment.
The dashed box in the middle of each plot marks the in‑distribution training region \([75\%,125\%]\times[75\%,125\%]\); everything outside is held‑out OOD. 
Black squares denote settings where at least one evaluation seed failed to finish a lap.

\paragraph{IND results.}
\Cref{tab:bop_ind} reports averages \textit{inside} the dashed box, as opposed to the OOD results from \Cref{tab:bop_ood}. 
All methods finish virtually every lap, and SPARC matches the two‑phase RMA and the privileged Oracle on both speed (BIAI $\approx$ 0.981) and reliability (100\% completion).

\begin{figure}[H]
  \centering
  \begin{subfigure}{0.48\textwidth}
    \includegraphics[width=\linewidth]{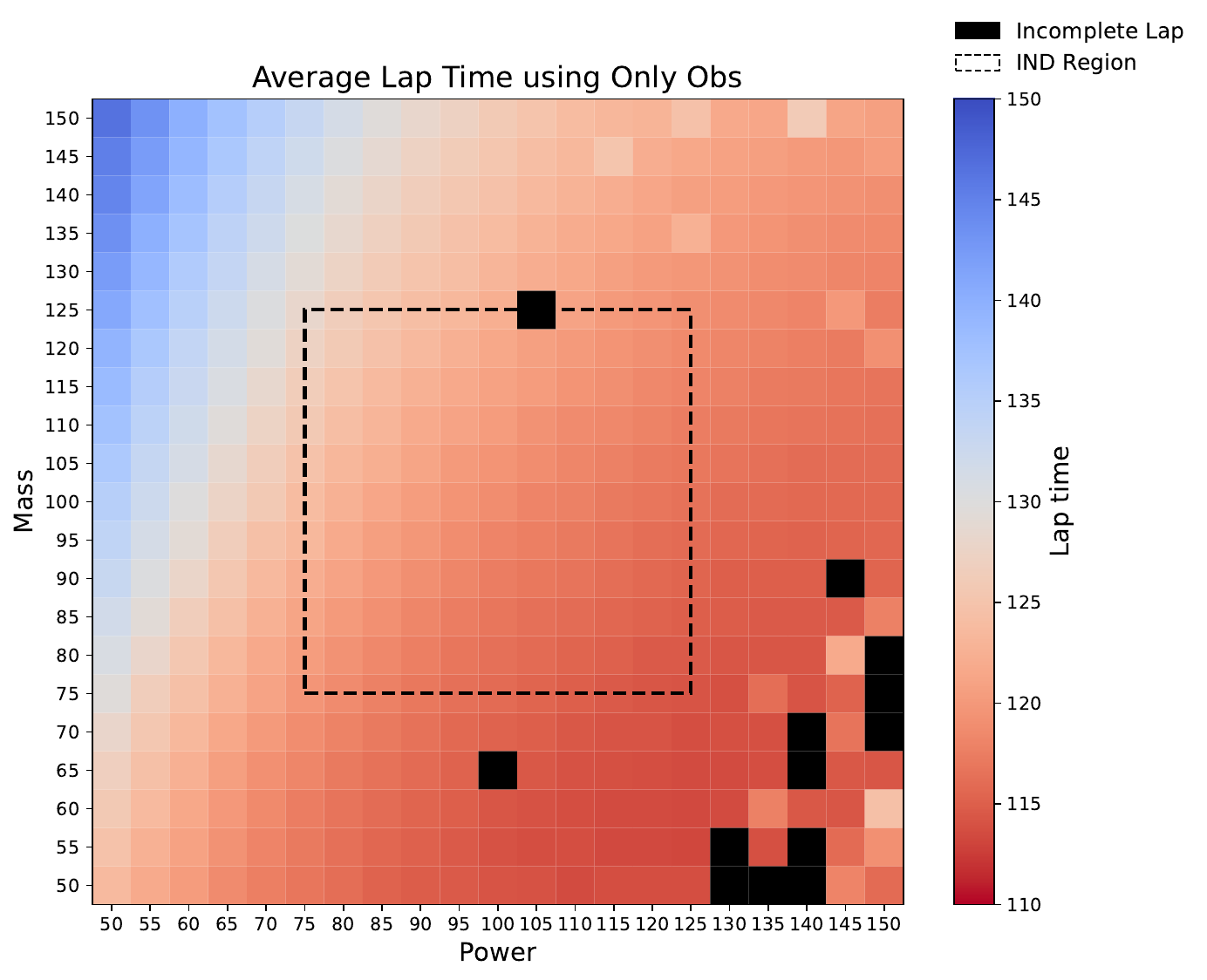}
    \caption{Only Obs}
    \label{fig:bop_only_obs}
  \end{subfigure}
  \hfill
  \begin{subfigure}{0.48\textwidth}
    \includegraphics[width=\linewidth]{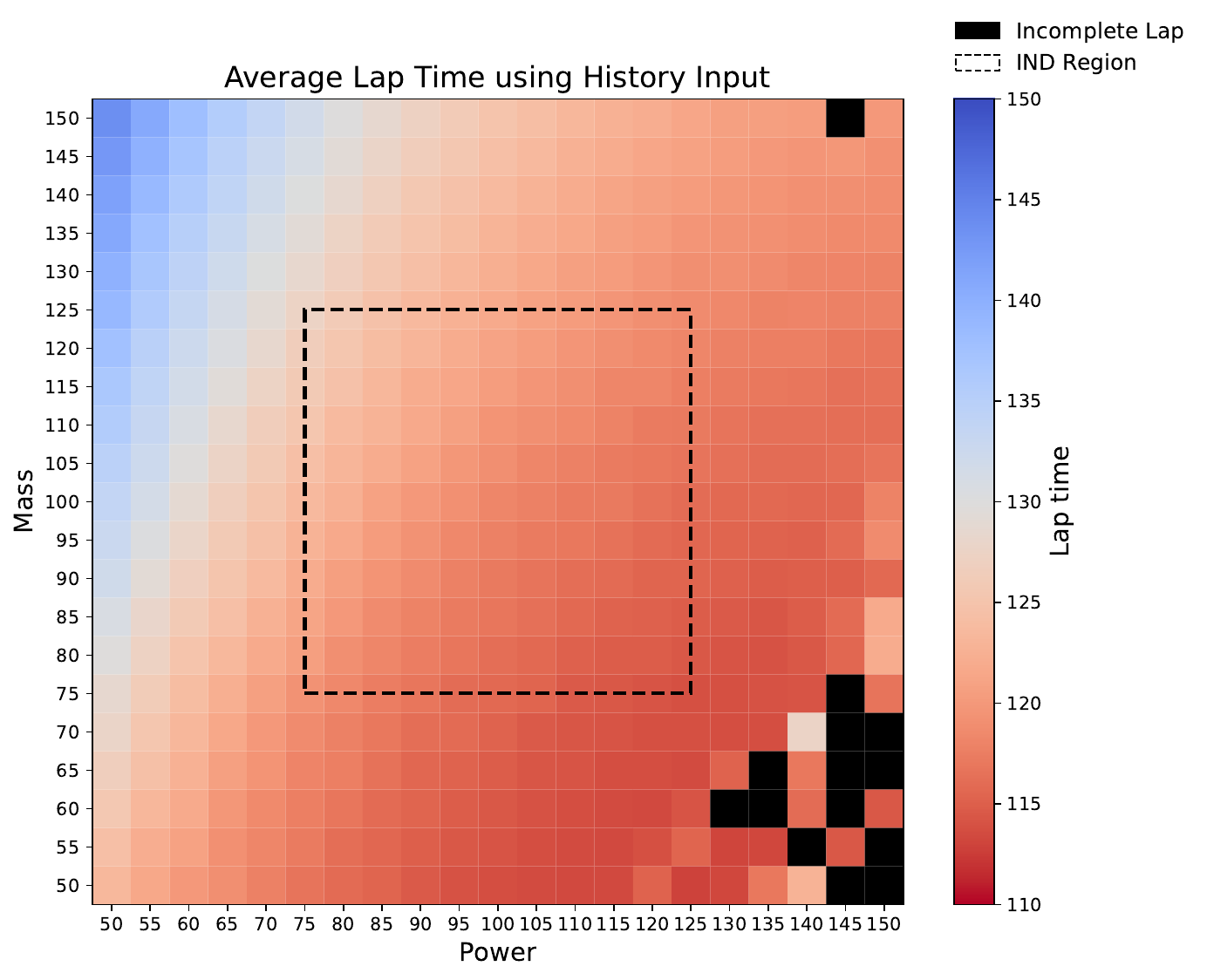}
    \caption{History Input}
    \label{fig:bop_history_input}
  \end{subfigure}
    \hfill
  \begin{subfigure}{0.48\textwidth}
    \includegraphics[width=\linewidth]{figures/bop/bop_context_plot_grand-valley_RMA.pdf}
    \caption{RMA}
    \label{fig:bop_RMA}
  \end{subfigure}
  \hfill
  \begin{subfigure}{0.48\textwidth}
    \includegraphics[width=\linewidth]{figures/bop/bop_context_plot_grand-valley_SPARC.pdf}
    \caption{SPARC}
    \label{fig:bop_sparc}
  \end{subfigure}
  \hfill
  \begin{subfigure}{0.48\textwidth}
    \includegraphics[width=\linewidth]{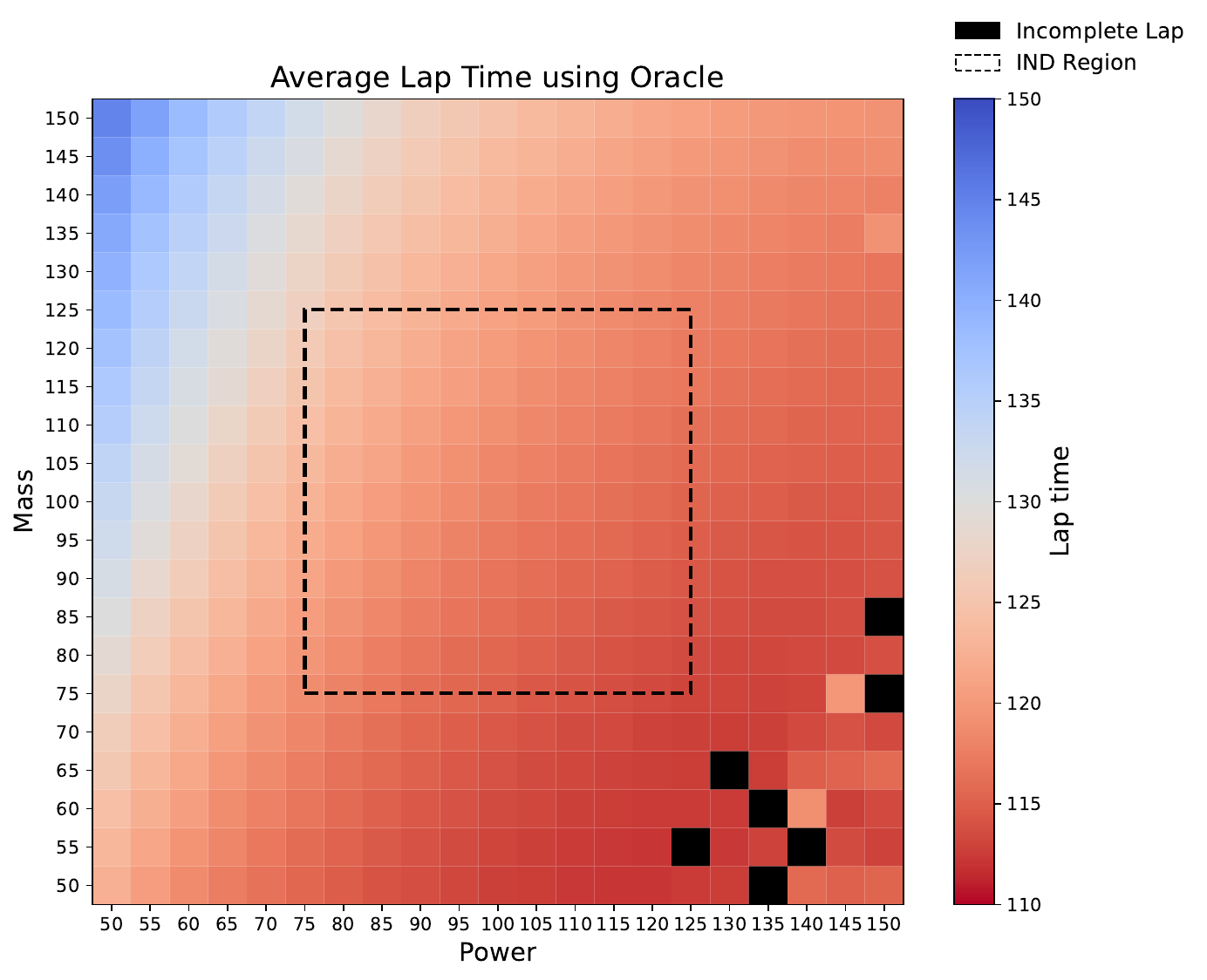}
    \caption{Oracle}
    \label{fig:bop_oracle}
  \end{subfigure}
  \caption{Lap times on the \emph{Power \& Mass} experiment.
  Colours denote average lap time (red\ =\ fast, blue\ =\ slow); black squares indicate at least one unfinished lap; the dashed box is the IND region.
  SPARC completes almost the entire grid and keeps lap times low even in the most extreme OOD settings, whereas other baselines either slow down notably or fail to finish.}
  \label{fig:bop_all}
\end{figure}

\begin{figure}[H]
  \centering
  \begin{subfigure}{0.46\textwidth}
    \includegraphics[width=\linewidth]{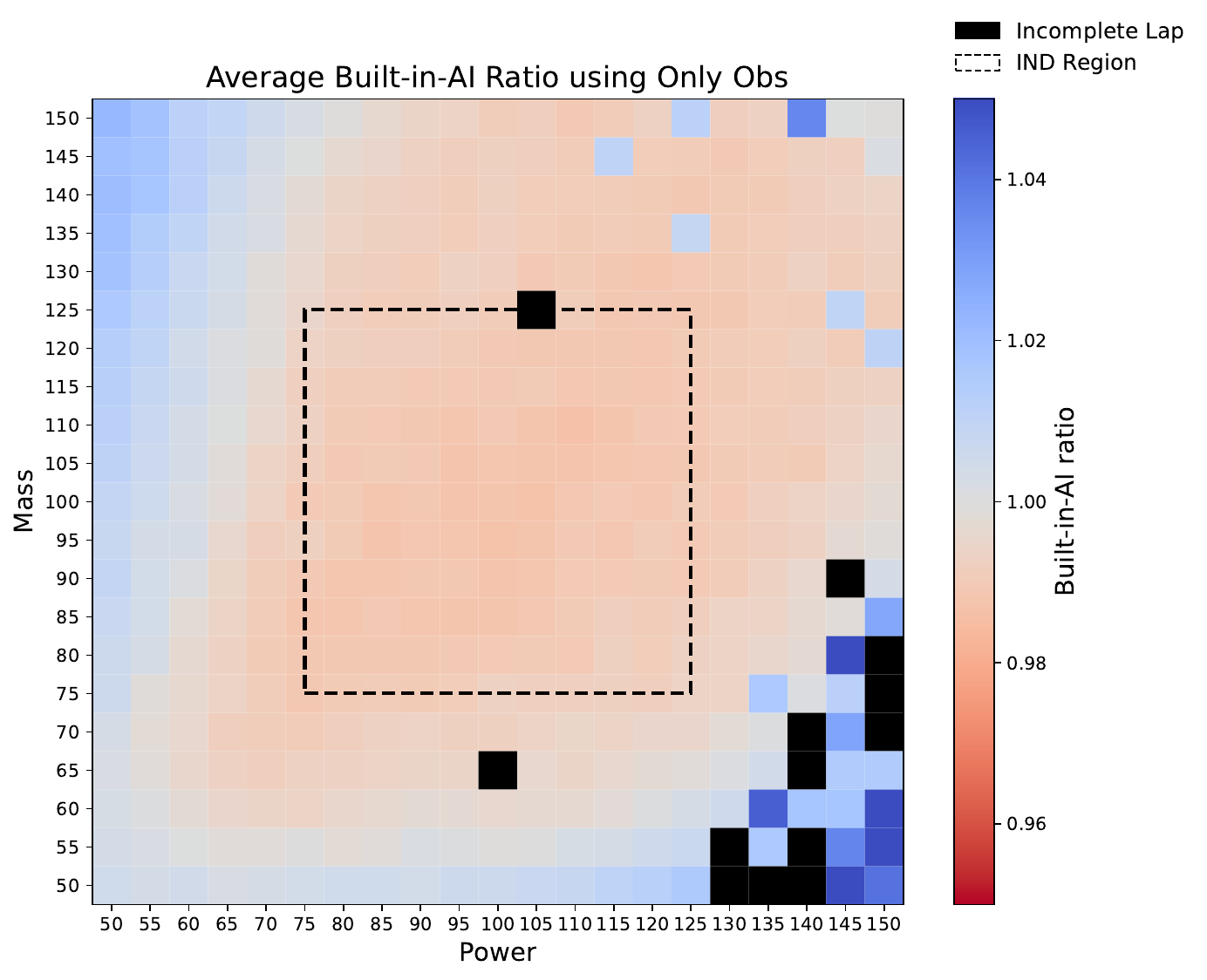}
    \caption{Only Obs}
    \label{fig:bop_biai_only_obs}
  \end{subfigure}
  \hfill
  \begin{subfigure}{0.46\textwidth}
    \includegraphics[width=\linewidth]{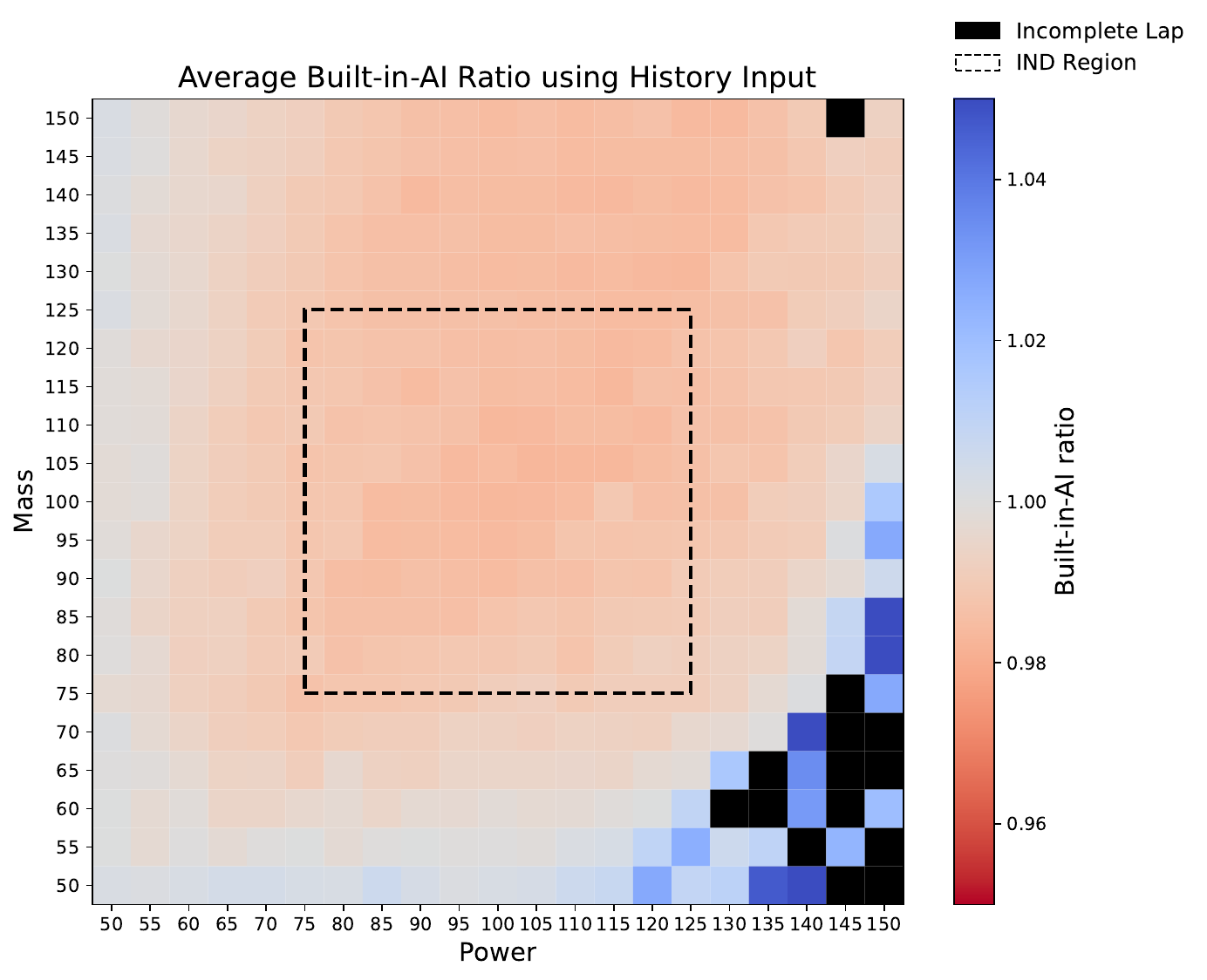}
    \caption{History Input}
    \label{fig:bop_biai_history_input}
  \end{subfigure}
  \hfill
    \hfill
  \begin{subfigure}{0.46\textwidth}
    \includegraphics[width=\linewidth]{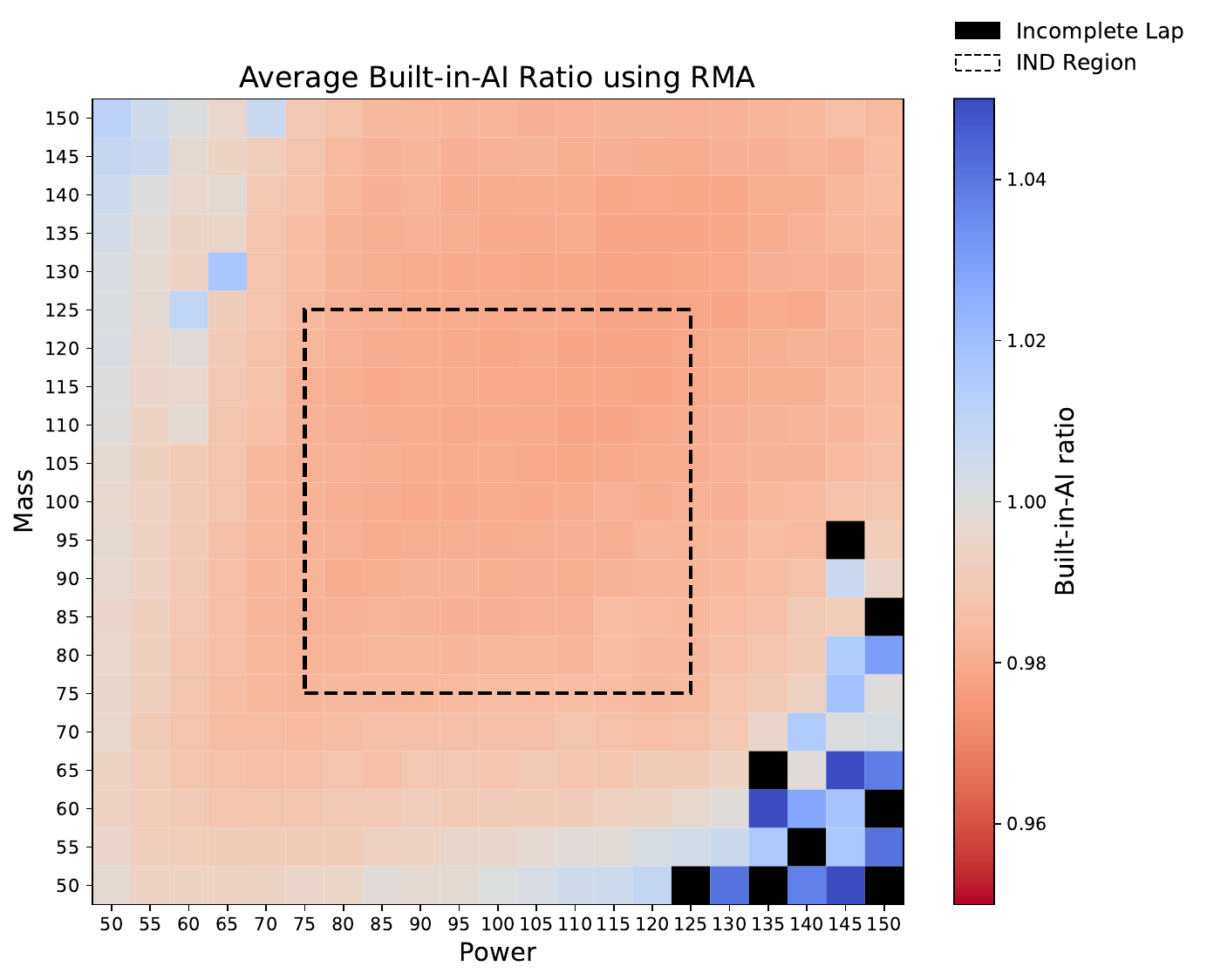}
    \caption{RMA}
    \label{fig:bop_biai_RMA}
  \end{subfigure}
  \hfill
  \begin{subfigure}{0.46\textwidth}
    \includegraphics[width=\linewidth]{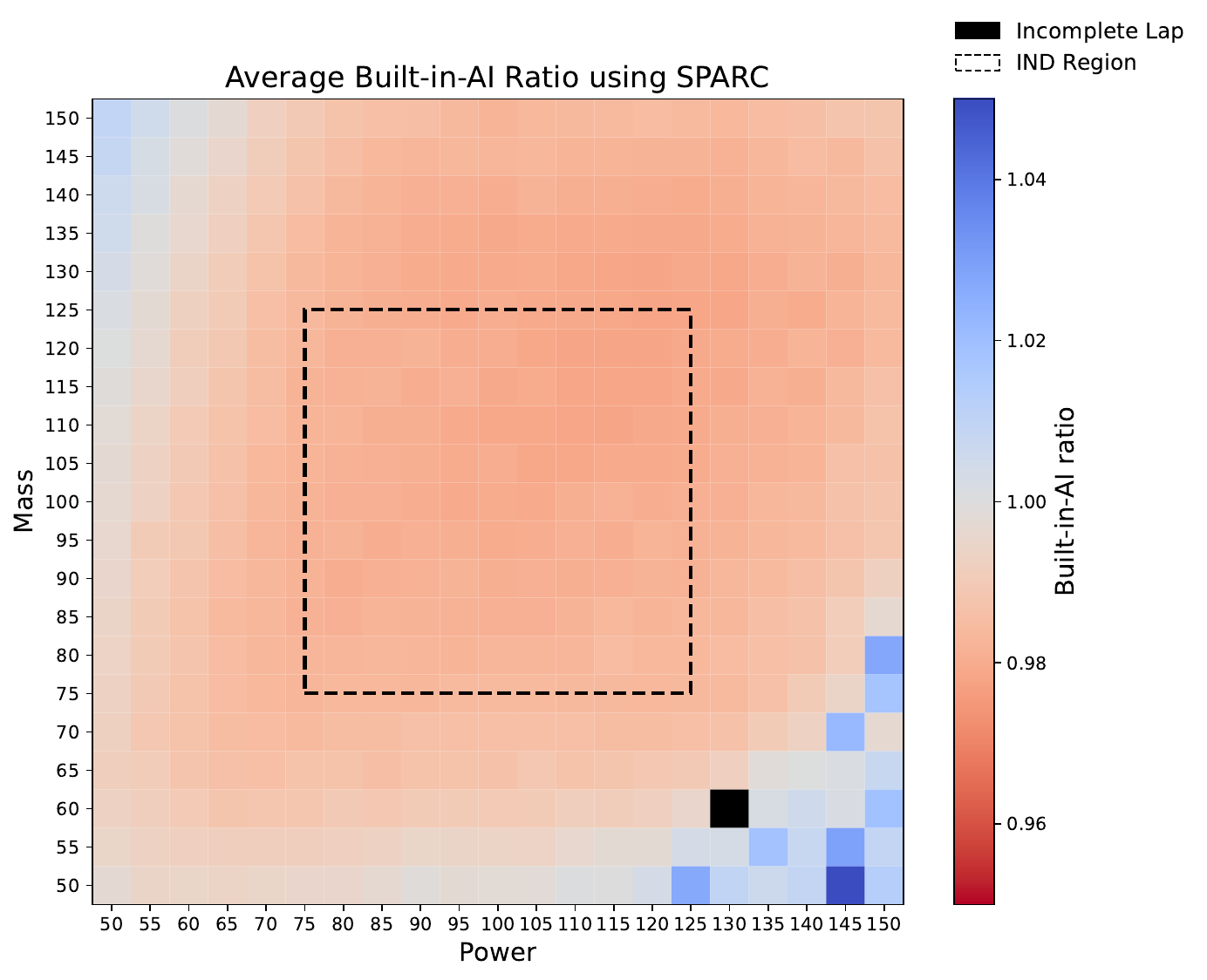}
    \caption{SPARC}
    \label{fig:bop_biai_sparc}
  \end{subfigure}
  \hfill
  \begin{subfigure}{0.46\textwidth}
    \includegraphics[width=\linewidth]{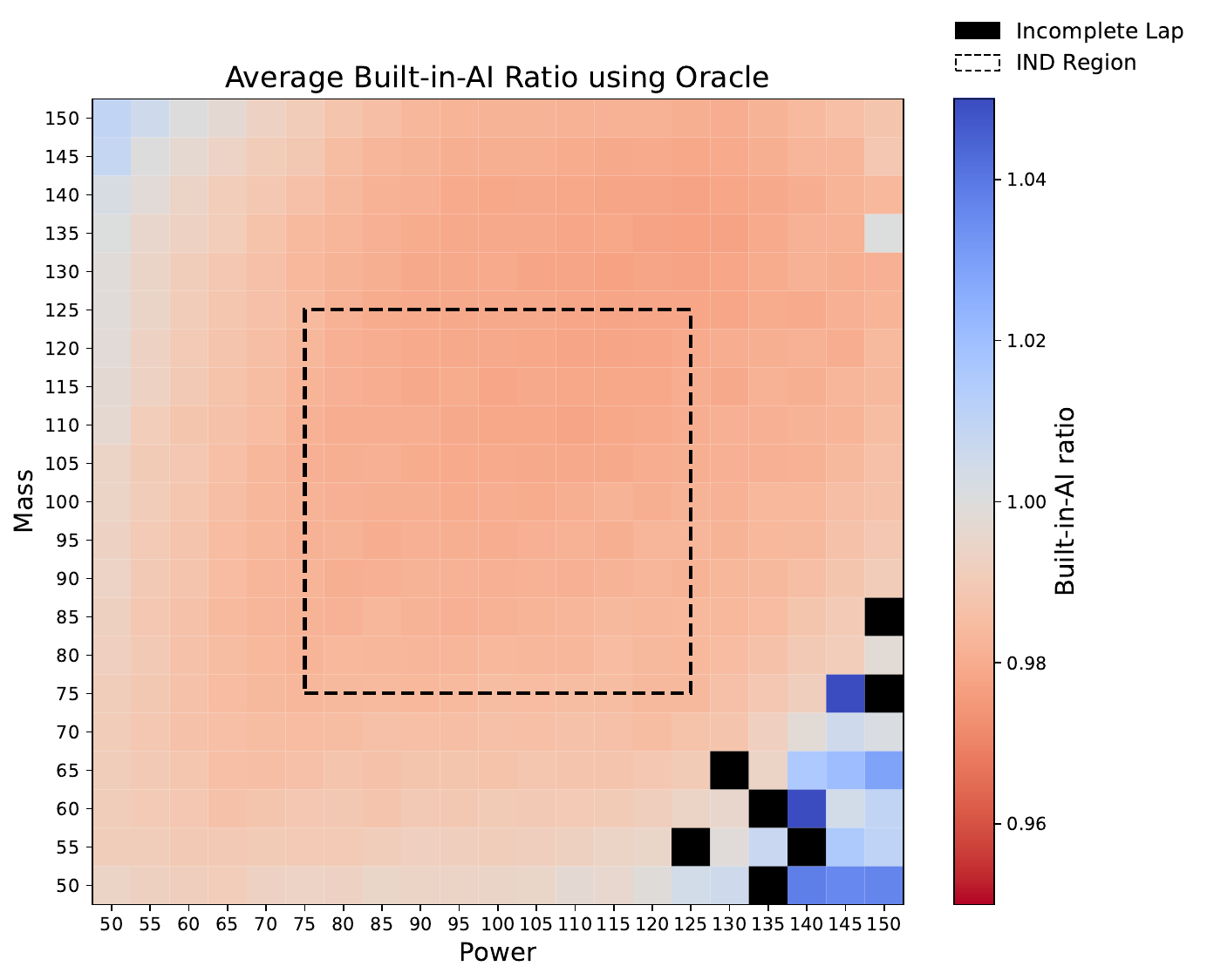}
    \caption{Oracle}
    \label{fig:bop_biai_oracle}
  \end{subfigure}
  \caption{
  Built-in-AI ratios on the \emph{Power \& Mass} experiment.   
  Normalising by the built‑in AI removes the effect of absolute car speed; values above 1.0 (blue) indicate slower than the scripted driver.
  SPARC attains sub‑BIAI lap times over the widest OOD area and shows the fewest non‑completions, highlighting its superior adaptability.}
  \label{fig:bop_biai_all}
\end{figure}

\paragraph{OOD results.}
\Cref{fig:bop_all,fig:bop_biai_all} visualise performance across the entire \([50\%,150\%]^2\) grid.  
Colour encodes either raw lap time (\Cref{fig:bop_all}) or BIAI ratio (\Cref{fig:bop_biai_all}); cooler colours indicate slower driving.  
Outside the training box, Only‑Obs, History‑Input, and RMA accumulate many black squares in the lower‑right corner (the difficult contextual setting of light cars with powerful engines), whereas SPARC only fails in a single extreme cell and achieves visibly faster laps over the viable region.  
This qualitative picture matches the quantitative OOD numbers in \Cref{tab:bop_ood}: SPARC completes 99.9\% of OOD settings and records the lowest average BIAI ratio (0.9907).

\begin{table}[b]
    \centering
    \caption{Performance summary of the \textit{Power \& Mass} experiments (\textbf{IND}), averaged over 3 seeds. Results show the mean built-in-AI lap-time ratios and the percentage of settings with a successfully completed lap ($\pm$ s.e.m.).
    All methods finish almost every lap; SPARC is tied with RMA and Oracle for fastest average time.}
    \label{tab:bop_ind}
    \resizebox{0.52\linewidth}{!}{%
    \begin{tabular}{l@{\hspace{1em}}cc}
        \toprule
        \multicolumn{1}{c}{Method} & \multicolumn{1}{c}{Built-in-AI lap-time ratio ($\downarrow$)} & \multicolumn{1}{c}{Success ($\uparrow$)} \\
        \midrule
        Only Obs          & 0.9924 $\pm$ 0.0028 & 99.72 $\pm$ 0.28 \\
        History Input     & 0.9868 $\pm$ 0.0017 & 100.00 $\pm$ 0.00 \\
        RMA               & 0.9810 $\pm$ 0.0002 & 100.00 $\pm$ 0.00 \\
        \colorRow
        SPARC             & 0.9810 $\pm$ 0.0002 & 100.00 $\pm$ 0.00 \\
        Oracle            & 0.9811 $\pm$ 0.0004 & 100.00 $\pm$ 0.00 \\
        \bottomrule
    \end{tabular}%
    }
\end{table}

\subsection{MuJoCo Experiments}
\label{sec:add_results_mujoco}

\begin{figure}[H]
  \centering
  \begin{subfigure}{0.48\textwidth}
    \includegraphics[width=\linewidth]{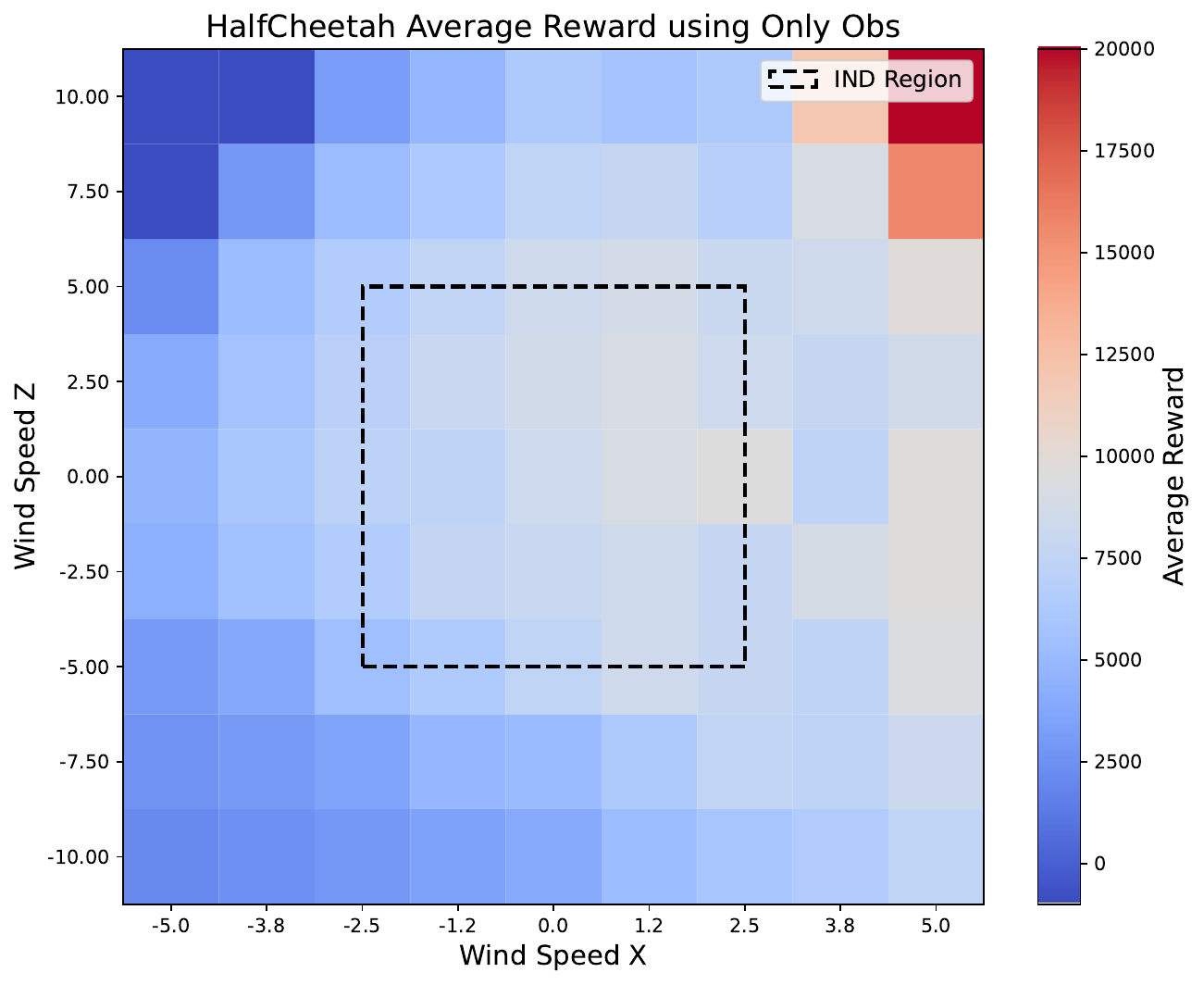}
    \caption{Only Obs}
    \label{fig:halfcheetah_only_obs}
  \end{subfigure}
  \hfill
  \begin{subfigure}{0.48\textwidth}
    \includegraphics[width=\linewidth]{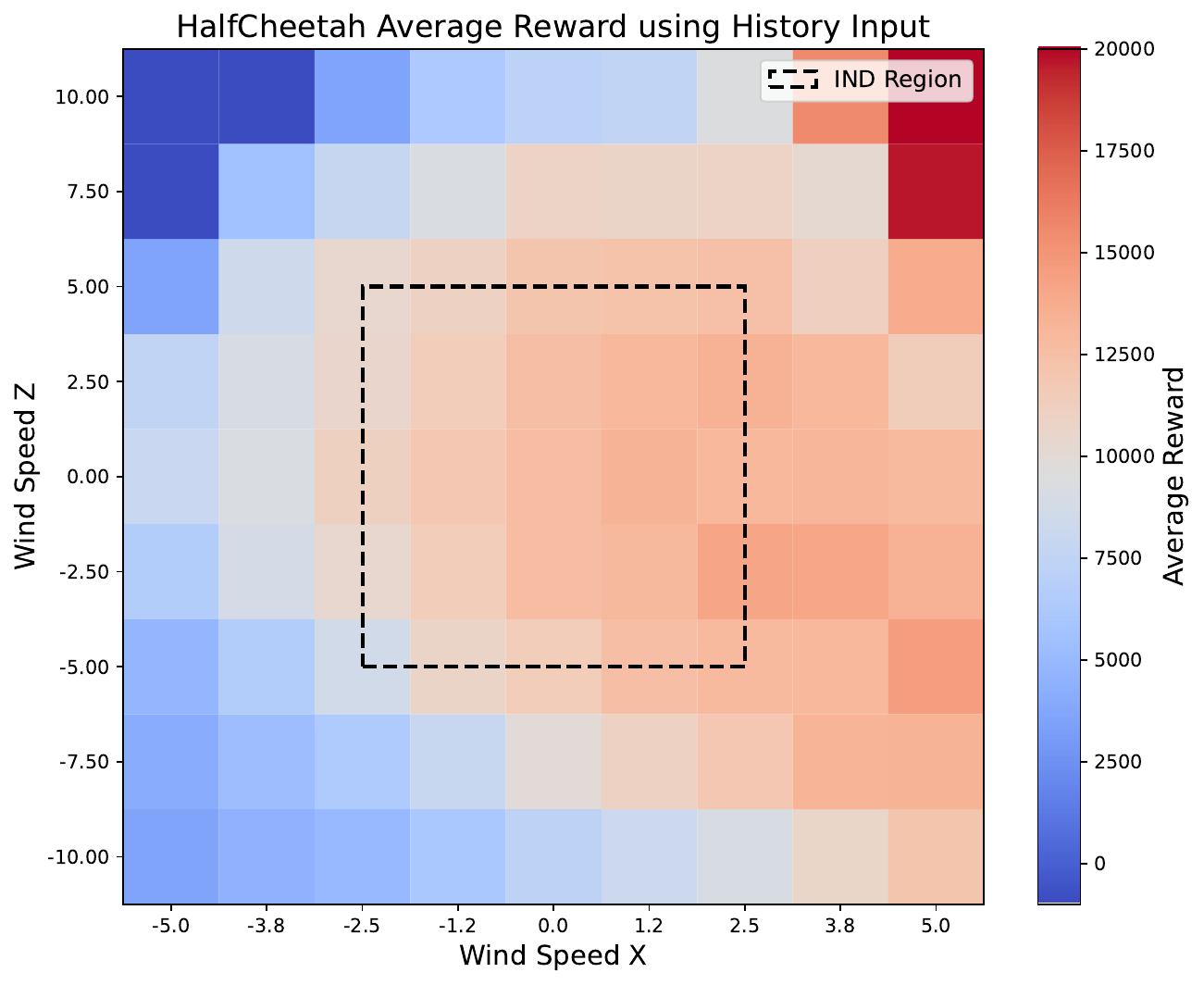}
    \caption{History Input}
    \label{fig:halfcheetah_history_input}
  \end{subfigure}
    \hfill
  \begin{subfigure}{0.48\textwidth}
    \includegraphics[width=\linewidth]{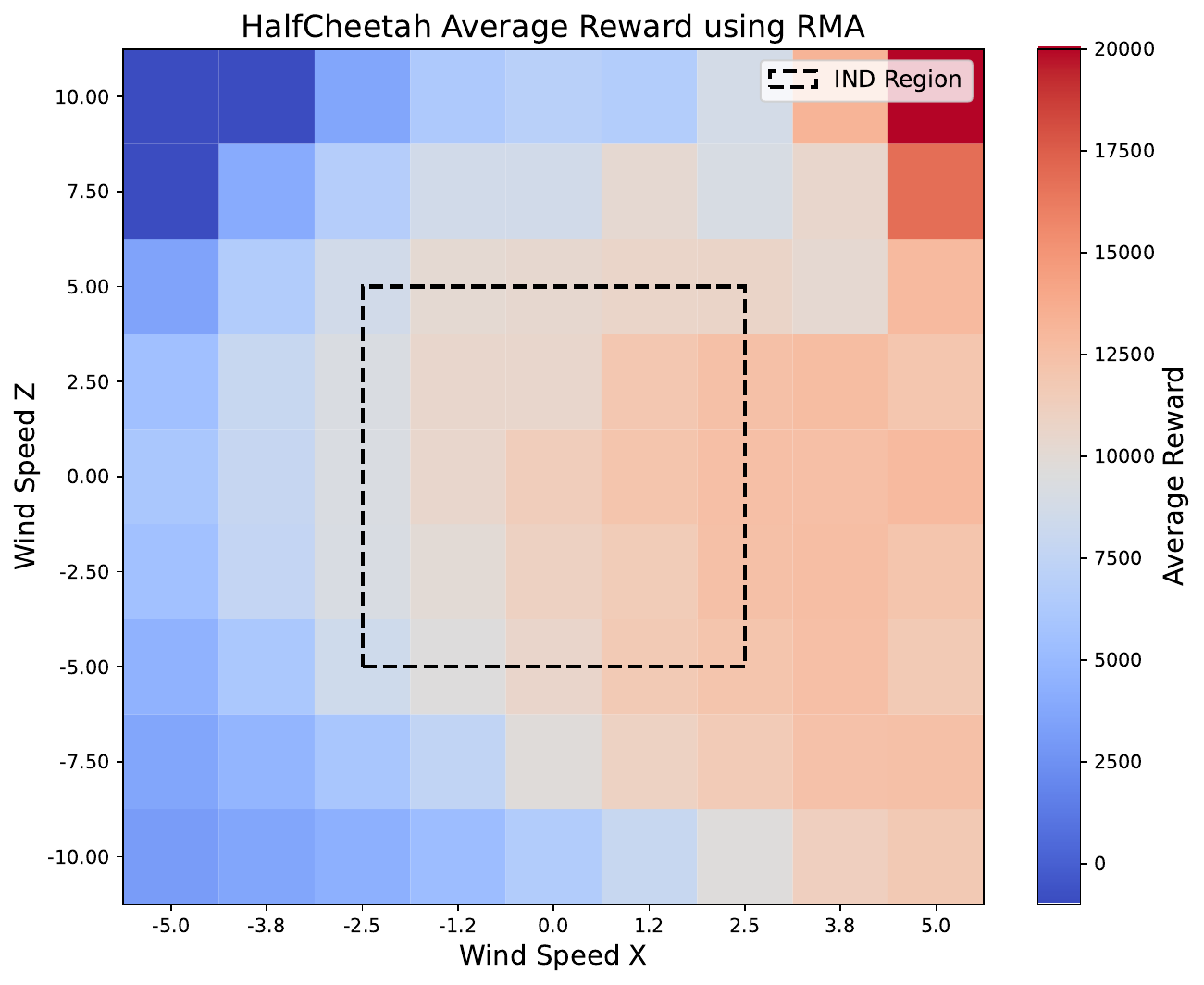}
    \caption{RMA}
    \label{fig:halfcheetah_RMA}
  \end{subfigure}
  \hfill
  \begin{subfigure}{0.48\textwidth}
    \includegraphics[width=\linewidth]{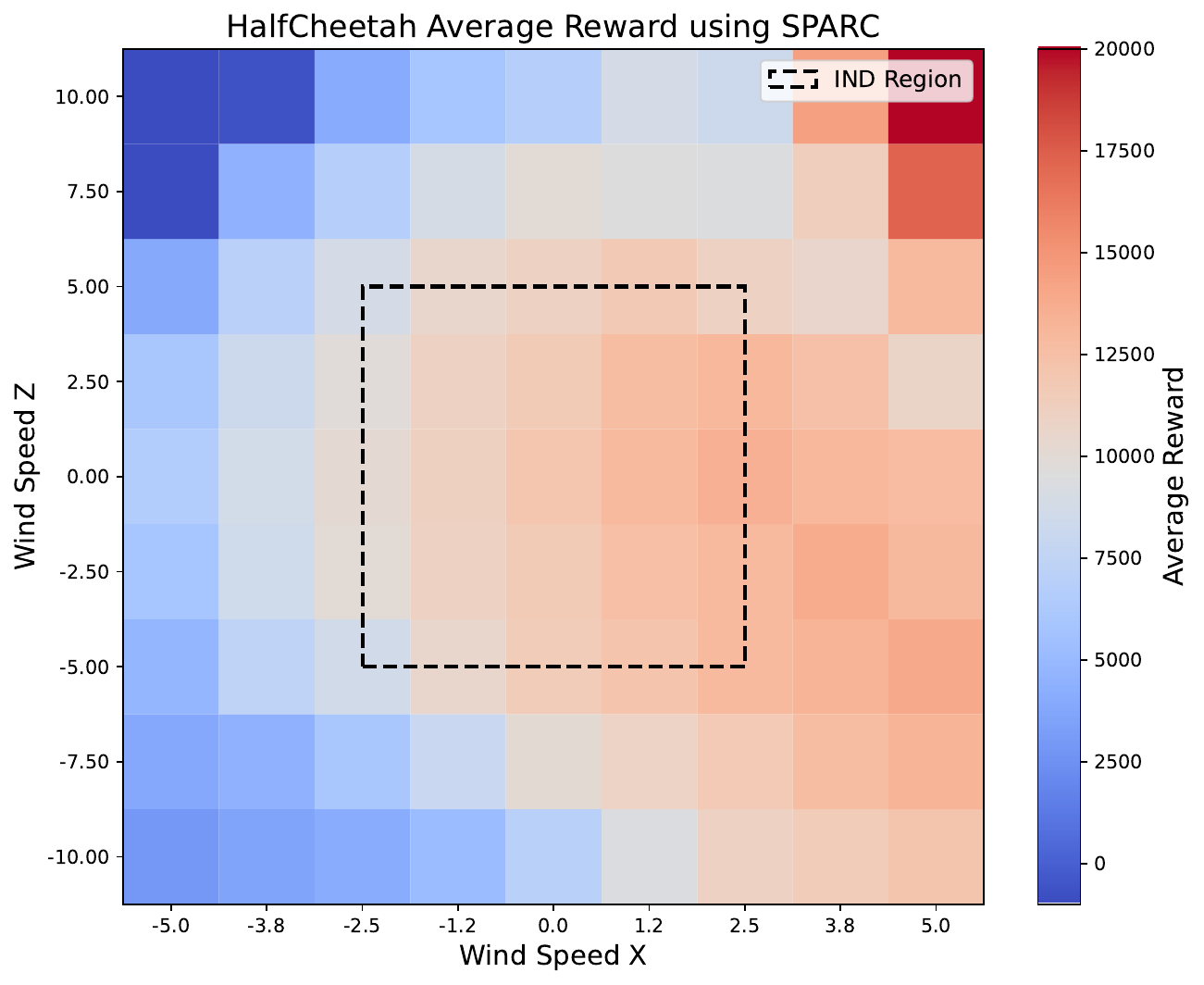}
    \caption{SPARC}
    \label{fig:halfcheetah_sparc}
  \end{subfigure}
  \hfill
  \begin{subfigure}{0.48\textwidth}
    \includegraphics[width=\linewidth]{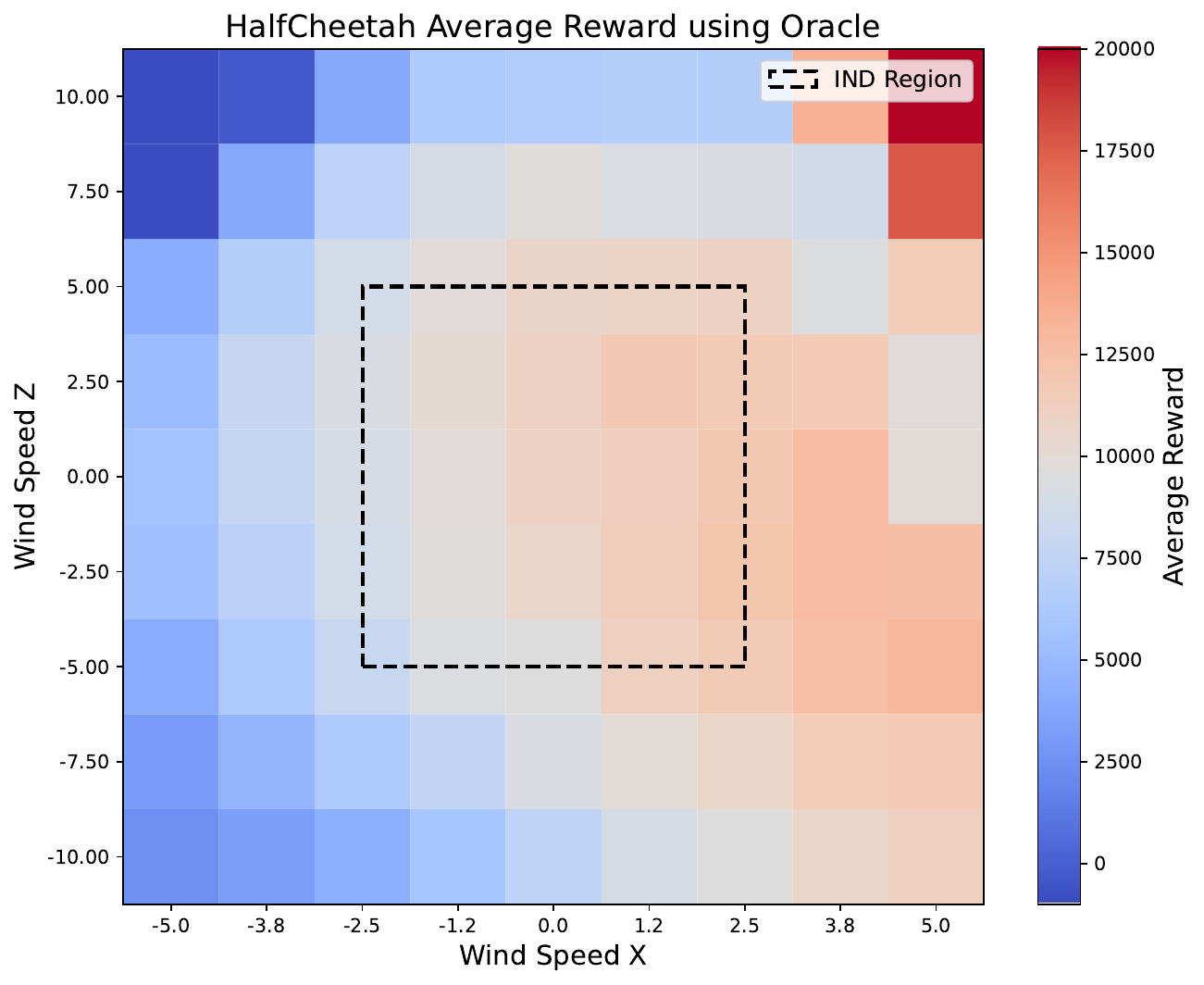}
    \caption{Oracle}
    \label{fig:halfcheetah_oracle}
  \end{subfigure}
  \caption{
  Average episode return of methods on \textbf{HalfCheetah-v5} with different wind perturbations (red is high reward, blue is low).  
  SPARC retains near‑Oracle performance throughout the grid.}
  \label{fig:halfcheetah_all}
\end{figure}

\begin{figure}[H]
  \centering
  \begin{subfigure}{0.48\textwidth}
    \includegraphics[width=\linewidth]{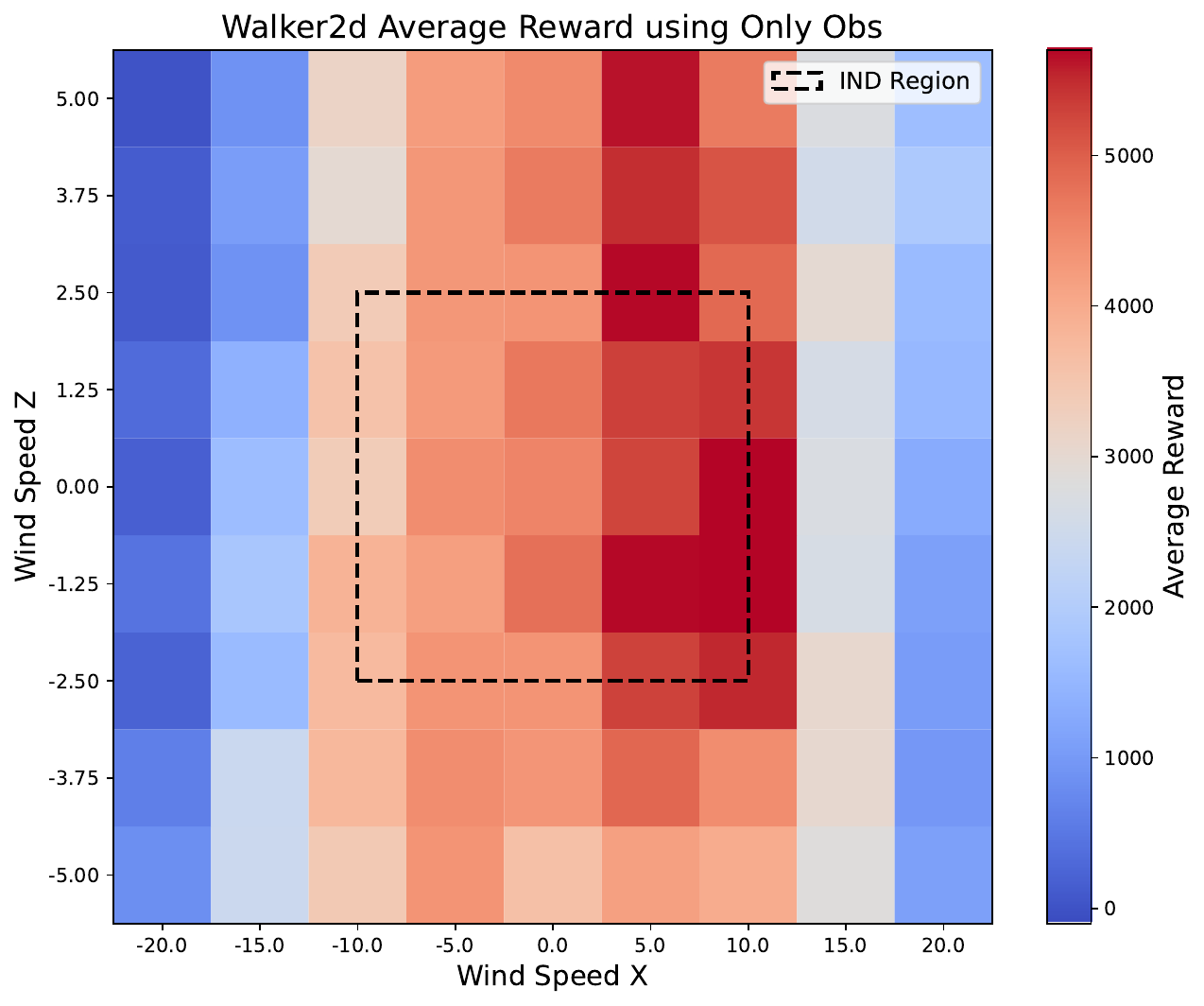}
    \caption{Only Obs}
    \label{fig:walker2d_only_obs}
  \end{subfigure}
  \hfill
  \begin{subfigure}{0.48\textwidth}
    \includegraphics[width=\linewidth]{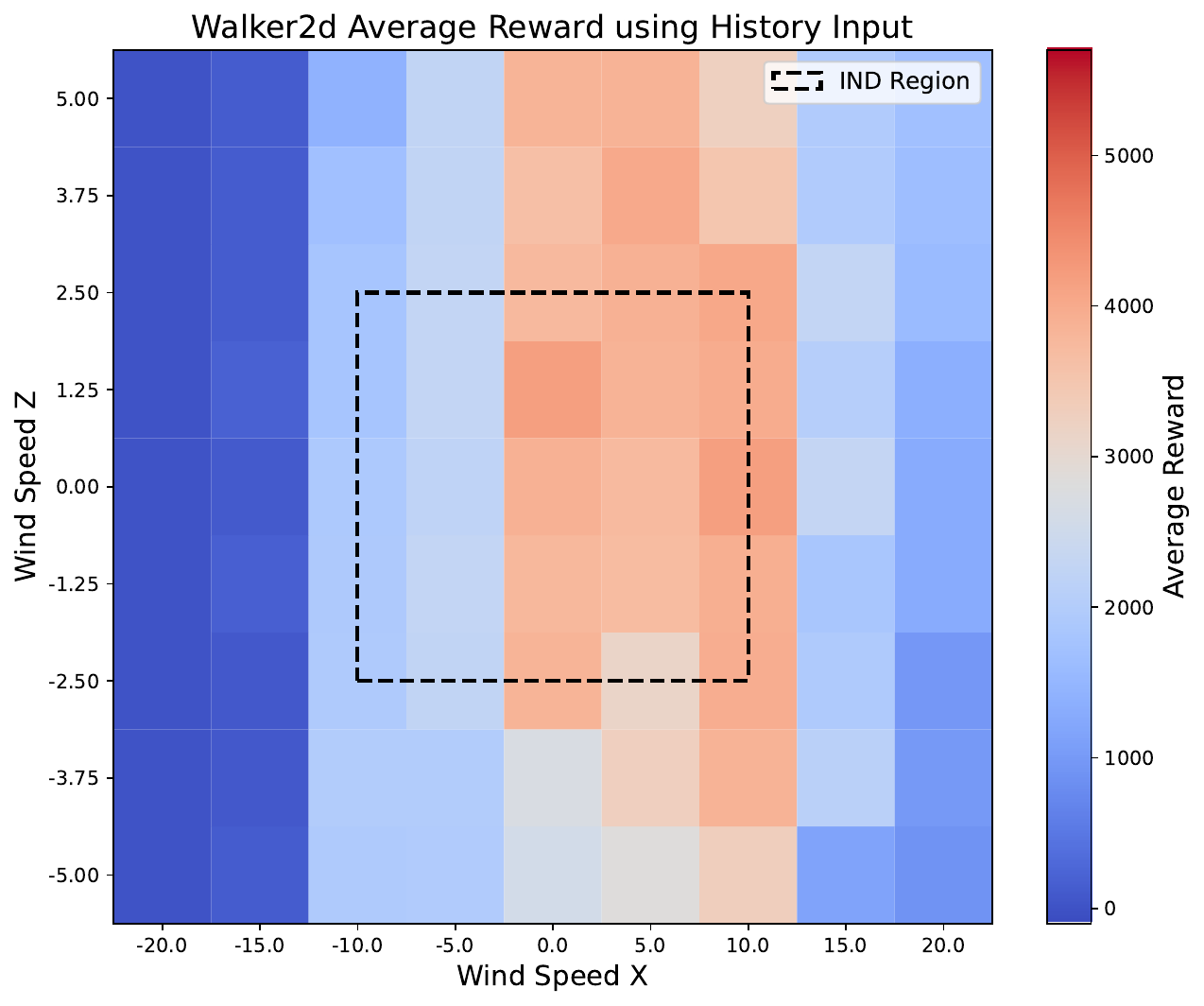}
    \caption{History Input}
    \label{fig:walker2d_history_input}
  \end{subfigure}
    \hfill
  \begin{subfigure}{0.48\textwidth}
    \includegraphics[width=\linewidth]{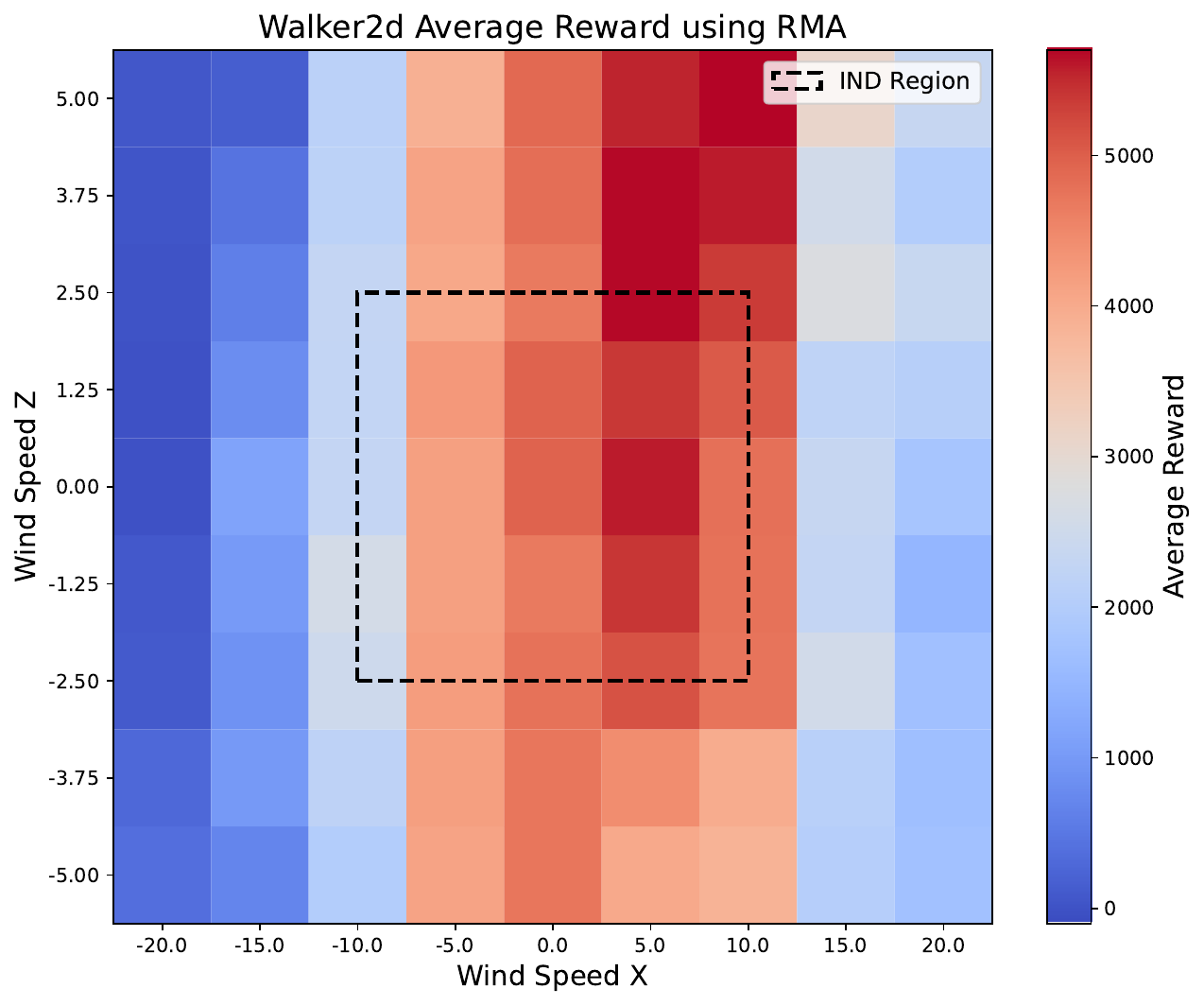}
    \caption{RMA}
    \label{fig:walker2d_RMA}
  \end{subfigure}
  \hfill
  \begin{subfigure}{0.48\textwidth}
    \includegraphics[width=\linewidth]{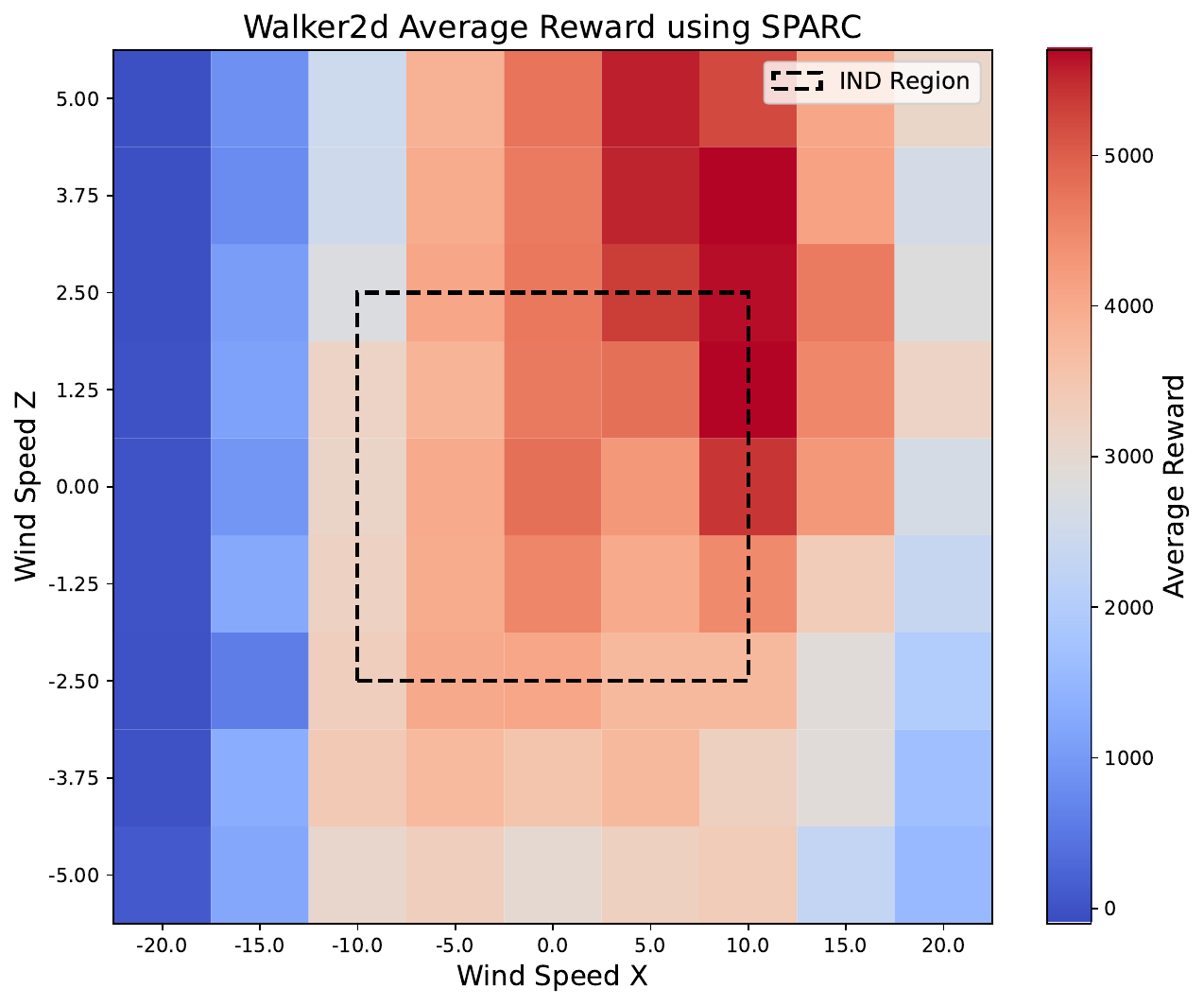}
    \caption{SPARC}
    \label{fig:walker2d_sparc}
  \end{subfigure}
  \hfill
  \begin{subfigure}{0.48\textwidth}
    \includegraphics[width=\linewidth]{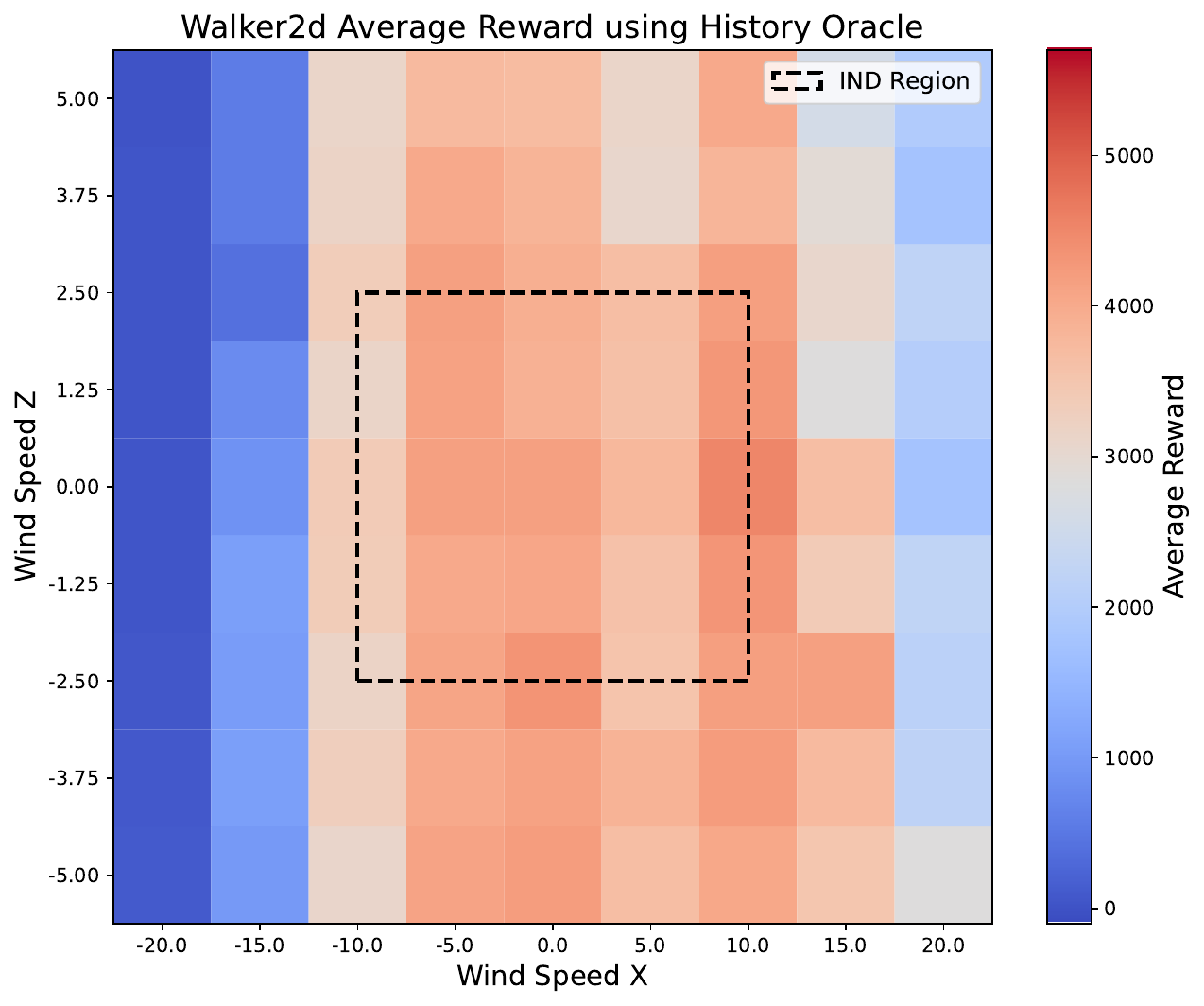}
    \caption{Oracle}
    \label{fig:walker2d_oracle}
  \end{subfigure}
  \caption{
  Average episode return of methods on \textbf{Walker2d-v5} with different wind perturbations (red is high reward, blue is low). 
  Returns exhibit a characteristic ridge along positive \(x\) winds where the walker is pushed forwards.  
  SPARC secures the highest rewards and records fewer falls than competing methods in many OOD settings.}
  \label{fig:walker2d_all}
\end{figure}

To probe fine‑grained robustness in continuous‑control domains, in \Cref{sec:mujoco_results} we add a \emph{horizontal} (\(x\)) and \emph{vertical} (\(z\)) wind force to every MuJoCo agent, as context within the environment. 
We present results of each baseline on all three environments (HalfCheetah-v5, Hopper-v5, Walker2d-v5) in \Cref{fig:halfcheetah_all,fig:hopper_all,fig:walker2d_all}. Each heat‑map cell shows the mean episode return over three seeds; darker red is higher reward, darker blue is lower.  
SPARC consistently matches or exceeds other baselines \emph{inside} the dashed IND region and, crucially, maintains high returns as winds grow stronger. 

\begin{figure}[H]
  \centering
  \begin{subfigure}{0.48\textwidth}
    \includegraphics[width=\linewidth]{figures/mujoco/wind_plot_halfcheetah_SPARC_delta_PRGn.pdf}
    \caption{HalfCheetah}
    \label{fig:mujoco_delta_halfcheetah_prgn}
  \end{subfigure}
  \hfill
  \begin{subfigure}{0.48\textwidth}
    \includegraphics[width=\linewidth]{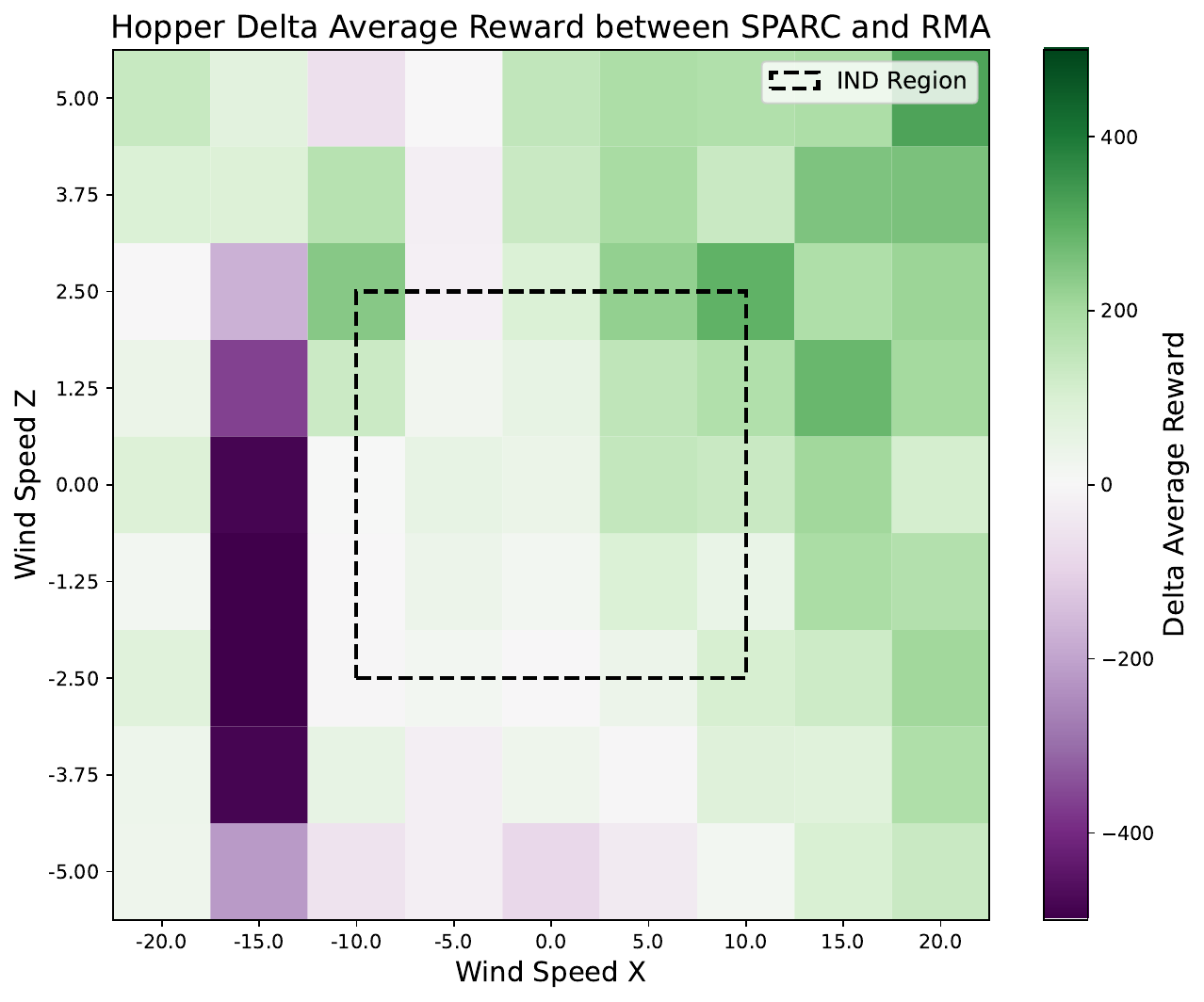}
    \caption{Hopper}
    \label{fig:mujoco_delta_hopper_prgn}
  \end{subfigure}
  \hfill
  \begin{subfigure}{0.48\textwidth}
    \includegraphics[width=\linewidth]{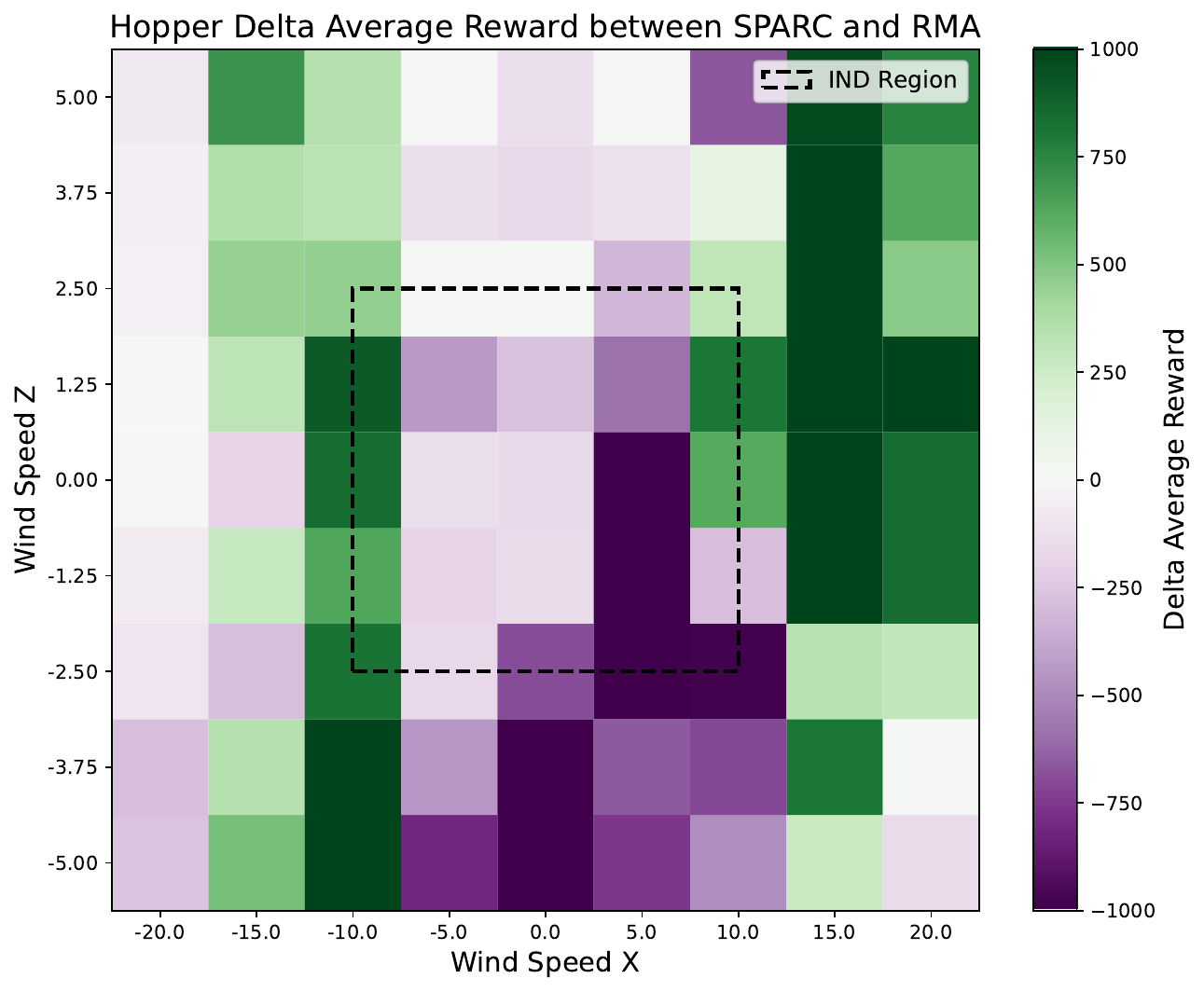}
    \caption{Walker2d}
    \label{fig:mujoco_delta_walker2d_prgn}
  \end{subfigure}
  \caption{
  Delta average return of SPARC vs RMA for MuJoCo environments with different wind perturbations. In \textcolor{deltagreen}{green: SPARC is better} in that wind setting, while in \textcolor{deltapurple}{purple: RMA scores higher}. SPARC demonstrates a stronger robustness overall, especially in out-of-distribution contextual settings.
  }
  \label{fig:mujoco_delta_all_PRGn}
\end{figure}

\begin{figure}[H]
  \centering
  \begin{subfigure}{0.48\textwidth}
    \includegraphics[width=\linewidth]{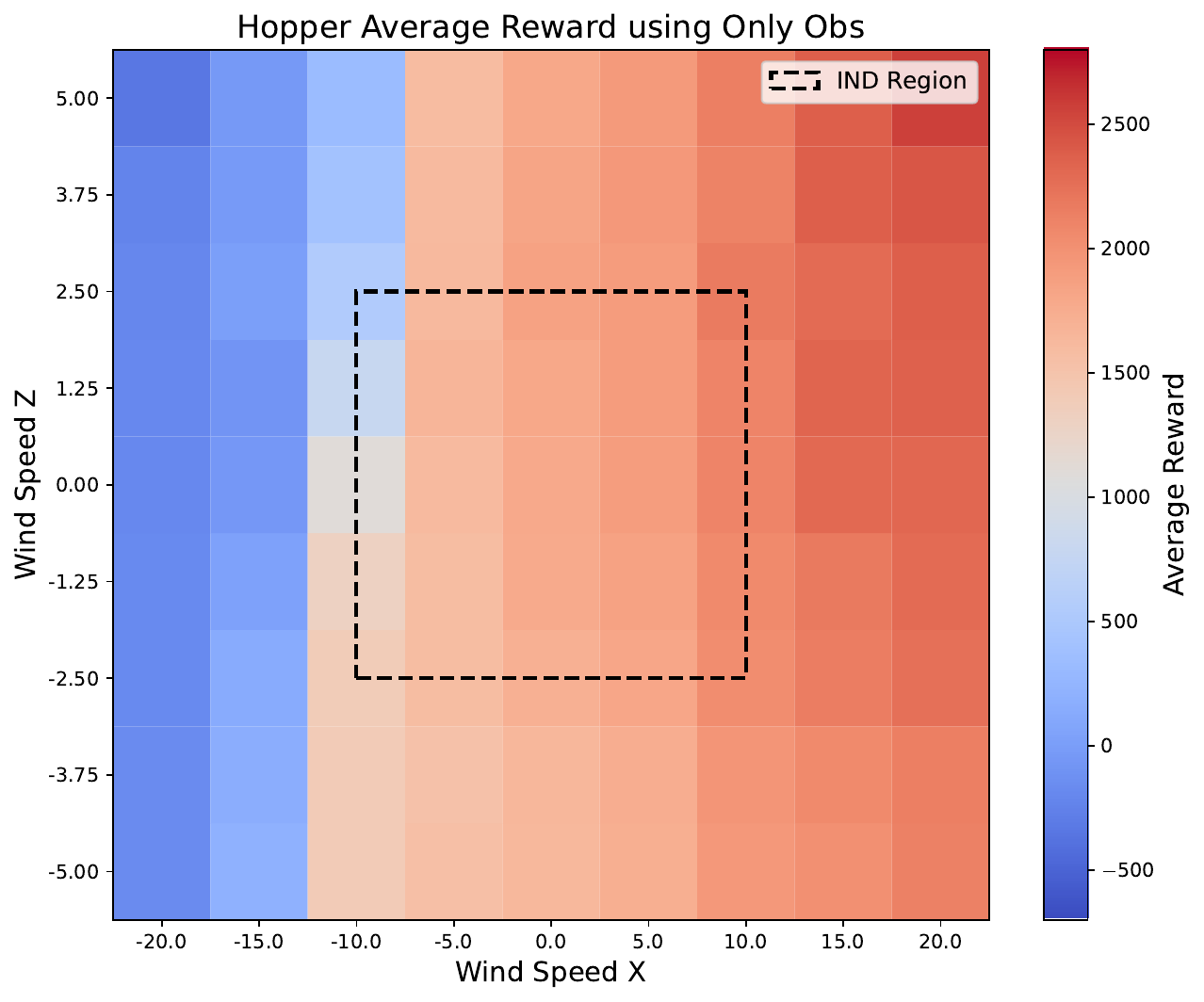}
    \caption{Only Obs}
    \label{fig:hopper_only_obs}
  \end{subfigure}
  \hfill
  \begin{subfigure}{0.48\textwidth}
    \includegraphics[width=\linewidth]{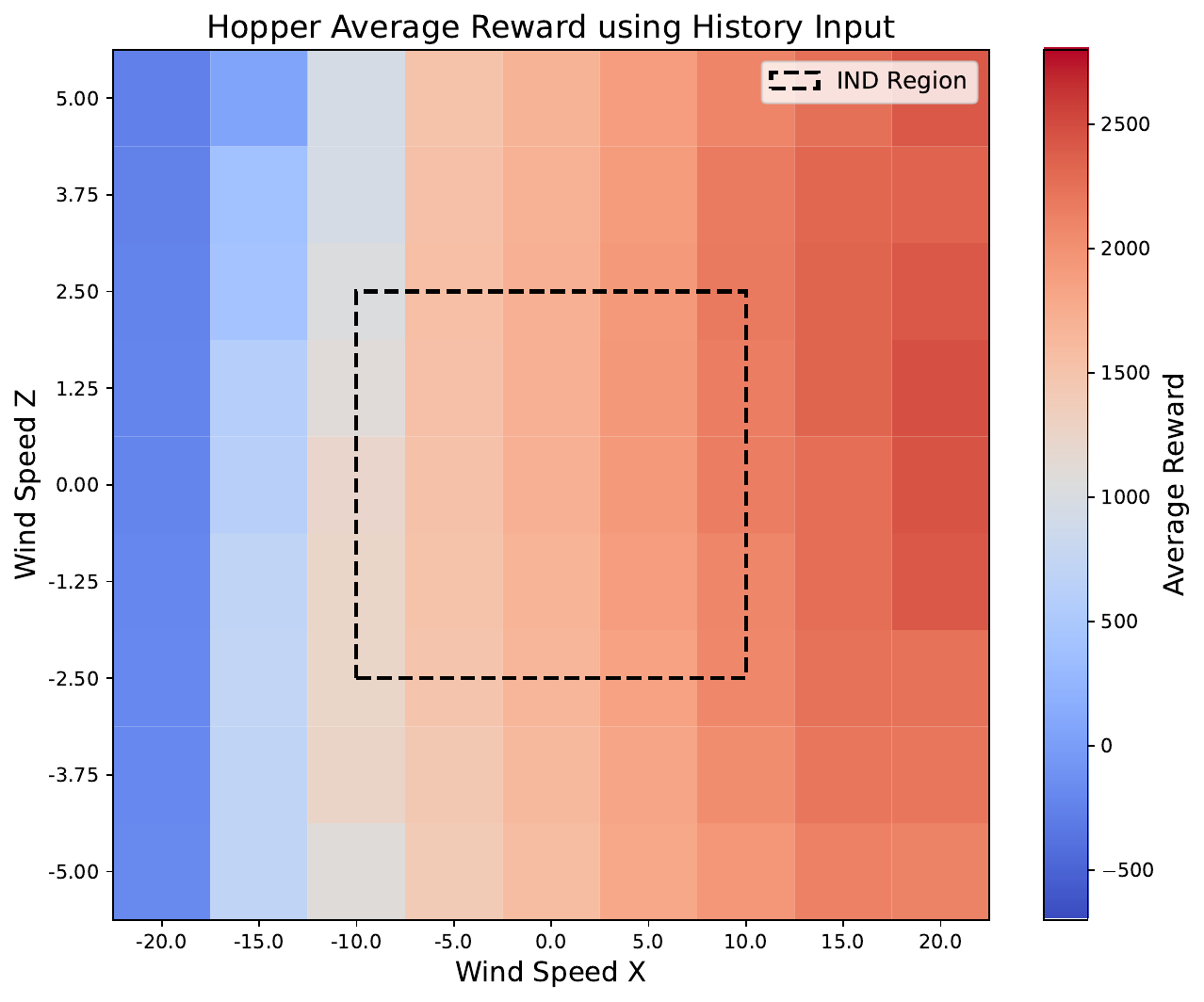}
    \caption{History Input}
    \label{fig:hopper_history_input}
  \end{subfigure}
    \hfill
  \begin{subfigure}{0.48\textwidth}
    \includegraphics[width=\linewidth]{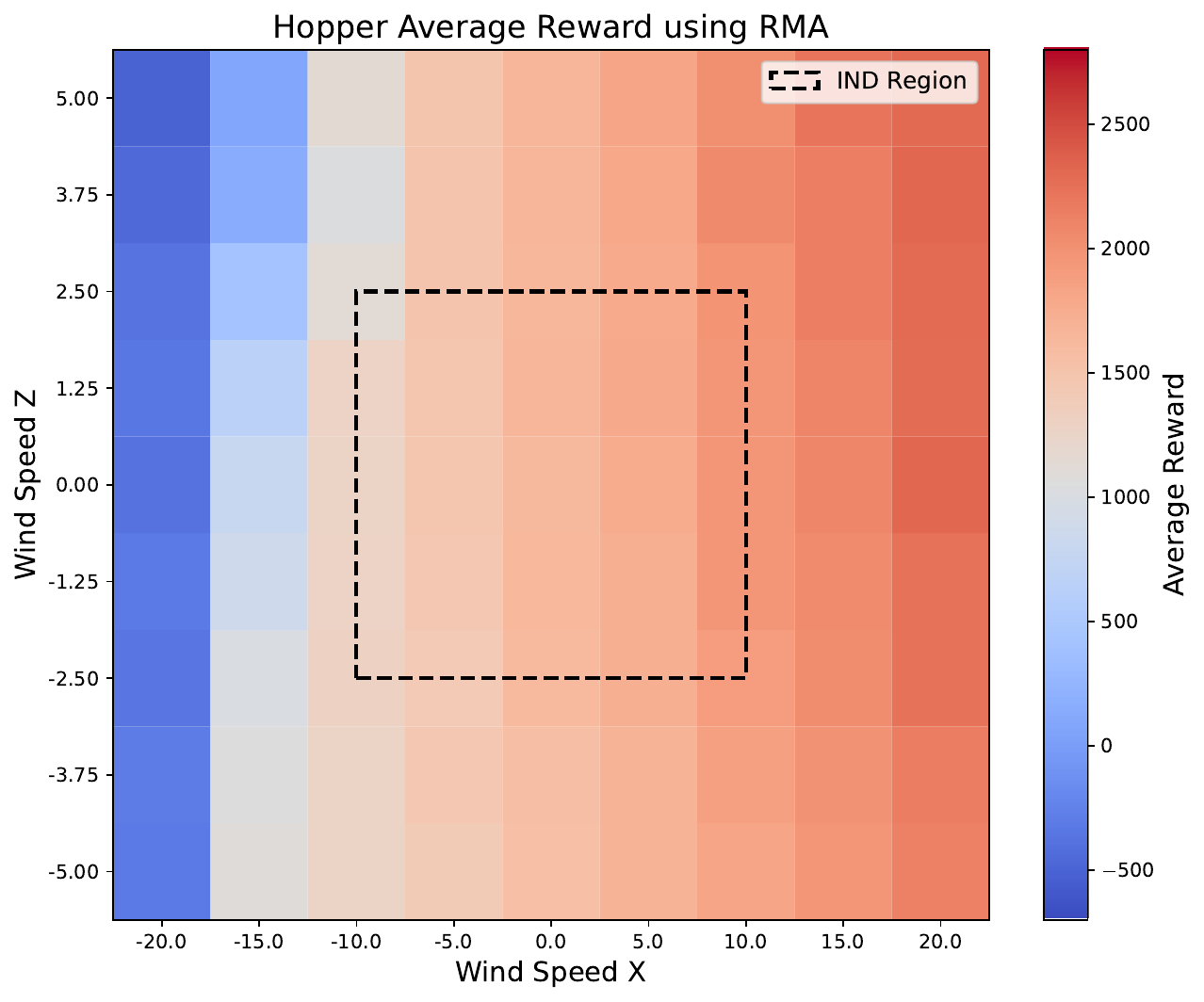}
    \caption{RMA}
    \label{fig:hopper_RMA}
  \end{subfigure}
  \hfill
  \begin{subfigure}{0.48\textwidth}
    \includegraphics[width=\linewidth]{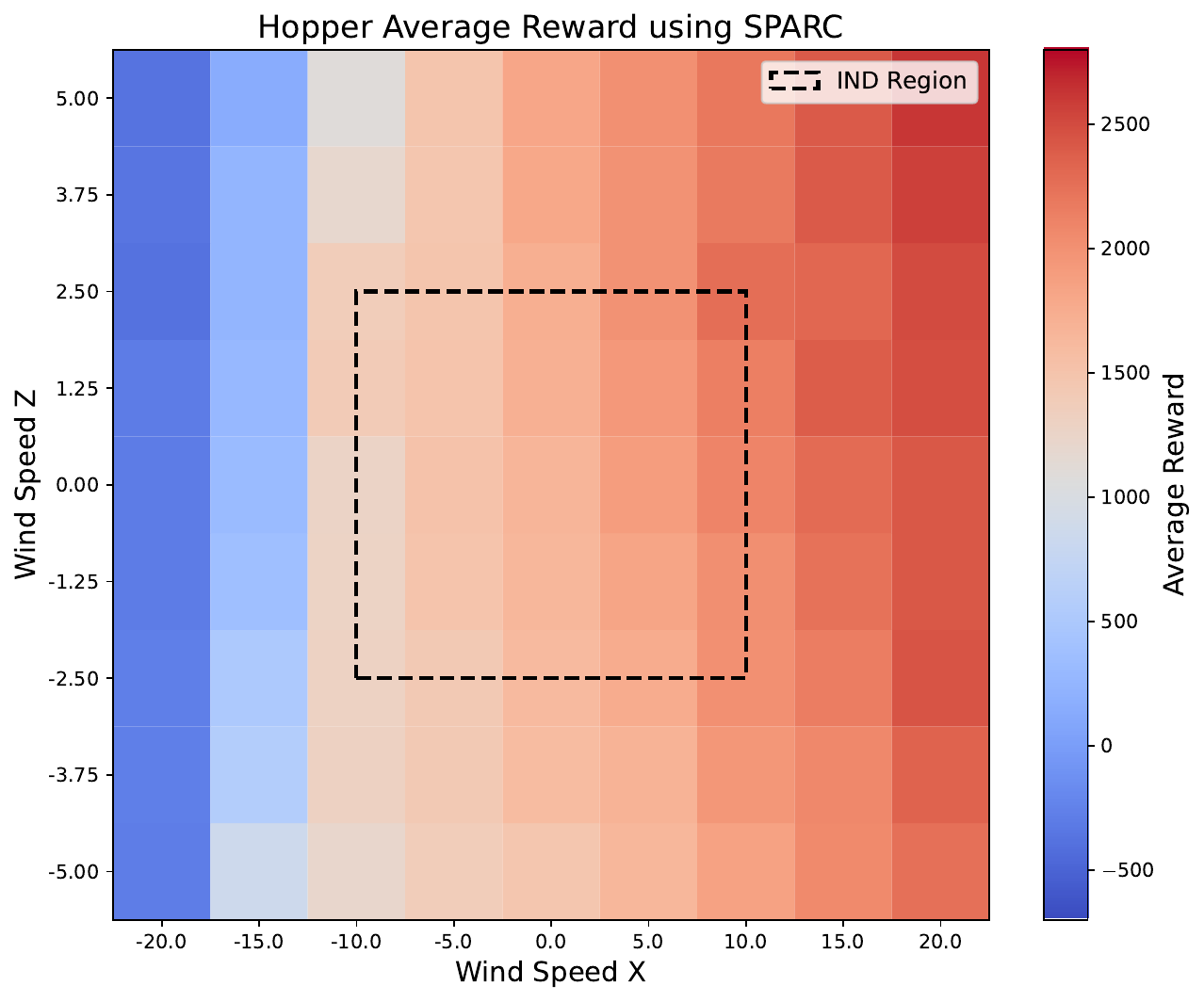}
    \caption{SPARC}
    \label{fig:hopper_sparc}
  \end{subfigure}
  \hfill
  \begin{subfigure}{0.48\textwidth}
    \includegraphics[width=\linewidth]{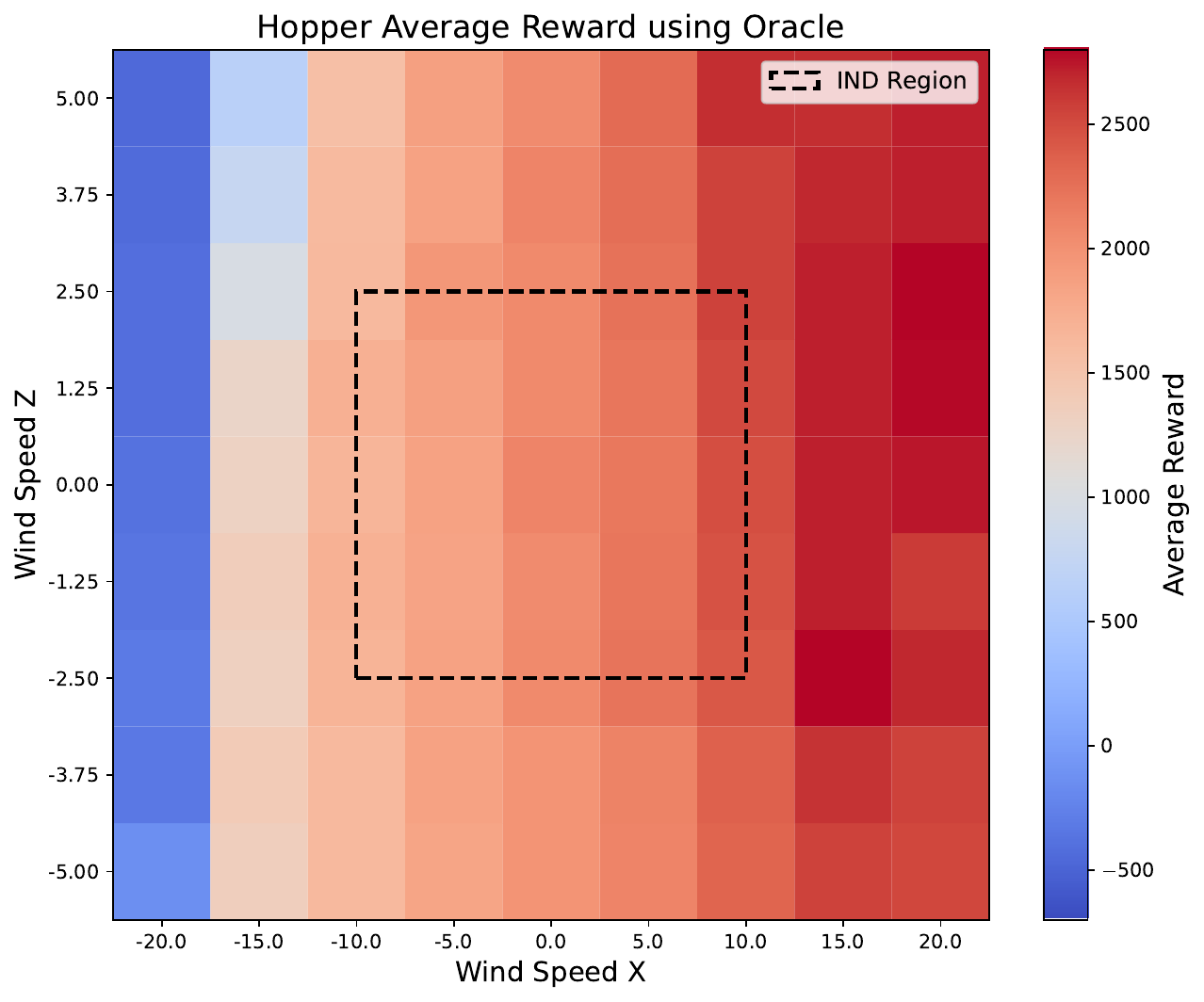}
    \caption{Oracle}
    \label{fig:hopper_oracle}
  \end{subfigure}
  \caption{
  Average episode return of methods on \textbf{Hopper-v5} with different wind perturbations (red is high reward, blue is low). 
  SPARC preserves stable hopping across almost the entire grid and outperforms RMA on most OOD cells (see also \Cref{fig:mujoco_delta_all_PRGn}).}
  \label{fig:hopper_all}
\end{figure}

The difference plot in \Cref{fig:mujoco_delta_all_PRGn} makes an explicit comparison between the strong RMA baseline \cite{kumar2021rma} and SPARC. Most OOD cells are green (SPARC\ $>$\ RMA), confirming that the history adapter can indeed be trained simultaneously with the expert policy to generalise across a two‑dimensional disturbance space.

\end{document}